%% file: main.tex
\documentclass{article}
\input{preamble_icml}

\input{numbers_r1}

\icmltitlerunning{Closed Gates Leak: Decomposing Adjacent-Layer Cancellation}

\begin{document}

\twocolumn[
\icmltitle{Closed Gates Leak: Decomposing Adjacent-Layer Cancellation\\in Gated Language Models}
\icmlsetsymbol{equal}{*}
\begin{icmlauthorlist}
\icmlauthor{Harvey Dam}{utah}
\icmlauthor{Ganesh Gopalakrishnan}{utah}
\end{icmlauthorlist}
\icmlaffiliation{utah}{Kahlert School of Computing, University of Utah, Salt Lake City, USA}
\icmlcorrespondingauthor{Harvey Dam}{\mbox{harvey.dam@utah.edu}}
\icmlkeywords{mechanistic interpretability, residual stream, gated MLP, SwiGLU, transformer}
\vskip 0.3in
]

\printAffiliationsAndNotice{}

\begin{abstract}
\input{abstract}
\end{abstract}

\input{intro}
\input{setup}
\input{law}
\input{mechanism}
\input{interventions}
\input{related}
\input{discussion}

\section*{Impact Statement}
This paper decomposes a measured property of trained language models and locates the neurons that carry it. It bears on how layer contributions are read in interpretability work and on activation-function design; we see no direct societal consequence beyond those of interpretability research in general.

\bibliographystyle{icml2026}
\bibliography{references}

\newpage
\appendix
\onecolumn
\input{appendix}

\end{document}

%% file: preamble_icml.tex
\usepackage{microtype}
\usepackage{graphicx}
\usepackage{booktabs}
\usepackage{hyperref}
\usepackage[preprint]{icml2026}
\usepackage[utf8]{inputenc}
\usepackage[T1]{fontenc}
\usepackage{amsmath}
\usepackage{amssymb}
\usepackage{amsthm}
\usepackage{float}
\usepackage{array}
\usepackage{multirow}
\usepackage{adjustbox}
\newcommand{\fittable}[1]{\begin{adjustbox}{max width=\linewidth}#1\end{adjustbox}}
\usepackage{pgfplots}
\pgfplotsset{compat=1.18}
\usepgfplotslibrary{groupplots,fillbetween}
\usepackage{pgfplotstable}
\pgfplotstableset{col sep=comma}
\usepackage{newunicodechar}
\newunicodechar{ℝ}{\ensuremath{\mathbb{R}}}
\newunicodechar{ι}{\ensuremath{\iota}}
\newunicodechar{α}{\ensuremath{\alpha}}
\newunicodechar{∀}{\ensuremath{\forall}}
\newunicodechar{∈}{\ensuremath{\in}}
\newunicodechar{∑}{\ensuremath{\sum}}
\newunicodechar{≤}{\ensuremath{\le}}
\newunicodechar{⟪}{\ensuremath{\langle\!\langle}}
\newunicodechar{⟫}{\ensuremath{\rangle\!\rangle}}
\newunicodechar{→}{\ensuremath{\to}}
\newunicodechar{•}{\ensuremath{\bullet}}
\newunicodechar{‖}{\ensuremath{\|}}
\newunicodechar{⊥}{\ensuremath{\perp}}
\newunicodechar{∘}{\ensuremath{\circ}}
\newcommand{\E}{\mathbb{E}}
\newcommand{\R}{\mathbb{R}}
\newcommand{\tr}{\operatorname{tr}}
\definecolor{series1}{HTML}{2a78d6}
\definecolor{series2}{HTML}{eb6834}
\definecolor{series3}{HTML}{1baf7a}
\definecolor{series1dark}{HTML}{1c5cab}
\definecolor{series1light}{HTML}{86b6ef}
\definecolor{inkmuted}{HTML}{898781}
\definecolor{inksecond}{HTML}{52514e}
\definecolor{gridgray}{HTML}{e1e0d9}
\definecolor{axisgray}{HTML}{c3c2b7}
\pgfplotsset{
  paper/.style={grid=major, grid style={line width=0.3pt, draw=gridgray}, axis line style={draw=axisgray, line width=0.4pt},
    tick style={draw=axisgray, line width=0.4pt}, major tick length=2pt, tick label style={font=\scriptsize, text=inksecond},
    label style={font=\footnotesize}, title style={font=\footnotesize, yshift=-2pt}, legend style={draw=none, fill=none, font=\scriptsize},
    legend cell align=left, every axis plot/.append style={line width=0.9pt}, scaled ticks=false,
    xticklabel style={/pgf/number format/fixed, /pgf/number format/precision=2}, yticklabel style={/pgf/number format/fixed, /pgf/number format/precision=2}},
  zero x/.style={extra x ticks={0}, extra x tick labels={}, extra x tick style={grid style={draw=black!50, line width=0.45pt}, major tick length=0pt}},
  zero y/.style={extra y ticks={0}, extra y tick labels={}, extra y tick style={grid style={draw=black!50, line width=0.45pt}, major tick length=0pt}},
  ringed/.style={mark options={draw=white, line width=0.35pt, fill=#1}, color=#1},
  dot/.style={only marks, mark=*, mark size=1.9pt, ringed=#1},
  diamond/.style={only marks, mark=diamond*, mark size=2.4pt, ringed=#1},
  tri/.style={only marks, mark=triangle*, mark size=2.2pt, ringed=#1},
  sq/.style={only marks, mark=square*, mark size=1.7pt, ringed=#1},
  tick mark/.style={only marks, mark=|, mark size=4pt, mark options={line width=1pt, draw=#1}, color=#1},
  hbar/.style={quiver={u=\thisrow{#1}, v=0}, line cap=butt},
  modelrows/.style 2 args={ytick={#2}, yticklabels from table={#1}{name}, y tick label style={font=\scriptsize, text=black}, major y tick style={draw=none}, ymajorgrids=false},
}
\tikzset{group label/.style={font=\scriptsize\itshape, text=inkmuted}}
\makeatletter\g@addto@macro\UrlBreaks{\do-}\makeatother

%% file: numbers_r1.tex
\newcommand{\ArmDetachLossMin}{0.040}

\newcommand{\ArmElasticityOne}{0.58}
\newcommand{\ArmElasticityZero}{0.62}

\newcommand{\ArmLamOneLoss}{3.441}

\newcommand{\ArmLamZeroLoss}{3.439}

\newcommand{\ArmLamZeroShare}{32}

\newcommand{\ArmStLossMin}{0.320}

\newcommand{\CatsFiveDLossMax}{0.35}
\newcommand{\CatsFiveDLossMin}{0.06}

\newcommand{\CatsFiveDampRatioMax}{0.89}
\newcommand{\CatsFiveDampRatioMin}{0.74}

\newcommand{\CatsFiveMassMax}{36}
\newcommand{\CatsFiveMassMin}{21}

\newcommand{\CatsFiveZeroedClosedMax}{85}
\newcommand{\CatsFiveZeroedClosedMin}{59}

\newcommand{\CatsNModels}{10}

\newcommand{\CcCatsCosDeltaMax}{0.030}
\newcommand{\CcCatsCosDeltaMin}{0.007}

\newcommand{\CcCatsDRatioMax}{0.89}
\newcommand{\CcCatsDRatioMin}{0.74}

\newcommand{\CcCtlCosDeltaAbsMax}{0.008}

\newcommand{\CcCtlDRatioMax}{1.02}
\newcommand{\CcCtlDRatioMin}{0.98}

\newcommand{\CcCtlPMax}{13.1}
\newcommand{\CcCtlPMin}{3.7}

\newcommand{\CcNCtlCosRises}{2}
\newcommand{\CcNModels}{10}
\newcommand{\CcNModelsWord}{ten}
\newcommand{\CfSwiMax}{0.022}
\newcommand{\CfSwiMin}{-0.023}
\newcommand{\CpAnnealEnd}{2000}
\newcommand{\CpAnnealEndM}{131}
\newcommand{\CpAnnealStart}{200}
\newcommand{\CpAnnealStartM}{13}
\newcommand{\CpBatchTokensK}{66}
\newcommand{\CpCRatio}{0.41}

\newcommand{\CpControlLossDrift}{0.038}
\newcommand{\CpControlValDrift}{0.004}
\newcommand{\CpEndAnnealA}{-0.056}

\newcommand{\CpEndAnnealC}{0.082}

\newcommand{\CpEndAnnealCos}{0.021}

\newcommand{\CpEndAnnealD}{-0.002}

\newcommand{\CpEndAnnealDbar}{0.020}

\newcommand{\CpEndAnnealLaw}{0.028}

\newcommand{\CpEndAnnealLoss}{2.888}

\newcommand{\CpEndAnnealValLoss}{2.873}

\newcommand{\CpEndAnnealWeak}{0.043}

\newcommand{\CpEndControlA}{-0.057}

\newcommand{\CpEndControlC}{0.202}

\newcommand{\CpEndControlCos}{-0.170}

\newcommand{\CpEndControlD}{0.320}

\newcommand{\CpEndControlDbar}{0.291}

\newcommand{\CpEndControlLaw}{-0.174}

\newcommand{\CpEndControlLoss}{2.723}

\newcommand{\CpEndControlValLoss}{2.702}

\newcommand{\CpEndControlWeak}{0.030}

\newcommand{\CpEndLossGap}{0.165}

\newcommand{\CpEndValGap}{0.171}

\newcommand{\CpLamZeroControlValLoss}{2.739}

\newcommand{\CpLamZeroValLoss}{3.049}

\newcommand{\CpLlamaThreeOneBCRatio}{0.44}

\newcommand{\CpLlamaThreeOneBControlLossDrift}{0.236}
\newcommand{\CpLlamaThreeOneBControlValDrift}{0.035}

\newcommand{\CpLlamaThreeOneBEndAnnealC}{0.108}

\newcommand{\CpLlamaThreeOneBEndAnnealCos}{0.148}

\newcommand{\CpLlamaThreeOneBEndAnnealD}{-0.079}

\newcommand{\CpLlamaThreeOneBEndAnnealDbar}{-0.030}

\newcommand{\CpLlamaThreeOneBEndAnnealLaw}{0.198}

\newcommand{\CpLlamaThreeOneBEndAnnealWeak}{0.015}

\newcommand{\CpLlamaThreeOneBEndControlC}{0.247}

\newcommand{\CpLlamaThreeOneBEndControlCos}{-0.181}

\newcommand{\CpLlamaThreeOneBEndControlD}{0.438}

\newcommand{\CpLlamaThreeOneBEndControlLaw}{-0.202}

\newcommand{\CpLlamaThreeOneBEndControlWeak}{0.010}

\newcommand{\CpLlamaThreeOneBEndLossGap}{0.341}

\newcommand{\CpLlamaThreeOneBEndValGap}{0.273}

\newcommand{\CpLlamaThreeOneBHours}{12}

\newcommand{\CpLlamaThreeOneBLamZeroValGap}{0.459}

\newcommand{\CpLlamaThreeOneBMidCos}{-0.154}

\newcommand{\CpLlamaThreeOneBMidLam}{0.67}

\newcommand{\CpLlamaThreeOneBStartC}{0.275}

\newcommand{\CpLlamaThreeOneBStartCos}{-0.222}
\newcommand{\CpLlamaThreeOneBStartD}{0.497}

\newcommand{\CpLlamaThreeOneBZeroCos}{0.151}

\newcommand{\CpLr}{5\times10^{-5}}

\newcommand{\CpMidControlCos}{-0.163}
\newcommand{\CpMidCos}{-0.143}

\newcommand{\CpMidLam}{0.67}

\newcommand{\CpQwenHalfBControlLossDrift}{0.038}

\newcommand{\CpQwenHalfBHours}{9}

\newcommand{\CpSeedRangeMax}{0.008}

\newcommand{\CpStartC}{0.208}

\newcommand{\CpStartCos}{-0.184}
\newcommand{\CpStartD}{0.336}

\newcommand{\CpSteps}{3800}
\newcommand{\CpTokensM}{249}

\newcommand{\CpZeroCos}{0.022}

\newcommand{\CpZeroStep}{2400}
\newcommand{\CpZeroTokensM}{118}

\newcommand{\CtlPrevAboveTopExcess}{20}
\newcommand{\CtlPrevAboveTopName}{Phi-3-mini}

\newcommand{\CtlPrevGemmaTwoB}{0.96}
\newcommand{\CtlPrevHiGemmaTwoB}{0.97}
\newcommand{\CtlPrevHiLlamaThreeEightB}{1.03}
\newcommand{\CtlPrevHiLlamaThreeOneB}{1.04}
\newcommand{\CtlPrevHiLlamaTwoSevenB}{1.07}
\newcommand{\CtlPrevHiMistralSevenB}{0.99}
\newcommand{\CtlPrevHiOLMoE}{1.11}
\newcommand{\CtlPrevHiOLMoOneB}{1.06}
\newcommand{\CtlPrevHiPhiThreeMini}{1.21}
\newcommand{\CtlPrevHiQwenHalfB}{1.09}
\newcommand{\CtlPrevHiTinyLlama}{1.06}
\newcommand{\CtlPrevLlamaThreeEightB}{1.03}
\newcommand{\CtlPrevLlamaThreeOneB}{1.04}
\newcommand{\CtlPrevLlamaTwoSevenB}{1.07}
\newcommand{\CtlPrevLoGemmaTwoB}{0.96}
\newcommand{\CtlPrevLoLlamaThreeEightB}{1.03}
\newcommand{\CtlPrevLoLlamaThreeOneB}{1.04}
\newcommand{\CtlPrevLoLlamaTwoSevenB}{1.06}
\newcommand{\CtlPrevLoMistralSevenB}{0.97}
\newcommand{\CtlPrevLoOLMoE}{1.08}
\newcommand{\CtlPrevLoOLMoOneB}{1.06}
\newcommand{\CtlPrevLoPhiThreeMini}{1.19}
\newcommand{\CtlPrevLoQwenHalfB}{1.08}
\newcommand{\CtlPrevLoTinyLlama}{1.05}
\newcommand{\CtlPrevMistralSevenB}{0.98}
\newcommand{\CtlPrevNAbove}{8}

\newcommand{\CtlPrevNMistralSevenB}{2}

\newcommand{\CtlPrevNWithinWord}{nine}
\newcommand{\CtlPrevOLMoE}{1.09}
\newcommand{\CtlPrevOLMoOneB}{1.06}

\newcommand{\CtlPrevPhiThreeMini}{1.20}
\newcommand{\CtlPrevQwenHalfB}{1.08}
\newcommand{\CtlPrevTinyLlama}{1.06}
\newcommand{\CtlPrevWithinMax}{9}

\newcommand{\CzLamZeroCos}{0.080}
\newcommand{\CzLamZeroD}{0.0271}

\newcommand{\CzLamZeroRw}{-0.081}
\newcommand{\CzLamZeroShareOfSwi}{48}
\newcommand{\CzLamZeroStr}{0.23}

\newcommand{\CzLamZeroVal}{2.986}
\newcommand{\CzLamZeroWeak}{0.0260}
\newcommand{\CzLossDiff}{0.005}

\newcommand{\CzSwiCos}{0.096}
\newcommand{\CzSwiD}{0.0562}
\newcommand{\CzSwiLobe}{0.0489}

\newcommand{\CzSwiProbeCosLast}{0.079}
\newcommand{\CzSwiProbeCosMax}{0.203}
\newcommand{\CzSwiProbeGainLast}{0.056}

\newcommand{\CzSwiRw}{0.134}
\newcommand{\CzSwiStr}{0.78}

\newcommand{\CzSwiVal}{2.992}

\newcommand{\DampKeepProSparse}{11}
\newcommand{\DampKeepReluLlama}{11}

\newcommand{\EigAmpPrevMax}{40}
\newcommand{\EigAmpPrevMin}{22}
\newcommand{\EigAmpRandMax}{39}
\newcommand{\EigAmpRandMin}{22}

\newcommand{\EigDimMinusOneLlamaThreeEightB}{4095}

\newcommand{\EigDimMinusOneQwenHalfB}{895}

\newcommand{\EigFullPrevMax}{32}
\newcommand{\EigFullPrevMin}{9}
\newcommand{\EigFullRandMax}{27}
\newcommand{\EigFullRandMin}{5}
\newcommand{\EigLobeQGapPctMax}{14}
\newcommand{\EigMidPrevMax}{45}
\newcommand{\EigMidPrevMin}{38}
\newcommand{\EigMidRandMax}{49}
\newcommand{\EigMidRandMin}{42}

\newcommand{\EigOtherPRFracMax}{0.33}
\newcommand{\EigOtherPRFracMin}{0.31}

\newcommand{\EigPrevPRFracMax}{0.30}
\newcommand{\EigPrevPRFracMin}{0.24}

\newcommand{\EigRandPRFracMax}{0.34}
\newcommand{\EigRandPRFracMin}{0.33}

\newcommand{\EigRestQAbsMax}{0.109}
\newcommand{\EigTailsPrevMax}{17}
\newcommand{\EigTailsPrevMin}{9}
\newcommand{\EigTailsRandMax}{11}
\newcommand{\EigTailsRandMin}{3}
\newcommand{\EigTopFiftyPrevMax}{33}
\newcommand{\EigTopFiftyPrevMin}{12}
\newcommand{\EigTopFiftyRandMax}{30}
\newcommand{\EigTopFiftyRandMin}{10}

\newcommand{\FamDampGatedMaxPct}{46}
\newcommand{\FamDampGatedMinPct}{15}
\newcommand{\FamLobeGatedMaxPct}{47}
\newcommand{\FamLobeGatedMinPct}{10}
\newcommand{\FigGYTicks}{9,8,7,6,5,4,3,2,1,0}

\newcommand{\FigRatioHalfMax}{0.005}
\newcommand{\FigSepCG}{7.05}
\newcommand{\FigSepNC}{4.35}
\newcommand{\FigYMax}{17.00}
\newcommand{\FigYTicks}{16.40,15.40,14.40,13.40,12.40,11.40,10.40,9.40,8.40,7.40,5.70,4.70,3.00,2.00,1.00,0.00}
\newcommand{\GoA}{0.063}

\newcommand{\GoC}{0.159}
\newcommand{\GoCRaw}{0.670}
\newcommand{\GoCf}{-0.003}
\newcommand{\GoClosed}{0.74}
\newcommand{\GoCos}{0.187}
\newcommand{\GoCosRaw}{0.341}
\newcommand{\GoD}{-0.013}
\newcommand{\GoDHi}{-0.010}
\newcommand{\GoDLo}{-0.015}
\newcommand{\GoDRaw}{0.113}
\newcommand{\GoDbar}{0.063}
\newcommand{\GoDbarAnalytic}{0.063}
\newcommand{\GoDprev}{0.125}
\newcommand{\GoDprevRatio}{2.0}

\newcommand{\GoLobe}{0.065}

\newcommand{\GoLoss}{3.789}
\newcommand{\GoNPairs}{23}

\newcommand{\GoRW}{0.128}

\newcommand{\GoStrSharePct}{78}
\newcommand{\GoTpWikiRaw}{0.297}

\newcommand{\GoXnormMax}{2493}
\newcommand{\GoXnormMin}{292}

\newcommand{\GrClosedWeakAbsMax}{0.013}

\newcommand{\GrDClosedStrShareLlamaThreeEightBBase}{105}

\newcommand{\GrDClosedStrShareOLMoOneB}{71}

\newcommand{\GrFirstAlign}{0.136}

\newcommand{\GrFirstLobe}{0.0129}

\newcommand{\GrFirstTok}{0.12}
\newcommand{\GrHalfWidth}{0.005}
\newcommand{\GrLastAlign}{0.196}

\newcommand{\GrLastLobe}{0.0380}

\newcommand{\GrLastTok}{1.00}
\newcommand{\GrLawGapMax}{10^{-4}}
\newcommand{\GrLocalGapMax}{0.031}

\newcommand{\GrMidAlign}{0.196}

\newcommand{\GrMidTok}{0.50}

\newcommand{\GrOpenStrMax}{-0.012}
\newcommand{\GrOpenStrMin}{-0.069}
\newcommand{\GrOpenWeakMax}{0.115}
\newcommand{\GrOpenWeakMin}{0.015}

\newcommand{\GrShareMax}{140}
\newcommand{\GrShareMin}{71}

\newcommand{\LagANegMax}{-0.007}
\newcommand{\LagANegMin}{-0.087}
\newcommand{\LagANegWord}{five}
\newcommand{\LagCRatioMax}{1.35}
\newcommand{\LagCRatioMin}{0.94}
\newcommand{\LagCosThreeMax}{0.037}
\newcommand{\LagCosThreeMin}{-0.047}
\newcommand{\LagCosTwoMax}{0.063}
\newcommand{\LagCosTwoMin}{-0.062}
\newcommand{\LagDRatioMax}{0.95}
\newcommand{\LagDRatioMin}{0.90}
\newcommand{\LagDeltaAMax}{0.091}
\newcommand{\LagDeltaAMin}{0.010}
\newcommand{\LagNCosThreeNeg}{4}
\newcommand{\LagNCosTwoNeg}{5}
\newcommand{\LagQMax}{0.042}
\newcommand{\LagQMin}{-0.243}
\newcommand{\LagRintMax}{0.099}
\newcommand{\LagRintMin}{0.005}

\newcommand{\LawNullRD}{-0.90}

\newcommand{\LawPosCosOLMoE}{0.031}
\newcommand{\LawPosDOLMoE}{0.118}
\newcommand{\LawPosLawOLMoE}{0.076}

\newcommand{\LawRCDGated}{0.81}
\newcommand{\LawRCDGatedP}{4\times10^{-3}}

\newcommand{\LawSdAC}{0.069}

\newcommand{\LawSdD}{0.146}

\newcommand{\LawSdRatio}{2.1}
\newcommand{\LawStrongCosMax}{-0.127}
\newcommand{\LawStrongCosMin}{-0.222}
\newcommand{\LawStrongMax}{0.244}
\newcommand{\LawStrongMin}{0.135}
\newcommand{\LawStrongNames}{Llama-3.2-1B, Qwen2.5-0.5B, Llama-3.1-8B-Instruct, Gemma-2B, OLMo-1B}
\newcommand{\LawWeakCosMax}{-0.029}
\newcommand{\LawWeakCosMin}{-0.052}
\newcommand{\LawWeakMax}{0.048}
\newcommand{\LawWeakMin}{0.010}
\newcommand{\LawWeakNames}{TinyLlama-1.1B, Llama-2-7B, Mistral-7B, Phi-3-mini}

\newcommand{\LjNoTlcmOtherWmeanLlamaThreeEightB}{-0.131}

\newcommand{\LjNoTlcmPrevWmeanLlamaThreeEightB}{-0.125}

\newcommand{\LjNoTlcmRandWmeanLlamaThreeEightB}{-0.131}

\newcommand{\LjTlcmAttnQuadAbsMax}{0.131}

\newcommand{\LjTlcmMlpQuadMax}{-0.170}
\newcommand{\LjTlcmMlpQuadMin}{-0.425}

\newcommand{\LjTlcmMlpQuadPrevLlamaThreeEightB}{-0.303}

\newcommand{\LjTlcmNModels}{11}
\newcommand{\LjTlcmNModelsWord}{eleven}

\newcommand{\LjTlcmNPrevWmeanBelowOther}{3}
\newcommand{\LjTlcmNPrevWmeanBelowRand}{11}

\newcommand{\LjTlcmOtherTopAboveLlamaThreeEightB}{16}

\newcommand{\LjTlcmOtherTopBelowLlamaThreeEightB}{41}

\newcommand{\LjTlcmOtherWmeanLlamaThreeEightB}{-0.350}

\newcommand{\LjTlcmOtherWmeanMax}{-0.245}
\newcommand{\LjTlcmOtherWmeanMin}{-0.507}

\newcommand{\LjTlcmPrevOverOtherPctMax}{+6}
\newcommand{\LjTlcmPrevOverOtherPctMin}{-21}
\newcommand{\LjTlcmPrevOverRandPctMax}{44}
\newcommand{\LjTlcmPrevOverRandPctMin}{1}

\newcommand{\LjTlcmPrevTopAboveLlamaThreeEightB}{29}

\newcommand{\LjTlcmPrevTopBelowLlamaThreeEightB}{37}

\newcommand{\LjTlcmPrevWmeanLlamaThreeEightB}{-0.339}

\newcommand{\LjTlcmPrevWmeanMax}{-0.229}
\newcommand{\LjTlcmPrevWmeanMin}{-0.470}

\newcommand{\LjTlcmRandTopAboveLlamaThreeEightB}{8}

\newcommand{\LjTlcmRandTopBelowLlamaThreeEightB}{18}

\newcommand{\LjTlcmRandWmeanLlamaThreeEightB}{-0.306}

\newcommand{\LjTlcmRandWmeanMax}{-0.190}
\newcommand{\LjTlcmRandWmeanMin}{-0.446}

\newcommand{\LoDampAbsMax}{0.052}

\newcommand{\LoLossMax}{13.3}
\newcommand{\LoLossMin}{3.9}

\newcommand{\LoNModels}{10}

\newcommand{\NsFCosGapMax}{0.001}

\newcommand{\NsFCosHwMax}{0.003}

\newcommand{\NsFCosHwSixteenMax}{0.005}

\newcommand{\NsFDGapMax}{0.002}

\newcommand{\NsFDbarGapMax}{0.002}

\newcommand{\NsFLawGapMax}{0.001}

\newcommand{\NsFNInsideSixteenInterval}{60}

\newcommand{\NsFNModelsWord}{fifteen}
\newcommand{\NsFNQuantities}{60}

\newcommand{\OcDampRatioMax}{1.03}
\newcommand{\OcDampRatioMin}{0.63}

\newcommand{\OcLossMax}{10.0}
\newcommand{\OcLossMin}{4.4}

\newcommand{\OcNCosNeg}{10}

\newcommand{\OdNAllFiveBWord}{thirteen}

\newcommand{\OdNModelsWord}{fifteen}
\newcommand{\OdOverlapBMin}{4}
\newcommand{\OdOverlapWikiMin}{3}
\newcommand{\OeAAbsMax}{0.063}

\newcommand{\OeCMax}{0.219}
\newcommand{\OeCMin}{0.176}

\newcommand{\OeCosFirst}{0.193}
\newcommand{\OeCosLast}{0.031}

\newcommand{\OeDwFirst}{-0.033}
\newcommand{\OeDwLast}{0.118}

\newcommand{\OeLawFirst}{0.272}
\newcommand{\OeLawLast}{0.076}

\newcommand{\OeLobeFirst}{0.002}
\newcommand{\OeLobeLast}{0.125}

\newcommand{\OeN}{10}

\newcommand{\OeRWFirst}{-0.024}
\newcommand{\OeRWLast}{0.199}

\newcommand{\OlAAbsMax}{0.031}

\newcommand{\OlALast}{-0.031}

\newcommand{\OlCLast}{0.175}
\newcommand{\OlCMax}{0.175}
\newcommand{\OlCMin}{0.120}

\newcommand{\OlCosLast}{-0.156}

\newcommand{\OlCrossDbar}{0.112}
\newcommand{\OlCrossLobe}{0.008}

\newcommand{\OlCrossTok}{64}
\newcommand{\OlCrossWeak}{0.109}

\newcommand{\OlDwFirst}{0.004}
\newcommand{\OlDwLast}{0.304}

\newcommand{\OlLawCrossC}{0.122}
\newcommand{\OlLawCrossD}{0.156}
\newcommand{\OlLawCrossTok}{129}

\newcommand{\OlLawPreC}{0.126}
\newcommand{\OlLawPreCrossTok}{64}
\newcommand{\OlLawPreD}{0.122}

\newcommand{\OlLobeLast}{0.165}

\newcommand{\OlN}{15}
\newcommand{\OlPreCrossTok}{33}

\newcommand{\OlRWLast}{0.038}

\newcommand{\OlRWMinVal}{-0.301}

\newcommand{\OlWeakLast}{0.074}

\newcommand{\PairCosProSparse}{0.096}

\newcommand{\PairCosReluLlama}{0.113}

\newcommand{\PairDampProSparse}{-0.130}
\newcommand{\PairDampProSparseHi}{-0.127}
\newcommand{\PairDampProSparseLo}{-0.132}

\newcommand{\PairDampReluLlama}{-0.129}
\newcommand{\PairDampReluLlamaHi}{-0.127}
\newcommand{\PairDampReluLlamaLo}{-0.132}

\newcommand{\PerpDRatioMax}{0.99}
\newcommand{\PerpDRatioMin}{0.81}

\newcommand{\PrevRatioAboveMax}{20}
\newcommand{\PrevRatioAboveMin}{8}

\newcommand{\PrevRatioExcessPythia}{31}
\newcommand{\PrevRatioGemmaHi}{8}
\newcommand{\PrevRatioGemmaLo}{7}
\newcommand{\PrevRatioHiMax}{22}
\newcommand{\PrevRatioLoMin}{8}

\newcommand{\PrevRatioNAbove}{9}

\newcommand{\PsLateCtlDAbsMax}{0.069}

\newcommand{\PtLobeNegDocsMin}{16}

\newcommand{\PtLobePartialMax}{-0.08}
\newcommand{\PtLobePartialMin}{-0.52}

\newcommand{\PtLobeTokMax}{-0.07}
\newcommand{\PtLobeTokMin}{-0.49}

\newcommand{\PtTotalPartialMax}{-0.11}
\newcommand{\PtTotalPartialMin}{-0.54}

\newcommand{\PtWeakPartialMax}{0.12}
\newcommand{\PtWeakPartialMin}{-0.29}

\newcommand{\PyAMax}{0.062}

\newcommand{\PyCMax}{0.160}
\newcommand{\PyCMin}{0.115}

\newcommand{\PyDwFirst}{0.023}
\newcommand{\PyDwLast}{0.361}

\newcommand{\PyFlipAfter}{8{,}000}
\newcommand{\PyFlipBefore}{4{,}000}

\newcommand{\RClosedMax}{0.83}
\newcommand{\RClosedMin}{0.62}

\newcommand{\RClosedSideShareGemmaTwoB}{90}
\newcommand{\RClosedSideShareOLMoOneB}{66}
\newcommand{\RClosedSideSharePhiThreeMini}{77}

\newcommand{\RCosLlamaThreeEightB}{-0.127}
\newcommand{\RCosLlamaThreeEightBBase}{-0.130}

\newcommand{\RCosLlamaTwoSevenB}{-0.041}

\newcommand{\RCosMeanRemPhiThreeMini}{-0.047}
\newcommand{\RCosMeanRemProSparse}{0.058}

\newcommand{\RCosPhiThreeMini}{-0.052}

\newcommand{\RCosRawGemmaTwoB}{-0.015}

\newcommand{\RCosRawPhiThreeMini}{0.083}
\newcommand{\RCosRawProSparse}{-0.218}

\newcommand{\RCosStdPhiThreeMini}{-0.067}

\newcommand{\RCosTopFivePhiThreeMini}{0.020}

\newcommand{\RCosTopOneGemmaTwoB}{-0.121}

\newcommand{\RCosTopOnePhiThreeMini}{0.059}

\newcommand{\RCosTopTwentyPhiThreeMini}{-0.053}
\newcommand{\RCosTopTwentyProSparse}{-0.255}

\newcommand{\RCosTopTwoPhiThreeMini}{0.030}

\newcommand{\RFamHalfWidth}{0.006}

\newcommand{\RLobeShareMin}{59}

\newcommand{\RLossTwoLlamaTwoSevenB}{2.02}

\newcommand{\RLossTwoProSparse}{2.15}

\newcommand{\RLossTwoReluLlama}{2.10}

\newcommand{\RRWGPTTwo}{-0.096}
\newcommand{\RRWGPTTwoMedium}{-0.102}

\newcommand{\RRWLlamaTwoSevenB}{0.321}

\newcommand{\RRWProSparse}{0.269}
\newcommand{\RRWPythiaSmall}{-0.362}

\newcommand{\RRWReluLlama}{0.299}

\newcommand{\RRadialAbsMax}{0.0001}

\newcommand{\RRouterAbsMaxOLMoE}{0.0005}
\newcommand{\RRouterOLMoE}{0.0002}
\newcommand{\RRouterShareOLMoE}{0.2}
\newcommand{\RTraceMaxGap}{0.037}

\newcommand{\RTraceMedGap}{0.0002}
\newcommand{\RTraceNLayers}{376}

\newcommand{\SelFlipMax}{13}
\newcommand{\SelFlipMin}{9}

\newcommand{\SelFlipWMax}{6.0}
\newcommand{\SelFlipWMin}{3.4}

\newcommand{\SelLagTwoAbsMax}{8}

\newcommand{\SelNModelsOtherZeroWord}{eight}

\newcommand{\SelNPrevAboveOtherMax}{3}

\newcommand{\SelOdocRandMax}{26}
\newcommand{\SelOdocRandMin}{5}

\newcommand{\SelOtherRandMax}{22}
\newcommand{\SelOtherRandMin}{4}

\newcommand{\SelRatioPrevAttnMax}{1.06}
\newcommand{\SelRatioPrevAttnMin}{0.84}

\newcommand{\SelRatioPrevOdocMax}{0.96}
\newcommand{\SelRatioPrevOdocMin}{0.83}

\newcommand{\SelRatioPrevOtherLlamaThreeEightBBase}{0.91}

\newcommand{\SelRatioPrevOtherMax}{0.99}
\newcommand{\SelRatioPrevOtherMin}{0.88}

\newcommand{\SelRatioPrevRandLlamaThreeEightBBase}{1.04}

\newcommand{\SelShufRandAbsMax}{0.4}

\newcommand{\SpAOneGPTTwo}{0.140}

\newcommand{\SpAlayerOneGPTTwo}{0.131}

\newcommand{\SpCtlALlamaThreeEightBBase}{-0.008}
\newcommand{\SpCtlCLlamaThreeEightBBase}{0.146}
\newcommand{\SpCtlDLlamaThreeEightBBase}{0.276}

\newcommand{\SpDOneProSparse}{-0.011}
\newcommand{\SpDOnePythiaSmall}{0.361}

\newcommand{\SpDOneReluLlama}{-0.006}

\newcommand{\SpecContractMax}{0.78}
\newcommand{\SpecContractMin}{0.62}

\newcommand{\SpecLobeContractMax}{0.85}
\newcommand{\SpecLobeContractMin}{0.61}

\newcommand{\SpecLobeMeanMax}{0.458}
\newcommand{\SpecLobeMeanMin}{0.092}

\newcommand{\SpecQFiveMax}{-0.33}
\newcommand{\SpecQFiveMin}{-0.47}

\newcommand{\SpecQNinetyFiveMax}{1.31}
\newcommand{\SpecQNinetyFiveMin}{0.77}

\newcommand{\SpecRestAbsMaxSwi}{0.027}

\newcommand{\SpecRestMeanGemmaTwoB}{0.046}

\newcommand{\SpecRestMeanPhiThreeMini}{0.047}

\newcommand{\StrShareMax}{91}
\newcommand{\StrShareMin}{53}

\newcommand{\SwIdentityGap}{1.3\times10^{-3}}
\newcommand{\SwLayerRatioMax}{1.05}
\newcommand{\SwLayerRatioMin}{0.91}
\newcommand{\SwLinErrMax}{0.22}
\newcommand{\SwLinErrMed}{0.13}

\newcommand{\TlAMax}{0.047}

\newcommand{\TlCMax}{0.164}
\newcommand{\TlCMin}{0.114}

\newcommand{\TlDwFirst}{-0.016}

\newcommand{\TlDwLast}{0.145}

\newcommand{\TlDwStepFlip}{0.065}
\newcommand{\TlDwStepMaxOther}{0.025}
\newcommand{\TlFlipAfter}{3{,}000}
\newcommand{\TlFlipBefore}{2{,}500}

\newcommand{\TlLobeA}{0.040}

\newcommand{\TlLobeF}{0.153}
\newcommand{\TlLobeFirst}{0.040}

\newcommand{\TlLobeLast}{0.171}

\newcommand{\TlLossA}{2.72}
\newcommand{\TlLossB}{3.30}
\newcommand{\TlLossC}{2.56}

\newcommand{\TlRWA}{0.170}

\newcommand{\TlRWFirst}{0.170}

\newcommand{\TlRWLast}{0.362}

\newcommand{\TlWeakA}{0.009}

\newcommand{\TlWeakF}{0.007}

\newcommand{\TpCorpusGapMax}{0.055}

\newcommand{\TpMidLessNegWord}{six}
\newcommand{\TpNArticles}{40}

\newcommand{\TpPosFirstLlamaThreeEightBBase}{-0.133}

\newcommand{\TpPosGapMax}{0.037}

\newcommand{\TpPosLastLlamaThreeEightBBase}{-0.141}

\newcommand{\TpWikiMedianLlamaThreeEightBBase}{-0.150}

\newcommand{\TpWikiModeLlamaThreeEightBBase}{-0.170}

\newcommand{\TpWikiRawLlamaThreeEightB}{-0.129}
\newcommand{\TpWikiRawLlamaThreeEightBBase}{-0.139}

\newcommand{\TpWindowNegWord}{ten}
\newcommand{\TpxAllPairsLlamaThreeEightB}{-0.025}
\newcommand{\TpxAllPairsLlamaThreeEightBBase}{-0.034}
\newcommand{\TpxDocsLlamaThreeEightB}{2{,}001}

\newcommand{\TpxMinPairLlamaThreeEightB}{-0.212}
\newcommand{\TpxMinPairLlamaThreeEightBBase}{-0.229}

\newcommand{\TpxNPairsBelowPointFifteenLlamaThreeEightB}{6}
\newcommand{\TpxNPairsBelowPointFifteenLlamaThreeEightBBase}{9}

\newcommand{\TpxStrongPairIdxLlamaThreeEightB}{11}

\newcommand{\TpxTokensMLlamaThreeEightB}{1.02}

\newcommand{\TpxWindowLlamaThreeEightB}{-0.123}
\newcommand{\TpxWindowLlamaThreeEightBBase}{-0.133}

\newcommand{\WinAllPairsPosNames}{TinyLlama-1.1B, Llama-2-7B, Mistral-7B, Phi-3-mini}

\newcommand{\WinFracNegAllGatedMax}{0.87}
\newcommand{\WinFracNegAllGatedMin}{0.35}

\newcommand{\WinFracNegWinGatedMax}{1.00}
\newcommand{\WinFracNegWinGatedMin}{0.69}

\newcommand{\WkACGapMax}{0.032}

\newcommand{\WkDGapMax}{0.013}

\newcommand{\WkLawRank}{0.99}

\newcommand{\WkLongDRatioMax}{1.17}
\newcommand{\WkLongDRatioMin}{1.01}

\newcommand{\WkLongLawMax}{0.222}
\newcommand{\WkLongLawMin}{-0.255}

\newcommand{\WkLongNGatedWord}{ten}

\newcommand{\WkLongNModelsWord}{fourteen}
\newcommand{\WkLongNSignAgree}{14}

\newcommand{\WkNModelsWord}{sixteen}

\newcommand{\WkNSignAgreeWord}{sixteen}
\newcommand{\WkSdRatio}{2.2}
\newcommand{\WkSignDisagreeNames}{none}

%% file: abstract.tex
In transformer language models with gated multilayer perceptrons (MLPs), adjacent layers write partly against each other into the residual stream. Previous work interpreted this cancellation as layers proposing features and the next layer selectively rejecting the inappropriate ones. We decompose the cancellation by removing the previous layer's write from the input of the next MLP: what the next layer writes along that write splits into what it writes anyway and how much less its MLP writes because the write is present, which we call the damping, and we trace the damping to the neurons that carry it. In gated MLPs the damping is carried by neurons whose gates are closed. SiLU and GELU are slightly negative below zero, so a closed gate still passes a small negative multiple of what the neuron reads, and a neuron whose output weight points along its input weight then subtracts a fraction of the direction it read. Summed over such neurons, closed gates contract the average direction a gated MLP reads, and the previous layer's write is not singled out: the same layer's write at another token is damped as much or more, and the eigenvalue signature from which that work concluded that the correction is selective appears for another token's write as well. Over training the damping grows while what the next layer writes anyway stays flat, and cancellation begins when the damping overtakes it; annealing the negative side of the gate activation away in continued pretraining removes the damping and the cancellation. Adjacent-layer cancellation is therefore weak evidence about which features a model rejects: it is produced by a contraction that gated MLPs apply, on average over directions, to whatever they read.

%% file: intro.tex
\section{Introduction}
\label{sec:intro}

\citet{elhage2021framework} describe a transformer's residual stream as a communication channel: each layer reads from the stream and writes back by adding to it, so what one layer writes stays unless another component writes against it. \citet{patrawala2025tlcm} found that adjacent layers do write against each other. Averaged over tokens, the cosine between what layer $i$ adds to the stream and what layer $i+1$ adds is $\approx -0.2$ in five of seven model families (Llama~3, OLMo, Mistral, Gemma and Qwen2), most strongly in the first two-thirds of the network, while layers further apart do not cancel on average. The effect is absent at initialisation and grows during pretraining, and they find the multilayer perceptrons (MLPs) primarily responsible. They interpret the cancellation through a \emph{propose-and-reject} framework, in which layers propose candidate features and subsequent layers selectively correct the inappropriate ones, a hypothesis they state they do not test; from the eigenvalues of each layer's Jacobian they conclude that the correction is selective rather than a uniform attenuation. If that interpretation holds, adjacent-layer cancellation shows which features a model rejects, and the components that write against the previous layer are where the rejection is computed.

We ask what in the next layer writes against the previous one, and whether it singles out what that layer wrote. Removing the previous layer's write from the next MLP's input splits the next layer's output along that write into three parts: what its attention writes along it, what its MLP writes along it with the write absent (\emph{co-writing}), and how much less the MLP writes along it because the write is present (\emph{damping}). By definition the pair cancels when the damping exceeds the other two. The content is in the sizes of the parts, in which neurons carry the damping, and in whether the write is damped differently from other directions.

\textbf{We find that the next MLP damps the previous layer's write less than it damps the same layer's write at another token; that in gated models this damping is carried by neurons whose gates are closed, where SiLU and GELU are slightly negative; and that adjacent layers cancel where this contraction outweighs what the next layer writes along the previous write anyway.} A gated MLP \citep{shazeer2020glu} computes $m(x) = W_{\mathrm{down}}\big(\phi(W_{\mathrm{gate}} z)\odot W_{\mathrm{up}} z\big)$ with $z$ the normalised stream and $\phi$ SiLU or GELU. Both are slightly negative below zero, and \citet{gerstner2025gated} describe how that reverses what a neuron does: a closed gate still passes a small negative multiple of the neuron's input (it \emph{leaks}), so a neuron whose output weight points along its input weight adds the direction it reads when its gate is open and subtracts a small multiple of that direction when the gate is closed. Between \StrShareMin\% and \StrShareMax\% of the middle layers' neurons are of this \emph{strengthening} kind, and between \RClosedMin{} and \RClosedMax{} of the gates are closed at a token. Summed over the closed strengthening neurons, the reversal that prior work described one neuron at a time becomes a contraction of everything the MLP reads, and this contraction carries most or all of the damping of the previous write in all ten gated models we examined, a mixture of experts among them.

This limits what cancellation shows. In all ten gated models the previous write is damped less than the same layer's write at another token. How much more it is damped than a random direction depends on the frame: along the controlled write it is within \CtlPrevWithinMax\% of a random direction in nine of the ten (\CtlPrevAboveTopExcess\% above it in \CtlPrevAboveTopName), along the raw write \PrevRatioAboveMin\% to \PrevRatioAboveMax\% above it, and through the whole layer's Jacobian \LjTlcmPrevOverRandPctMin\% to \LjTlcmPrevOverRandPctMax\% above it. A direction written into the stream is, on average over directions, written against by the next gated MLP by \FamDampGatedMinPct\% to \FamDampGatedMaxPct\% of its size; single directions spread widely around that average, and a fifth to a third of them are amplified. \textbf{Cancellation is therefore weak evidence about which features a model rejects, and analyses that read a layer's contribution, or the next layer's response to it, should expect the next gated MLP to write against most contributions, not only against the ones a model rejects.}

The contraction depends on an architectural choice, not only on what training learns: 51M-parameter models trained with the gate's negative region removed have \ArmLamZeroShare\% of the damping, at a loss two seeds cannot distinguish, and two ReLU conversions of Llama-2-7B keep about a tenth of it. In a model that cancels, the cancellation goes with the negative region. Continued pretraining of Qwen2.5-0.5B with the negative region annealed away leaves a model whose window cosine has gone from $\CpStartCos$ to $+\CpEndAnnealCos$, at a loss cost of \CpEndLossGap{} nats, while the same training with the activation unchanged leaves it at $\CpEndControlCos$ (three seeds); Llama-3.2-1B behaves the same (one seed).

\paragraph{Contributions.}
\begin{itemize}\setlength{\itemsep}{2pt}\setlength{\parsep}{0pt}\setlength{\topsep}{2pt}
\item \textbf{A decomposition} (Section~\ref{sec:law}) of the next layer's output along the previous write into attention, co-writing and damping, by a stated counterfactual, in sixteen models on two corpora. The damping varies \LawSdRatio{} times as much across models as the other two together, and in released training runs it grows while co-writing stays flat.
\item \textbf{The source of the damping} (Section~\ref{sec:mechanism}): an exact split by neuron through the normaliser's Jacobian. Closed strengthening neurons carry most or all of the damping of the previous write in all ten gated models; models without a gate damp through active neurons whose output weights oppose their input weights.
\item \textbf{What the damping singles out} (Section~\ref{sec:mechanism}): nothing. Against a random direction, the same layer's writes at other tokens and in other documents, and over the eigendirections of the MLP's and the whole layer's Jacobians, the previous layer's write is not singled out.
\item \textbf{Interventions} (Section~\ref{sec:interventions}): annealing the negative region away in continued pretraining of two models that cancel removes the cancellation in both while the matched controls keep it; removing it in training from scratch, at inference, by thresholding and in public ReLU conversions lowers the damping.
\end{itemize}

%% file: setup.tex
\section{Setup}
\label{sec:setup}

\paragraph{Notation.} Write $x_l\in\R^d$ for the residual stream at a token before layer $l$ and $c_l = x_{l+1} - x_l$ for layer $l$'s contribution, the sum of its attention write $a_l$ and its MLP write $m_l$. In a sequential pre-norm block the MLP reads $x'_l = x_l + a_l$ through $z = N(x'_l)$, with $N$ an RMSNorm \citep{zhang2019rmsnorm} or LayerNorm \citep{ba2016layernorm}; in Pythia's parallel blocks \citep{biderman2023pythia} it reads $x_l$ itself. A gated MLP \citep{dauphin2017glu, shazeer2020glu} has neurons $k$ with gate weight $g_k$, linear input weight $u_k$ and output weight $d_k$, $m(x') = \sum_k \phi(g_k\cdot z)\,(u_k\cdot z)\,d_k$, with $\phi$ the SiLU $t\,\sigma(t)$ \citep{elfwing2018silu, ramachandran2017swish} (SwiGLU), the tanh-approximate GELU \citep{hendrycks2016gelu} (GeGLU) or ReLU (ReGLU). A neuron's gate is \emph{closed} at a token when $g_k\cdot z \le 0$. SiLU and GELU are negative there (minima $-0.278$ and $-0.170$) and ReLU is zero. We call a neuron \emph{strengthening} when $\langle w\odot d_k, u_k\rangle > 0$ and \emph{weakening} otherwise, with $w$ the normaliser's gain; this inner product, not the cosine, is what enters the Jacobian (Section~\ref{sec:mechanism}). Being a sign split weighted by the gain, it counts weakly aligned neurons that the $\pm0.5$ cosine thresholds of \citet{gerstner2025gated} (enrichment and depletion neurons) exclude; the read--write cosines we report are unweighted.

\paragraph{Cancellation.} We measure cancellation by the cosine between adjacent contributions, $\cos(c_{j-1}, c_j)$, averaged over tokens. \citet{patrawala2025tlcm} average the raw cosine. Because a few activations at the first token and some delimiters are orders of magnitude larger than the rest \citep{sun2024massive, yu2024superweight} and a few dimensions that are large at every token dominate cosine similarity \citep{timkey2021rogue}, our \emph{controlled} cosine drops position~0, zeroes the five dimensions with the largest mean absolute contribution over all layers and tokens, and subtracts each layer's mean contribution over tokens. That mean is a component every token shares, and whether it is removed sets the sign of the window cosine for Phi-3-mini and ProSparse-7B (Appendix~\ref{app:raw}). The controls were fixed before the measurements, and the five dimensions recur on other documents and on another corpus (Appendix~\ref{app:controls}). Means over layers use a \emph{window} of pairs $i\to i+1$ with $\tfrac{L}{8} \le i < \tfrac{5L}{8}$ for a model with $L$ layers, which generalises the window of \citet{patrawala2025tlcm}. On their corpus the raw window cosine differs from ours by at most \TpCorpusGapMax{} (Figure~\ref{fig:protocol}), and their released protocol gives $\TpxWindowLlamaThreeEightB$ over the window pairs of their main model, against the $\approx -0.2$ they report; its strongest pairs reach $\TpxMinPairLlamaThreeEightB$ (Appendix~\ref{app:controls}). The window is where they report the effect, not where it is largest: over all adjacent pairs the controlled cosine is negative in \WinFracNegAllGatedMin{} to \WinFracNegAllGatedMax{} of the pairs of the gated models that cancel in the window (in the window, \WinFracNegWinGatedMin{} to \WinFracNegWinGatedMax), and its all-pairs mean is positive in \WinAllPairsPosNames{} (Table~\ref{tab:cosctl}).

\paragraph{Damping.} For the MLP branch $m_l$ and its Jacobian $J_l$ at the token's stream, the damping is $\bar D_l = -\,\E_u \langle u, J_l\, u\rangle$ with $u$ uniform on the unit sphere orthogonal to $x'_l$; $\bar D_l > 0$ means the MLP contracts the average direction. We estimate it by finite differences with one random direction per token, $\|\epsilon u\| = 0.02\,\|x'_l\|$, and average over tokens. In expectation it equals $-(d-1)^{-1}\tr(P J_l P)$, with $P$ the projection orthogonal to $x'_l$: the mean of the damping over directions. Single directions spread widely around that mean (in every gated model the 5th percentile over directions is at most $\SpecQFiveMax$ and the 95th at least $+\SpecQNinetyFiveMin$; Appendix~\ref{sec:spectrum}), and the average damping is distinct from the damping $D$ of a particular write (Section~\ref{sec:law}). We also measure $D^{\mathrm{prev}}_l$, the local damping along the previous layer's write $c_{l-1}$ projected orthogonal to $x'_l$. The part of a write along $x'_l$ is invisible to the normaliser and is excluded from all three.

\paragraph{Models and data.} Ten gated models: Llama-3.2-1B \citep{meta2024llama32}, Llama-2-7B \citep{touvron2023llama2}, Llama-3.1-8B-Instruct \citep{llama3}, Qwen2.5-0.5B \citep{qwen25}, Mistral-7B \citep{jiang2023mistral}, Gemma-2B \citep{gemma}, Phi-3-mini \citep{phi3}, and three with their released training checkpoints, TinyLlama-1.1B \citep{zhang2024tinyllama}, OLMo-1B (July 2024 release; \citealp{groeneveld2024olmo,olmo1b0724}) and OLMoE-1B-7B \citep{muennighoff2024olmoe}, a mixture of experts with 64 SwiGLU experts of width 1{,}024 per layer and eight routed to each token (Appendix~\ref{app:decomp}). We also measure two ReLU conversions of Llama-2-7B, ReluLLaMA-7B \citep{relullama} and ProSparse-7B \citep{song2025prosparse}; four non-gated models, Pythia-160m with its checkpoints \citep{biderman2023pythia}, GPT-2 and GPT-2 medium \citep{radford2019gpt2} and OPT-1.3B \citep{zhang2022opt}; and the models we train (Section~\ref{sec:interventions}). Llama-3.2-1B is pruned and distilled from Llama 3.1 8B and 70B; Llama-3.1-8B-Instruct is the instruction-tuned release of the main model of \citet{patrawala2025tlcm} (the base release: Appendix~\ref{app:controls}); Phi-3-mini and ProSparse-7B also saw instruction data. Every measurement of a pretrained model uses the same sixteen FineWeb-Edu documents \citep{penedo2024fineweb} truncated to 192 tokens, each document one unit for intervals and tests; the 95\% half-width of every window average damping and cosine is at most \RFamHalfWidth{} (Table~\ref{tab:family}), and on 64 documents every window mean moves by at most \NsFDGapMax{} (Appendix~\ref{app:controls}). The 51M models are measured on sixteen validation sequences of 1{,}024 tokens. Everything ran on one RTX~5090, about 150 GPU-hours in all; Appendix~\ref{app:controls} specifies every measurement in full.

%% file: law.tex
\section{What the next layer writes along the previous write}
\label{sec:law}

\paragraph{A decomposition by counterfactual.} Take a layer $j$ and an earlier layer $b$ whose write $v = c_b$ we follow ($b = j-1$ for adjacent layers). Layer $j$'s contribution is its attention write plus its MLP write, so its projection on the unit vector $\hat v$ along $v$ is $\langle \hat v, a_j\rangle + \langle \hat v, m_j\rangle$. We also evaluate the MLP with $v$ removed from its input, $m_j(x'_j - v)$, with everything else unchanged, and define, in units of $\|v\|$, $A = \langle \hat v, a_j\rangle/\|v\|$, $C = \langle \hat v, m_j(x'_j - v)\rangle/\|v\|$ and $D = \langle \hat v, m_j(x'_j - v) - m_j(x'_j)\rangle/\|v\|$, so that
\begin{equation}
\frac{\langle \hat v, c_j\rangle}{\|v\|} \;=\; A + C - D
\label{eq:switch}
\end{equation}
holds at every token by construction. $C$, the \emph{co-writing}, is what the MLP writes along $v$ when $v$ is absent from its input; $D$, the \emph{damping}, is how much less it writes along $v$ because $v$ is present; $A$ is attention's projection on $v$, including attention's own response to $v$, which we do not separate. By definition, layer $j$'s contribution points against $v$ when $D > A + C$; what the decomposition adds is the counterfactual and the sizes of the terms. We report means over tokens with 95\% intervals over documents, in the frame of the controlled cosine: $v$ is the controlled $c_b$, and every projected vector has its mean over tokens, in the same condition, removed, so that the sign of the left side at each token is that of the controlled cosine (raw vectors give the raw cosine, Table~\ref{tab:switchraw}). Appendix~\ref{app:controls} checks that $D$ is a property of the MLP rather than of how the write is removed.
\begin{figure*}[t]
\centering
\input{figures/law}
\caption{Adjacent layers cancel when the next MLP's damping of the previous layer's write exceeds what the next layer writes along that write anyway. (a) Adjacent layers in sixteen models (window means, in units of the previous write's norm): what the next layer writes along the previous write anyway, $A + C$ (gray), and the damping $D$ (blue); the gap is the next layer's projection on the previous write, $A + C - D$, and the pair cancels when the damping is the larger (blue segment). Values with 95\% intervals in Table~\ref{tab:switchci}. (b) Over the released checkpoints of OLMo-1B, $A + C$ stays flat while the damping grows past it; in the shaded interval the cosine between adjacent writes, and then $A + C - D$, turns negative. (c) In the same run the average damping $\bar D$ comes first from open neurons whose output weights oppose their input weights and later from closed strengthening neurons (Figure~\ref{fig:sources}).}
\label{fig:law}
\end{figure*}

\paragraph{The damping is the term that separates the models.} In \LawStrongNames{} the damping exceeds attention and co-writing together by \LawStrongMin{} to \LawStrongMax{} and the controlled cosine is between $\LawStrongCosMin$ and $\LawStrongCosMax$; in \LawWeakNames{} it exceeds them by \LawWeakMin{} to \LawWeakMax{} and the cosine is between $\LawWeakCosMin$ and $\LawWeakCosMax$ (Figure~\ref{fig:law}a). OLMoE-1B-7B, the mixture of experts, does not cancel (cosine $+\LawPosCosOLMoE$): its damping of the previous write is the smallest of the gated models' ($D = \LawPosDOLMoE$) and falls \LawPosLawOLMoE{} short of attention and co-writing. gpt-oss-20b \citep{openai2025gptoss}, a 32-expert mixture measured as released in 4-bit, writes along the previous write in attention and MLP and does not damp it ($A = +\GoA$, $C = \GoC$, $D = \GoD$; cosine $+\GoCos$; Appendix~\ref{app:decomp}). The two ReLU conversions do not damp the previous write ($D = \SpDOneReluLlama$ and $\SpDOneProSparse$) and reinforce it. Without a gate, Pythia-160m damps the previous write by \SpDOnePythiaSmall, within the range of the cancelling gated models, and cancels; GPT-2 and GPT-2 medium damp too but write more along the previous write in attention and co-writing, so they reinforce it; OPT-1.3B does not damp.

Across the sixteen models the damping varies \LawSdRatio{} times as much as $A + C$ (standard deviations \LawSdD{} and \LawSdAC; a correlation of $D$ with $A + C - D$, which share the term $D$, would be near $\LawNullRD$ even if $D$ and $A + C$ were independent). Within the ten gated models co-writing and damping rise together ($r = \LawRCDGated$; permutation $p = \LawRCDGatedP$, over models that share lineages), so their difference, not the damping alone, orders those models. At lag~2 later MLPs still damp an older write (\LagDRatioMin{} to \LagDRatioMax{} of the lag-1 damping per unit), but the layer in between has removed part of it and attention writes along it, so the controlled cosine two apart lies between $\LagCosTwoMin$ and $+\LagCosTwoMax$ (negative in \LagNCosTwoNeg{} of the ten gated models; Appendix~\ref{app:lags}): the concentration in adjacent pairs that \citet{patrawala2025tlcm} report needs no selectivity of the damping. On sixteen WikiText-103 articles the terms change by at most \WkDGapMax{} ($D$) and \WkACGapMax{} ($A + C$) and the sign of $A + C - D$ is unchanged in every model (Table~\ref{tab:wiki}); at 2{,}048 tokens the gated models' damping is \WkLongDRatioMin{} to \WkLongDRatioMax{} of its 192-token value.

\paragraph{Damping grows during training, co-writing does not.} Over the released checkpoints of OLMo-1B from 2B to 3T tokens (Figure~\ref{fig:law}b), co-writing stays between \OlCMin{} and \OlCMax{} and attention within \OlAAbsMax{} of zero, while the damping of the previous write grows from $\OlDwFirst$ to $\OlDwLast$; the cosine turns negative between \OlPreCrossTok{}B and \OlCrossTok{}B tokens and $A + C - D$ between \OlLawPreCrossTok{}B and \OlLawCrossTok{}B. Pythia-160m, which has no gate, TinyLlama-1.1B and the mixture OLMoE-1B-7B show the same shape (Appendix~\ref{app:training}), the mixture without reaching the crossing by 5.1T tokens. Each is a single released run.

%% file: figures/law.tex
\begin{minipage}[b]{0.5\linewidth}\centering
\begin{tikzpicture}
\begin{axis}[paper, zero x, scale only axis, width=\dimexpr\linewidth-2.75cm\relax, height=4.5cm,
  xmin=-0.07, xmax=0.53, ymin=-0.6, ymax=\FigYMax, modelrows={figures/data/figlaw.csv}{\FigYTicks}, enlarge y limits=false,
  title={(a) the next layer along the previous write}, xlabel={per unit of the previous write}, xtick={0,0.1,0.2,0.3,0.4,0.5},
  extra y ticks={\FigSepCG, \FigSepNC}, extra y tick labels={}, extra y tick style={grid=major, grid style={draw=axisgray, line width=0.4pt}, major tick length=0pt},
  legend style={at={(0.98,0.555)}, anchor=south east, fill=white, legend columns=1}, clip=false]
\addplot[series1light, line width=1.6pt, hbar=segl, forget plot] table[x=segc, y=y] {figures/data/figlaw.csv};
\addplot[black!18, line width=1.6pt, hbar=segl, forget plot] table[x=segr, y=y] {figures/data/figlaw.csv};
\addplot[diamond=black!60] table[x=AC, y=y] {figures/data/figlaw.csv}; \addlegendentry{written anyway, $A + C$}
\addplot[dot=series1] table[x=D, y=y] {figures/data/figlaw.csv}; \addlegendentry{damping $D$}
\node[group label, anchor=east] at (axis cs:0.525,7.4) {gated};
\node[group label, anchor=east] at (axis cs:0.525,4.7) {ReLU conversions};
\node[group label, anchor=east] at (axis cs:0.525,0) {no gate};
\end{axis}
\end{tikzpicture}
\end{minipage}\hfill
\begin{minipage}[b]{0.48\linewidth}\centering
\begin{tikzpicture}
\begin{groupplot}[group style={group size=1 by 2, vertical sep=0.85cm, xticklabels at=edge bottom}, paper, scale only axis, width=\dimexpr\linewidth-2.3cm\relax, height=1.85cm, clip=false,
  xmode=log, xmin=1.5, xmax=3400, log basis x=10, every axis plot/.append style={line width=0.8pt, mark size=1.3pt},
  legend style={legend columns=1, fill=white, fill opacity=0.85, text opacity=1}]
\nextgroupplot[title={(b) over the training of OLMo-1B}, zero y, ymin=-0.2, ymax=0.33, ytick={-0.1,0,0.1,0.2,0.3}]
\fill[gridgray!70] (axis cs:33,-0.2) rectangle (axis cs:129,0.33);
\addplot[series1, mark=*] table[x=x, y=Dw] {figures/data/tabolmotrain.csv};
\addplot[black!60, mark=diamond*] table[x=x, y expr=\thisrow{A}+\thisrow{C}] {figures/data/tabolmotrain.csv};
\addplot[black, dashed, line width=0.7pt] table[x=x, y=cos] {figures/data/tabolmotrain.csv};
\node[anchor=west, font=\scriptsize] at (axis cs:3400,\OlDwLast) {$D$};
\node[anchor=west, font=\scriptsize] at (axis cs:3400,{\OlALast+\OlCLast}) {$A + C$};
\node[anchor=west, font=\scriptsize] at (axis cs:3400,\OlCosLast) {cosine};
\nextgroupplot[title={(c) sources of $\bar D$ over the same run}, ymin=0, ymax=0.32, ytick={0,0.1,0.2,0.3}, xlabel={training tokens (billions)}, legend style={at={(0.02,0.97)}, anchor=north west}]
\fill[gridgray!70] (axis cs:33,0) rectangle (axis cs:129,0.32);
\addplot[series1, mark=*] table[x=x, y=lobe] {figures/data/tabolmotrain.csv}; \addlegendentry{closed strengthening}
\addplot[series2, mark=diamond*] table[x=x, y=weak] {figures/data/tabolmotrain.csv}; \addlegendentry{open weakening}
\addplot[black!85, line width=0.7pt] table[x=x, y=D] {figures/data/tabolmotrain.csv}; \addlegendentry{$\bar D$}
\end{groupplot}
\end{tikzpicture}
\end{minipage}

%% file: mechanism.tex
\section{Closed gates leak}
\label{sec:mechanism}

\paragraph{Where a gated MLP's average damping $\bar D$ comes from.} With an RMSNorm, $z = \sqrt{d}\,\|x'\|^{-1}\, w\odot x'$, and the Jacobian of $z$ with respect to $x'$ is $\sqrt d\,\|x'\|^{-1}\mathrm{diag}(w)\,P$, where $P$ projects out the direction of $x'$ (we omit the norm's $\epsilon$). Differentiating $m$ gives the MLP's Jacobian $J = M P$ with $M = \sqrt d\,\|x'\|^{-1}\sum_k d_k\big[\phi'(g_k\cdot z)\,(u_k\cdot z)\,(w\odot g_k)^{\top} + \phi(g_k\cdot z)\,(w\odot u_k)^{\top}\big]$, and $\bar D$ is minus the trace of $J$ on the subspace orthogonal to $x'$, divided by $d-1$. Since $\tr(PJP) = \tr M - \hat x^{\top} M \hat x$,
\begin{equation}
\begin{aligned}
\bar D = -\kappa \sum\nolimits_k \big[&\phi'(t_k)(u_k\!\cdot\! z)\langle w\odot d_k, g_k\rangle \\
&+ \phi(t_k)\langle w\odot d_k, u_k\rangle\big] + \frac{\hat x^{\top} M \hat x}{d-1} ,
\end{aligned}
\label{eq:trace}
\end{equation}
exactly, with $\kappa = \sqrt d\,(d-1)^{-1}\|x'\|^{-1}$ and $t_k = g_k\cdot z$: a term through the gate, a term through the linear input, and the part of $M$'s trace along $x'$ itself (expanded in Appendix~\ref{app:decomp}). Over \RTraceNLayers{} layers of the sixteen models it differs from the finite-difference estimate of $\bar D$ by a median of \RTraceMedGap{} and at most \RTraceMaxGap, and the last term is negligible (at most \RRadialAbsMax). Split by gate state (closed, $g_k\cdot z \le 0$, or open) and by alignment, the sum has six terms (Table~\ref{tab:decomp}). The term that carries the damping is the one through the linear input of closed strengthening neurons, where $\phi(t_k) < 0$ for SiLU and GELU and $\langle w\odot d_k, u_k\rangle > 0$, so each such neuron adds to $\bar D$: with $v_k = w\odot u_k$ the direction it reads, its term maps $r$ to $\phi(t_k)\langle v_k, r\rangle d_k$, whose trace $\phi(t_k)\langle v_k, d_k\rangle$ is negative and which contracts $v_k$ itself. We call the sum over these neurons the \emph{lobe term}, after the negative lobe of SiLU and GELU; with ReLU gating it is exactly zero. Only the part of each output weight along $v_k$ enters the trace; the remainder spreads the damping across directions.
\begin{figure*}[t]
\centering
\input{figures/sources}
\caption{Where the damping comes from. (a) The average damping $\bar D$ of a random direction (tick) and two of its sources: closed strengthening neurons through their linear input (blue, the lobe term) and open neurons whose output weight opposes their input weight (orange); in models without a gate, active neurons with negative slope and aligned weights take the blue role. (b) The damping $D$ of the previous write split by neuron group (each neuron's group fixed at the model's actual input); the parts sum to the total (tick). (c) The local damping along the previous write, along the same layer's write at another token, and along a random direction, on the same tokens (95\% half-widths at most \FigRatioHalfMax). Values in Tables~\ref{tab:family} and~\ref{tab:group}.}
\label{fig:sources}
\end{figure*}

\paragraph{The lobe term carries the average damping in gated models.} In all ten gated models the lobe term is at least \RLobeShareMin\% of the window average damping $\bar D$ (Figure~\ref{fig:sources}a). In Gemma-2B, Phi-3-mini and OLMo-1B the closed side carries \RClosedSideShareGemmaTwoB\%, \RClosedSideSharePhiThreeMini\% and \RClosedSideShareOLMoOneB\% of it, and open weakening neurons, the route through which OLMo-1B first began to cancel (Section~\ref{sec:law}), supply most of the rest. In the other seven, all SwiGLU and the mixture OLMoE-1B-7B among them, the closed side is all of it: removing every closed-gate term from Equation~\eqref{eq:trace} leaves between $\CfSwiMin$ and $+\CfSwiMax$, because the open side's terms cancel each other, open strengthening neurons expanding what they read and open weakening ones contracting it. Between \RClosedMin{} and \RClosedMax{} of gate activations are closed, and between \StrShareMin\% and \StrShareMax\% of the window's neurons are strengthening.

\paragraph{The lobe term carries the damping of the previous write itself.} The average damping $\bar D$ is not the quantity in Equation~\eqref{eq:switch}, which is the damping $D$ of the particular direction $v = c_{j-1}$, so we split that damping by neuron group as well. With each neuron's group fixed by its state at the model's actual input (gate closed or open; strengthening or weakening), MLP $j$'s output is a sum of four group outputs, and $C$ and $D$ split exactly into four parts each (totals within $\GrLawGapMax$ of Figure~\ref{fig:law}; Figure~\ref{fig:sources}b and Table~\ref{tab:group}). In every gated model the closed strengthening neurons carry most or all of the write's damping ($D_g$ between \GrShareMin\% and \GrShareMax\% of $D$). Open strengthening neurons push the other way, by $\GrOpenStrMin$ to $\GrOpenStrMax$, because an open gate amplifies; open weakening neurons add \GrOpenWeakMin{} to \GrOpenWeakMax, most in OLMo-1B, Gemma-2B and Phi-3-mini; closed weakening neurons contribute within \GrClosedWeakAbsMax{} of zero. The same split of the local damping along $v$ gives the same picture (Table~\ref{tab:grouplocal}; closed strengthening group within \GrLocalGapMax), and removing $v$ flips the gate state of \SelFlipMin\% to \SelFlipMax\% of the neurons, carrying \SelFlipWMin\% to \SelFlipWMax\% of the activation mass. Across tokens and window pairs, the controlled projection of $c_j$ on $c_{j-1}$ falls as MLP $j$'s lobe term at that token rises: partial Spearman correlation $\PtLobeTokMin$ to $\PtLobeTokMax$ in the ten gated models, with pair, position, norms, token class and frequency partialled out (Appendix~\ref{app:pertoken}).
\paragraph{Non-gated models build the damping from active neurons.} A GELU or ReLU MLP without a gate has no closed gate to leak: an aligned neuron contracts what it reads only where the slope of its activation is negative, which for GELU is below $t = -0.75$ with slope never below $-0.13$, and for ReLU nowhere. In Pythia-160m, GPT-2 and GPT-2 medium the damping comes instead from active neurons whose output weight opposes their input weight (Figure~\ref{fig:sources}a; median read--write cosines $\RRWPythiaSmall$, $\RRWGPTTwo$ and $\RRWGPTTwoMedium$): in a gated MLP the activation's value enters the trace and in a non-gated MLP only its slope does (Appendix~\ref{app:decomp}), so the two reach damping through opposite read--write signs.

\paragraph{Is the previous write singled out?} \citet{patrawala2025tlcm} interpret the cancellation as selective correction of the previous layer's write. Three comparisons on the same tokens show that the MLP's damping does not single that write out (Figure~\ref{fig:sources}c; Appendix~\ref{sec:spectrum}, Tables~\ref{tab:selectivity} and~\ref{tab:eigen}, with intervals). First, against a uniform random direction. Along the controlled write, the local damping is within \CtlPrevWithinMax\% of the damping along a random direction in \CtlPrevNWithinWord{} of the ten gated models and \CtlPrevAboveTopExcess\% above it in \CtlPrevAboveTopName; along the raw write it is \PrevRatioAboveMin\% to \PrevRatioAboveMax\% above it (Table~\ref{tab:family}). Second, against directions of the write's own kind. The write of the \emph{same} layer at \emph{another token} of the document is damped \emph{more} than the token's own previous write in every model (ratio \SelRatioPrevOtherMin{} to \SelRatioPrevOtherMax), and so is the same layer's write at a token of \emph{another document} (ratio \SelRatioPrevOdocMin{} to \SelRatioPrevOdocMax). Other layer writes exceed the random direction by \SelOtherRandMin\% to \SelOtherRandMax\%: layer writes as a class overlap the neurons' read directions more than a uniform direction does, and the token's own previous write is at the low end of that class. Third, the write's energy over the eigendirections of the MLP's symmetrised Jacobian. A direction selectively corrected would place its energy on a few strongly contracting eigendirections, as the fifty eigenvectors that carry most of the previous write do in the layer Jacobian of \citet{patrawala2025tlcm}. Here the write's energy has about twice a random direction's share in both tails of the spectrum and follows it elsewhere: \EigFullPrevMin\% to \EigFullPrevMax\% on eigendirections damped by more than $0.8$ against \EigFullRandMin\% to \EigFullRandMax\%, as much on expanded eigendirections as a random direction, and a participation ratio of \EigPrevPRFracMin{} to \EigPrevPRFracMax{} of $d - 1$ against \EigRandPRFracMin{} to \EigRandPRFracMax{} (Table~\ref{tab:eigen}).

The whole layer's Jacobian, which \citet{patrawala2025tlcm} analysed, gives the same answer on their model, their text and their token--layer pairs. The fifty eigendirections that carry most of the previous write have the bimodal eigenvalue distribution they report (\LjTlcmPrevTopBelowLlamaThreeEightB\% of them below $-0.75$ and \LjTlcmPrevTopAboveLlamaThreeEightB\% above $+0.25$), and so do the fifty that carry the same layer's write at another token (\LjTlcmOtherTopBelowLlamaThreeEightB\% and \LjTlcmOtherTopAboveLlamaThreeEightB\%). To first order the layer's response along a unit direction $c$ is $c^{\top}\bar J c$: $\LjTlcmPrevWmeanLlamaThreeEightB$ for the previous write, $\LjTlcmOtherWmeanLlamaThreeEightB$ for another token's write and $\LjTlcmRandWmeanLlamaThreeEightB$ for a random direction, the MLP's path carrying $\LjTlcmMlpQuadPrevLlamaThreeEightB$ of the first. Across the \LjTlcmNModelsWord{} models run on their text, the layer contracts the previous write \LjTlcmPrevOverRandPctMin\% to \LjTlcmPrevOverRandPctMax\% more than a random direction and $\LjTlcmPrevOverOtherPctMin$\% to $\LjTlcmPrevOverOtherPctMax$\% relative to another token's write (Appendix~\ref{sec:spectrum}, Table~\ref{tab:layerjac}).

%% file: figures/sources.tex
\begin{tikzpicture}
\begin{groupplot}[group style={group size=3 by 1, horizontal sep=0.45cm, yticklabels at=edge left}, paper, zero x, scale only axis, height=4.2cm,
  ymin=-0.6, ymax=\FigYMax, enlarge y limits=false, ytick={\FigYTicks}, major y tick style={draw=none}, ymajorgrids=false,
  extra y ticks={\FigSepCG, \FigSepNC}, extra y tick labels={}, extra y tick style={grid=major, grid style={draw=axisgray, line width=0.4pt}, major tick length=0pt}, clip=false]
\nextgroupplot[width=4.55cm, modelrows={figures/data/figsrc.csv}{\FigYTicks}, xmin=-0.02, xmax=0.53, xtick={0,0.1,0.2,0.3,0.4,0.5},
  title={(a) average damping $\bar D$, by source}, xlabel={damping of a random direction},
  legend to name=srcleg, legend columns=5, legend style={/tikz/every even column/.append style={column sep=0.35cm}}]
\addlegendimage{area legend, fill=series1, draw=series1, line width=0pt}\addlegendentry{closed strengthening neurons (lobe term)}
\addlegendimage{area legend, fill=series2, draw=series2, line width=0pt}\addlegendentry{open weakening neurons}
\addlegendimage{area legend, fill=series3, draw=series3, line width=0pt}\addlegendentry{open strengthening}
\addlegendimage{area legend, fill=inkmuted, draw=inkmuted, line width=0pt}\addlegendentry{closed weakening}
\addlegendimage{only marks, mark=|, mark size=3pt, mark options={line width=1pt, draw=black!85}}\addlegendentry{total}
\addplot[series1, line width=2.9pt, hbar=lobe, forget plot] table[x expr=0, y expr=\thisrow{y}+0.2] {figures/data/figsrc.csv};
\addplot[series2, line width=2.9pt, hbar=weak, forget plot] table[x expr=0, y expr=\thisrow{y}-0.2] {figures/data/figsrc.csv};
\addplot[tick mark=black!85, forget plot] table[x=Dbar, y=y] {figures/data/figsrc.csv};
\node[group label, anchor=east] at (axis cs:0.525,7.4) {gated};
\node[group label, anchor=east] at (axis cs:0.525,4.7) {ReLU conversions};
\node[group label, anchor=east] at (axis cs:0.525,0) {no gate};
\nextgroupplot[width=4.55cm, xmin=-0.1, xmax=0.6, xtick={-0.1,0,0.1,0.2,0.3,0.4,0.5}, title={(b) damping $D$ of the write, by group}, xlabel={per unit of the previous write}]
\addplot[series1, line width=5pt, hbar=csl, forget plot] table[x=css, y=y] {figures/data/figgrp.csv};
\addplot[series2, line width=5pt, hbar=owl, forget plot] table[x=ows, y=y] {figures/data/figgrp.csv};
\addplot[series3, line width=5pt, hbar=osl, forget plot] table[x=oss, y=y] {figures/data/figgrp.csv};
\addplot[inkmuted, line width=5pt, hbar=cwl, forget plot] table[x=cws, y=y] {figures/data/figgrp.csv};
\addplot[tick mark=black!85, forget plot] table[x=total, y=y] {figures/data/figgrp.csv};
\node[group label, align=center] at (axis cs:0.25,2.4) {no gate, so no closed\\ or open side};
\nextgroupplot[width=3.6cm, xmin=0.1, xmax=0.56, xtick={0.1,0.2,0.3,0.4,0.5}, title={(c) which direction is damped}, xlabel={local damping},
  legend style={at={(0.5,0.08)}, anchor=south, legend columns=1}]
\addplot[dot=series1] table[x=write, y=y] {figures/data/figratio.csv}; \addlegendentry{the previous write}
\addplot[only marks, mark=triangle*, mark size=2.6pt, series2, line width=0.8pt] table[x=other, y=y] {figures/data/figratio.csv}; \addlegendentry{the same layer's write at another token}
\addplot[only marks, mark=o, mark size=2.3pt, black!70, line width=0.8pt] table[x=rand, y=y] {figures/data/figratio.csv}; \addlegendentry{a random direction}
\end{groupplot}
\end{tikzpicture}

\smallskip
\pgfplotslegendfromname{srcleg}

%% file: interventions.tex
\section{Interventions}
\label{sec:interventions}

Sections~\ref{sec:law} and~\ref{sec:mechanism} attribute the damping to the closed side by measurement alone; the experiments below change it.

\begin{figure*}[t]
\centering
\input{figures/contpre_QwenHalfB}
\caption{Continued pretraining of Qwen2.5-0.5B for \CpTokensM M tokens with the closed side of every gate activation scaled by $\lambda$, annealed from $1$ to $0$ (solid; $\lambda$ in panel c, right axis) or left at $1$ (dashed), from the same weights and batches; window means on the common documents, means over three seeds with bands spanning the seeds (most are narrower than the line). (a) The damping $D$ of the previous write, $A + C$, and the cosine between adjacent contributions. (b) $\bar D$ and its lobe term. (c) Loss on the common documents. Values in Table~\ref{tab:contpre}; Llama-3.2-1B in Figure~\ref{fig:contpre-llama}.}
\label{fig:contpre}
\end{figure*}

\paragraph{In continued pretraining of models that cancel.} Qwen2.5-0.5B cancels on the common documents (window cosine $\CpStartCos$; damping of the previous write $D = \CpStartD$). We trained it for a further \CpTokensM M tokens of FineWeb-Edu in two arms from the same weights and the same batches, for three seeds of the batch order (Appendix~\ref{app:contpre}). In the \emph{annealed} arm the closed side of every gate activation is scaled by $\lambda$ in the forward pass and the gradient, with $\lambda$ falling linearly from $1$ to $0$ between \CpAnnealStartM M and \CpAnnealEndM M tokens and held at $0$ for the remaining \CpZeroTokensM M; the \emph{control} arm trains with the activation unchanged. Figure~\ref{fig:contpre} follows both. The control keeps the cancellation (cosine $\CpEndControlCos$, $D = \CpEndControlD$ at the end). In the annealed arm the damping falls with $\lambda$ (Figure~\ref{fig:contpre}a), and once $\lambda = 0$ the lobe term is zero by construction, the damping of the previous write is $\CpEndAnnealD$ (average damping \CpEndAnnealDbar{} against \CpEndControlDbar), and the window cosine is $+\CpEndAnnealCos$: the model no longer cancels, and it has not rebuilt the damping through its open neurons (weakening term \CpEndAnnealWeak{} against \CpEndControlWeak). Co-writing falls too, from \CpStartC{} to \CpEndAnnealC{} (control \CpEndControlC). The annealed model remains a language model but a worse one: loss \CpEndAnnealLoss{} on the common documents against \CpEndControlLoss{} (validation gap \CpEndValGap, still narrowing). The three seeds agree to within \CpSeedRangeMax{} on every quantity quoted. Llama-3.2-1B, trained the same way for one seed, does the same: its window cosine goes from $\CpLlamaThreeOneBStartCos$ to $+\CpLlamaThreeOneBEndAnnealCos$, against $\CpLlamaThreeOneBEndControlCos$ in the control, at a loss cost of \CpLlamaThreeOneBEndLossGap{} nats. In both models the damping and the cancellation went with the closed side: neither survived \CpZeroTokensM M tokens of training without it, while the same training with it kept both.

\paragraph{In training from scratch, at 51M parameters.} Llama-style models of 51.5M parameters trained on 1B tokens of FineWeb-Edu from shared initial weights and data order, differing only in the closed side of the gate activation (two seeds; Figure~\ref{fig:arms} and the pre-registered analysis in Appendix~\ref{app:train}), show that the closed side produces most of the damping at this scale: with it removed ($\lambda = 0$) the models damp \ArmLamZeroShare\% as much as their SwiGLU twins, through weakening neurons alone. From $\lambda = 0.5$ to $2$ the damping rises with an elasticity of \ArmElasticityZero{} and \ArmElasticityOne{} (seeds~0 and~1), inside the band that the pre-registered analysis labels partial compensation. Two seeds detect no loss benefit of the closed side (\ArmLamZeroLoss{} at $\lambda = 0$ against \ArmLamOneLoss{} at $\lambda = 1$). The two arms that keep one of the forward pass and the gradient through the closed side without the other train worse (by at least \ArmDetachLossMin{} and \ArmStLossMin), so they do not separate the absence of the closed side from disrupted training (Appendix~\ref{app:train}). No arm cancels at 1B tokens, nor does a 125M pair trained for 12B tokens (Appendix~\ref{app:train}), so these runs speak to the damping and not to the cancellation.

\paragraph{At inference, in the pretrained and converted models.} Removing the closed side from every MLP of the \LoNModels{} pretrained gated models brings the window damping within \LoDampAbsMax{} of zero but breaks the models (loss \LoLossMin{} to \LoLossMax{} nats); a size-matched removal of open entries also breaks them and leaves the damping (Appendix~\ref{app:inference}). CATS-style thresholding \citep{lee2024cats} at 50\% sparsity keeps the models working and lowers the damping to \CatsFiveDampRatioMin{} to \CatsFiveDampRatioMax{} of its value, with the cosine rising in all \CatsNModels{} models, at a loss cost of \CatsFiveDLossMin{} to \CatsFiveDLossMax{} nats. The degradation is not what moves them: zeroing open entries at random at the rate that matches each model's CATS loss (\CcCtlPMin\% to \CcCtlPMax\% of open entries) leaves the damping at \CcCtlDRatioMin{} to \CcCtlDRatioMax{} of its value and the cosine within \CcCtlCosDeltaAbsMax{} of where it was in all \CcNModelsWord{} models (Table~\ref{tab:catsctl}). Two public ReLU conversions of Llama-2-7B, ReluLLaMA-7B \citep{relullama} and ProSparse-7B \citep{song2025prosparse}, keep \DampKeepReluLlama\% and \DampKeepProSparse\% of its window damping while remaining working language models; but they differ from their parent in distillation or in instruction data and a sparsity regulariser, and the parent's own window cancellation is weak (controlled $\RCosLlamaTwoSevenB$; Appendix~\ref{app:inference}).

%% file: figures/contpre_QwenHalfB.tex
\begin{tikzpicture}
\begin{groupplot}[group style={group size=3 by 1, horizontal sep=1.2cm}, width=0.33\linewidth, height=3.1cm, paper, xlabel={training tokens (millions)}, xmin=0, xmax=\CpTokensM]
\nextgroupplot[zero y, title={(a) along the previous write}, ymin=-0.25, ymax=0.4,
  legend to name=cplegQwenHalfB, legend columns=6, legend style={/tikz/every even column/.append style={column sep=0.3cm}}]
\addplot[name path=aDhi, draw=none, forget plot] table[x=tok, y=aDmax] {figures/data/figcpQwenHalfB.csv};
\addplot[name path=aDlo, draw=none, forget plot] table[x=tok, y=aDmin] {figures/data/figcpQwenHalfB.csv};
\addplot[series1, fill opacity=0.18, draw=none, forget plot] fill between[of=aDhi and aDlo];
\addplot[name path=cDhi, draw=none, forget plot] table[x=tok, y=cDmax] {figures/data/figcpQwenHalfB.csv};
\addplot[name path=cDlo, draw=none, forget plot] table[x=tok, y=cDmin] {figures/data/figcpQwenHalfB.csv};
\addplot[series1, fill opacity=0.18, draw=none, forget plot] fill between[of=cDhi and cDlo];
\addplot[name path=aAChi, draw=none, forget plot] table[x=tok, y=aACmax] {figures/data/figcpQwenHalfB.csv};
\addplot[name path=aAClo, draw=none, forget plot] table[x=tok, y=aACmin] {figures/data/figcpQwenHalfB.csv};
\addplot[series2, fill opacity=0.18, draw=none, forget plot] fill between[of=aAChi and aAClo];
\addplot[name path=cAChi, draw=none, forget plot] table[x=tok, y=cACmax] {figures/data/figcpQwenHalfB.csv};
\addplot[name path=cAClo, draw=none, forget plot] table[x=tok, y=cACmin] {figures/data/figcpQwenHalfB.csv};
\addplot[series2, fill opacity=0.18, draw=none, forget plot] fill between[of=cAChi and cAClo];
\addplot[name path=acoshi, draw=none, forget plot] table[x=tok, y=acosmax] {figures/data/figcpQwenHalfB.csv};
\addplot[name path=acoslo, draw=none, forget plot] table[x=tok, y=acosmin] {figures/data/figcpQwenHalfB.csv};
\addplot[black, fill opacity=0.18, draw=none, forget plot] fill between[of=acoshi and acoslo];
\addplot[name path=ccoshi, draw=none, forget plot] table[x=tok, y=ccosmax] {figures/data/figcpQwenHalfB.csv};
\addplot[name path=ccoslo, draw=none, forget plot] table[x=tok, y=ccosmin] {figures/data/figcpQwenHalfB.csv};
\addplot[black, fill opacity=0.18, draw=none, forget plot] fill between[of=ccoshi and ccoslo];
\addplot[series1, mark=*, mark size=1.3pt] table[x=tok, y=aD] {figures/data/figcpQwenHalfB.csv}; \addlegendentry{$D$, annealed}
\addplot[series1, dashed, mark=o, mark size=1.3pt] table[x=tok, y=cD] {figures/data/figcpQwenHalfB.csv}; \addlegendentry{$D$, control}
\addplot[series2, mark=square*, mark size=1.3pt] table[x=tok, y=aAC] {figures/data/figcpQwenHalfB.csv}; \addlegendentry{$A + C$, annealed}
\addplot[series2, dashed, mark=square, mark size=1.3pt] table[x=tok, y=cAC] {figures/data/figcpQwenHalfB.csv}; \addlegendentry{$A + C$, control}
\addplot[black!70, mark=triangle*, mark size=1.5pt] table[x=tok, y=acos] {figures/data/figcpQwenHalfB.csv}; \addlegendentry{cosine, annealed}
\addplot[black!70, dashed, mark=triangle, mark size=1.5pt] table[x=tok, y=ccos] {figures/data/figcpQwenHalfB.csv}; \addlegendentry{cosine, control}
\nextgroupplot[zero y, title={(b) average damping and lobe term}, ymin=-0.02, ymax=0.35]
\addplot[name path=aDbarhi, draw=none, forget plot] table[x=tok, y=aDbarmax] {figures/data/figcpQwenHalfB.csv};
\addplot[name path=aDbarlo, draw=none, forget plot] table[x=tok, y=aDbarmin] {figures/data/figcpQwenHalfB.csv};
\addplot[series1, fill opacity=0.18, draw=none, forget plot] fill between[of=aDbarhi and aDbarlo];
\addplot[name path=cDbarhi, draw=none, forget plot] table[x=tok, y=cDbarmax] {figures/data/figcpQwenHalfB.csv};
\addplot[name path=cDbarlo, draw=none, forget plot] table[x=tok, y=cDbarmin] {figures/data/figcpQwenHalfB.csv};
\addplot[series1, fill opacity=0.18, draw=none, forget plot] fill between[of=cDbarhi and cDbarlo];
\addplot[name path=alobehi, draw=none, forget plot] table[x=tok, y=alobemax] {figures/data/figcpQwenHalfB.csv};
\addplot[name path=alobelo, draw=none, forget plot] table[x=tok, y=alobemin] {figures/data/figcpQwenHalfB.csv};
\addplot[series3, fill opacity=0.18, draw=none, forget plot] fill between[of=alobehi and alobelo];
\addplot[name path=clobehi, draw=none, forget plot] table[x=tok, y=clobemax] {figures/data/figcpQwenHalfB.csv};
\addplot[name path=clobelo, draw=none, forget plot] table[x=tok, y=clobemin] {figures/data/figcpQwenHalfB.csv};
\addplot[series3, fill opacity=0.18, draw=none, forget plot] fill between[of=clobehi and clobelo];
\addplot[series1, mark=*, mark size=1.3pt] table[x=tok, y=aDbar] {figures/data/figcpQwenHalfB.csv};
\addplot[series1, dashed, mark=o, mark size=1.3pt] table[x=tok, y=cDbar] {figures/data/figcpQwenHalfB.csv};
\addplot[series3, mark=diamond*, mark size=1.5pt] table[x=tok, y=alobe] {figures/data/figcpQwenHalfB.csv};
\addplot[series3, dashed, mark=diamond, mark size=1.5pt] table[x=tok, y=clobe] {figures/data/figcpQwenHalfB.csv};
\nextgroupplot[title={(c) loss on the common documents, $\lambda$}, ymin=2.6, ymax=3.05, axis y line*=left]
\addplot[name path=alosshi, draw=none, forget plot] table[x=tok, y=alossmax] {figures/data/figcpQwenHalfB.csv};
\addplot[name path=alosslo, draw=none, forget plot] table[x=tok, y=alossmin] {figures/data/figcpQwenHalfB.csv};
\addplot[black, fill opacity=0.18, draw=none, forget plot] fill between[of=alosshi and alosslo];
\addplot[name path=closshi, draw=none, forget plot] table[x=tok, y=clossmax] {figures/data/figcpQwenHalfB.csv};
\addplot[name path=closslo, draw=none, forget plot] table[x=tok, y=clossmin] {figures/data/figcpQwenHalfB.csv};
\addplot[black, fill opacity=0.18, draw=none, forget plot] fill between[of=closshi and closslo];
\addplot[black!85, mark=*, mark size=1.3pt] table[x=tok, y=aloss] {figures/data/figcpQwenHalfB.csv};
\addplot[black!85, dashed, mark=o, mark size=1.3pt] table[x=tok, y=closs] {figures/data/figcpQwenHalfB.csv};
\end{groupplot}
\begin{axis}[at={(group c3r1.south west)}, anchor=south west, width=0.33\linewidth, height=3.1cm, xmin=0, xmax=\CpTokensM, ymin=0, ymax=1.05, axis y line*=right, axis x line=none, ylabel={$\lambda$}, ylabel near ticks, ytick={0,0.5,1}, tick label style={font=\scriptsize, text=inksecond}, label style={font=\footnotesize}, axis line style={draw=axisgray, line width=0.4pt}, tick style={draw=axisgray, line width=0.4pt}]
\addplot[inkmuted, thick, no marks] table[x=tok, y=lam] {figures/data/figcpQwenHalfB.csv};
\end{axis}
\end{tikzpicture}

\smallskip
\pgfplotslegendfromname{cplegQwenHalfB}

%% file: related.tex
\section{Related work}
\label{sec:related}

\paragraph{Adjacent-layer cancellation.} \citet{patrawala2025tlcm} describe the phenomenon, show that it emerges in training, that the next layer's correction scales with the previous layer's write, and that it is concentrated in adjacent layers. Equation~\eqref{eq:switch} separates their scaling intervention into attention, co-writing and damping, and Appendix~\ref{app:lags} follows the terms to older writes, where the concentration in adjacent pairs sits in the terms rather than in any selectivity of the damping. Their two exceptions appear here as GPT-2 writing more along the previous write in attention and co-writing, and as Phi-3-mini cancelling once each layer's shared component is removed (Appendix~\ref{app:raw}). They find the normaliser's blindness to the stream's norm insufficient as an explanation (their Appendix~A); here the normaliser enters as the factor $\sqrt d\,\|x'\|^{-1}$, which carries the growth of the stream's norm with depth \citep{heimersheim2023norms, sun2025curse}, and as the projector $P$, the form of its Jacobian noted by \citet{xiong2020layernorm} and \citet{brody2023layernorm}. From the eigenvalues of the whole layer's symmetrised Jacobian, weighted by the previous write, they conclude that the correction is selective, the natural reading in the iterative-refinement view of the stream \citep{greff2017highway, jastrzebski2018residual}; on their model and text the same Jacobian contracts another token's write and a random direction as it does the previous write, and the MLP's Jacobian spreads the write's energy nearly as a random direction's (Section~\ref{sec:mechanism}). Per-layer Jacobians across depth \citep{fernando2026dynamics} and the deletion or swapping of layers \citep{lad2025stages, gromov2025deeper, csordas2025depth} are neighbouring measurements.

\paragraph{Input--output functionality of neurons.} \citet{gerstner2025gated} classify gated neurons by the cosines between their input, gate and output weights (thresholds $\pm0.5$), find \emph{conditional enrichment} neurons dominant in early and middle layers, and describe the sign flip under a weakly negative Swish value (their Sections~4.3 and~6.2). Their work gives the per-neuron reversal that this paper sums over neurons, and given their counts, that aligned neurons dominate the early and middle layers and rarely have positive gate values, it follows that those neurons contract what they read. This paper adds the sum: that it is the largest term of the MLP's contraction of every direction, its size (\FamLobeGatedMinPct--\FamLobeGatedMaxPct\% of a direction per layer), that it carries the damping of the previous layer's write, and that adjacent-layer cancellation decomposes into that damping. \citet{gerstner2026weakening}, whose terms strengthening and weakening we adopt, find that weakening neurons have a strong effect when their gate values are negative; GLUScope \citep{gerstner2026gluscope} shows a neuron's activations by sign combination; \citet{pearce2025bilinear} analyse gated MLPs without the nonlinearity from their weights; the reading of a neuron by its input and output weights is that of \citet{geva2021kv, geva2022promoting}. \citet{gurnee2024universal} note the near-opposite weights of very frequently active GPT-2 and Pythia neurons, and \citet{voita2024neurons} find OPT neurons that erase their triggering token; OPT-1.3B has no window damping.

\paragraph{Self-repair, erasure and activations.} Ablating a component makes later components compensate \citep{mcgrath2023hydra, rushing2024selfrepair}, and \citet{janiak2024memory} show heads erasing earlier writes; by the definition of $D$, removing layer $j-1$'s write makes the next MLP write $D\,\|c_{j-1}\|$ more along it. Copy suppression \citep{mcdougall2023copy} is a later component writing against an earlier one selectively; the damping here is an average over directions, tested at the token level by the comparison of a token's own previous write with another token's (Section~\ref{sec:mechanism}). \citet{stolfo2024confidence} describe erasure through the final normaliser's scale and \citet{feucht2024erasure} erasure of token identity in early layers. The shape of a gated activation's negative branch is itself a design choice \citep{huang2024expanded, huang2024xielu}, and \citet{kong2025negative} find the negative pre-activations of a small set of neurons functionally used. Converting the activation to ReLU and training further makes activations sparse \citep{mirzadeh2024relu, song2024turbosparse, song2025prosparse, relullama, zhang2024relusq}; CATS \citep{lee2024cats} and TEAL \citep{liu2025teal} sparsify by magnitude at inference, and at 50\% sparsity \CatsFiveZeroedClosedMin\% to \CatsFiveZeroedClosedMax\% of what CATS zeroes in the models here is closed. Transcoders fit an MLP's output from its input \citep{dunefsky2024transcoders} or from the stream at that and earlier layers \citep{ameisen2025circuit}; in gated models what they fit includes this contraction, and \citet{shabalin2026sparsified} find that many transcoder features are single gated neurons split by the sign of their activation.

%% file: discussion.tex
\section{Discussion and conclusion}
\label{sec:discussion}

\paragraph{Summary.} The cancellation decomposes, by a stated counterfactual, into attention, co-writing and damping, and across models the spread of the outcome is the spread of the damping. In every gated model closed strengthening neurons carry between \GrShareMin\% and \GrShareMax\% of the damping of the previous write, and that write is not singled out. In Qwen2.5-0.5B and Llama-3.2-1B the cancellation went with the closed side: annealed away in continued pretraining, it took the damping and the cancellation with it, at a loss cost of \CpEndLossGap{} and \CpLlamaThreeOneBEndLossGap{} nats, while the same training with the activation unchanged left both in place. Two questions the account leaves open: whether a model trained from scratch without the closed side would cancel by the open-neuron route of the non-gated models, and which gated models cancel, since within the gated family co-writing rises with the damping ($r = \LawRCDGated$) and the account explains the damping alone.

\paragraph{Is the damping useful?} \citet{patrawala2025tlcm} ask whether the cancellation is ``the most efficient mechanism available, or a consequence of architectural constraints''. At 51M parameters two seeds detect no loss benefit of the closed side, but a difference below about $0.01$ nats would not show, which is the ReGLU-to-SwiGLU gap \citet{shazeer2020glu} report at the T5-base scale, so the null says little; in Qwen2.5-0.5B and Llama-3.2-1B removing the closed side cost \CpEndLossGap{} and \CpLlamaThreeOneBEndLossGap{} nats that \CpZeroTokensM M tokens had not recovered. Whether the contraction helps optimisation at scale is open.

\paragraph{Limitations.} The models studied have at most 8B parameters; two are conversions of a third and one is distilled from another. The main measurements use sixteen documents of one corpus at 192 tokens, with intervals over documents within that corpus only (64 documents move the means by at most \NsFDGapMax); the decomposition is repeated on a second corpus and at 2{,}048 tokens (Appendix~\ref{app:controls}). The selectivity tests are at the level of directions; whether particular features are singled out would need a feature dictionary. The from-scratch models do not reach cancellation; the continued-pretraining result is two models, one seed of the larger, and \CpTokensM M tokens; the conversions differ from their parent in more than the activation. OLMo-2 and Gemma-2, which normalise the MLP's output, are not decomposed.

\paragraph{Implications.} Three follow, none tested here. For interpretability, the next gated MLP writes against most directions it reads, so methods that follow a feature or a steering direction across layers, or fit a layer's output from its input as transcoders do, could account for this contraction before reading a correction as content. For architecture, the negative region of SiLU and GELU shapes how adjacent layers interact, and trainable negative branches, ReLU conversion and magnitude-based sparsification each change it. For training, in every gated run we followed the closed side's damping and the alignment of the neurons' input and output weights rose together, which makes that alignment worth tracking during pretraining.

%% file: appendix.tex
\section{Measurement details}
\label{app:controls}

\paragraph{Texts.} The documents of the FineWeb-Edu sample-10BT subset's files \texttt{000\_00000} and \texttt{001\_00000}, in file order, are split so that the last 1\% are held out (the validation split of the 51M models); every text is tokenised with GPT-2's BPE (no special tokens) and decoded again, so that the common set is the same string for every model. The common sixteen documents are the first sixteen of the held-out documents with at least 900 characters, in order, truncated to 1{,}500 characters and then to 192 tokens of each model's tokenizer; the 64-document set is the first 64 such documents, and the check of the outlier dimensions uses the next sixteen after the common set. Position~0 is dropped from every measurement. The replication of Figure~\ref{fig:protocol} uses the first \TpNArticles{} articles of the WikiText-103 test split that reach 2{,}048 tokens of the model's tokenizer (1{,}024 for GPT-2), each article starting at its title line.

\paragraph{Contributions and the controlled frame.} A layer's contribution $c_l$ is its output minus its input, recorded at every layer. The five outlier dimensions are those with the largest mean absolute contribution over all layers and all tokens of the sixteen documents; they are zeroed in every contribution, and each layer's mean contribution over those tokens is then subtracted. Table~\ref{tab:cosctl} gives the window cosine under each control separately and the controlled cosine over all pairs.

\begin{table}[htbp]
\centering
\small
\setlength{\tabcolsep}{3.2pt}
\caption{Adjacent cosine under separate controls (FineWeb-Edu). Top-$k$: the $k$ largest-magnitude dimensions zeroed. Mean removed: each layer's mean contribution subtracted. Controlled: top-5 and mean removed. Standardised: each dimension z-scored over tokens \citep{timkey2021rogue}. Last column: the controlled cosine averaged over all adjacent pairs.}
\label{tab:cosctl}
\fittable{\input{supplement/table_cos_controls_r1}}
\end{table}

\paragraph{The window and the replication.} The window generalises the $4 \le i < 20$ that \citet{patrawala2025tlcm} use for their 32-layer model when calibrating a null (their Appendix~C.2); they find the cancellation strongest in the first two-thirds of the layers, and a quantity of a single layer is in the window when its pair is. Averaged over the middle half of the layers instead, which includes late pairs that do not cancel, the raw cosine is less negative in \TpMidLessNegWord{} of the \TpWindowNegWord{} models whose window cosine is negative (Figure~\ref{fig:protocol}a). Their protocol averages about 100{,}000 tokens across random WikiText documents; we approximate it by \TpNArticles{} WikiText-103 articles \citep{merity2017wikitext} of 2{,}048 tokens each, on which the raw window cosine differs from ours by at most \TpCorpusGapMax{} and across positions by at most \TpPosGapMax. For Llama 3.1 8B, their main model, we measure $\TpWikiRawLlamaThreeEightB$ with the Instruct release used in the tables and $\TpWikiRawLlamaThreeEightBBase$ with the base release, against the ``$\approx -0.2$'' they report; the base release cancels by $\RCosLlamaThreeEightBBase$ (controlled) on the common documents against $\RCosLlamaThreeEightB$ for Instruct, with the same decomposition ($A = \SpCtlALlamaThreeEightBBase$, $C = \SpCtlCLlamaThreeEightBBase$, $D = \SpCtlDLlamaThreeEightBBase$; closed strengthening neurons carry \GrDClosedStrShareLlamaThreeEightBBase\% of $D$, and the write is damped \SelRatioPrevRandLlamaThreeEightBBase{} times a random direction and \SelRatioPrevOtherLlamaThreeEightBBase{} times another token's write), and across positions $1$--$192$ to $1{,}024$--$2{,}048$ its raw window cosine moves from $\TpPosFirstLlamaThreeEightBBase$ to $\TpPosLastLlamaThreeEightBBase$; the per-token window cosines have median $\TpWikiMedianLlamaThreeEightBBase$ and mode $\TpWikiModeLlamaThreeEightBBase$ (bins of $0.02$). Their released code fixes the token set: WikiText-103 training documents of at least 2{,}000 characters, shuffled with seed 42, \TpxDocsLlamaThreeEightB{} documents truncated to 1{,}024 tokens, position~0 included, the model in bfloat16, and the per-layer contributions taken as differences of consecutive hidden states. Run as released (\TpxTokensMLlamaThreeEightB M tokens), it gives a mean over the window pairs of $\TpxWindowLlamaThreeEightB$ for the Instruct release and $\TpxWindowLlamaThreeEightBBase$ for the base release, within $0.01$ of ours, and over all adjacent pairs $\TpxAllPairsLlamaThreeEightB$ and $\TpxAllPairsLlamaThreeEightBBase$; the strongest single pair in the window is $\TpxMinPairLlamaThreeEightB$ (Instruct) and $\TpxMinPairLlamaThreeEightBBase$ (base), both at pair $\TpxStrongPairIdxLlamaThreeEightB \to \number\numexpr\TpxStrongPairIdxLlamaThreeEightB+1\relax$, and \TpxNPairsBelowPointFifteenLlamaThreeEightB{} and \TpxNPairsBelowPointFifteenLlamaThreeEightBBase{} of the sixteen window pairs lie below $-0.15$. Their text gives the adjacent cosine as averaging $\approx -0.2$; on their protocol we measure a window mean of $\TpxWindowLlamaThreeEightB$, strongest pairs at $\TpxMinPairLlamaThreeEightB$ to $\TpxMinPairLlamaThreeEightBBase$, and window means that agree with ours. Table~\ref{tab:cosctl} gives the window cosine under each control separately: the raw cosine, $k \in\{1, 2, 5, 20\}$ dimensions zeroed, the mean removed, and the standardised cosine of \citet{timkey2021rogue}; for Phi-3-mini, Gemma-2B and ProSparse-7B the sign of the window cosine depends on the control (Appendix~\ref{app:raw}).

\paragraph{The outlier dimensions on other documents.} The five outlier dimensions are chosen on the sixteen measured documents. On the next sixteen FineWeb-Edu documents of the same split, the same five dimensions are chosen in \OdNAllFiveBWord{} of the \OdNModelsWord{} models checked and at least \OdOverlapBMin{} of the five in all of them, and on sixteen WikiText-103 articles at least \OdOverlapWikiMin{} of the five, with all five of the common set's dimensions among the twenty largest on both other sets in every model checked: the dimensions are a property of the model, not of the documents.

\paragraph{The decomposition on a second corpus and at 2{,}048 tokens.} Table~\ref{tab:wiki} repeats the controlled decomposition of Table~\ref{tab:switchci} on the first sixteen WikiText-103 test articles of at least 900 characters, truncated as the FineWeb-Edu documents are (1{,}500 characters, 192 tokens), with the outlier dimensions and layer means recomputed on those articles. The damping changes by at most \WkDGapMax{} and $A + C$ by at most \WkACGapMax{} in any model, the sign of $A + C - D$ agrees in \WkNSignAgreeWord{} of the \WkNModelsWord{} models (disagreeing: \WkSignDisagreeNames), the across-model spread of $D$ is \WkSdRatio{} times that of $A + C$, and the rank order of $A + C - D$ across models has a Spearman correlation of \WkLawRank{} with the FineWeb-Edu order. At 2{,}048 tokens on WikiText-103 articles that reach that length (\WkLongNModelsWord{} models, GPT-2's context being 1{,}024), the damping of the \WkLongNGatedWord{} gated models is \WkLongDRatioMin{} to \WkLongDRatioMax{} of its 192-token value, the sign of $A + C - D$ agrees with the 192-token sign in \WkLongNSignAgree{} of \WkLongNModelsWord, and $A + C - D$ lies between $\WkLongLawMin$ and $\WkLongLawMax$.

\paragraph{Compute.} Each 51M run took about an hour and each 125M run 21 hours; the continued-pretraining arms took 9 hours per seed for Qwen2.5-0.5B and 12 for Llama-3.2-1B (39 hours in all); the measurements on the sixteen released models and their 40 training checkpoints, including the Jacobians of Appendix~\ref{sec:spectrum}, took on the order of 50 hours; about 150 GPU-hours in all on one RTX~5090.

\paragraph{Details that fix each measurement.} Random directions are Gaussian vectors projected orthogonal to $x'$ and normalised, one per token, redrawn per layer. The eigendecompositions of Appendix~\ref{sec:spectrum} use position 100 of each document (the last position for shorter documents) and, for the other token, position $100 + 37$ modulo the document's length; the other document is the next one in the common set at the same position; the four window layers are spread evenly over the window. The gate-flip share and the local damping by group fix each neuron's group at the model's actual input. CATS thresholds (Appendix~\ref{app:inference}) are set per layer as the quantile of $|\phi(t)|$ over a random two-million subsample of the gate activations on the first eight WikiText-103 test articles of at least 2{,}048 tokens, each cut to 2{,}048; the size-matched control for the closed side's removal zeroes, at each token, every open entry independently with probability equal to the closed side's share of that token's $\sum_k |\phi(t_k)|$; the loss-matched control for CATS zeroes open entries at rates of 0.5, 1, 2, 4, 8, 15 and 30\% and the rate that matches the CATS loss is read off by linear interpolation between the two bracketing rates. The released checkpoints followed are OLMo-1B-0724 at steps 1,000, 2,000, 4,000, 8,000, 16,000, 31,000, 62,000, 123,000, 245,000, 358,000, 477,000, 716,000, 954,000, 1,193,000 and 1,431,000 (2B to 3T tokens); Pythia-160m at steps 1,000, 2,000, 4,000, 8,000, 16,000, 32,000, 64,000 and 143,000; TinyLlama-1.1B at 105B, 503B, 1T, 1.5T, 2T, 2.5T and 3T tokens; OLMoE-1B-7B-0924 at steps 10,000, 50,000, 100,000, 200,000, 400,000, 600,000, 800,000, 1,000,000 and 1,200,000 and the release (41B to 5.1T tokens). For continued pretraining (Appendix~\ref{app:contpre}) the file \texttt{002\_00000} is tokenised with the model's tokenizer into one stream with the end-of-sequence token between documents; the first 330M tokens are the training stream, from which each batch takes 1{,}024-token windows at uniformly random offsets, and the next 4M tokens hold the validation windows. The 51M models (Appendix~\ref{app:train}) train on the 99\% of files \texttt{000\_00000} and \texttt{001\_00000} that is not held out, as one GPT-2 token stream with the end-of-text token between documents.

\paragraph{Sixty-four documents.} The window measurements of Tables~\ref{tab:family} and~\ref{tab:switchci} were repeated on 64 FineWeb-Edu documents, the common sixteen and the next 48 of the same split, with the outlier dimensions and layer means recomputed on the 64. Across the \NsFNModelsWord{} models the window means of $D$, $A + C - D$, the controlled cosine and $\bar D$ move by at most \NsFDGapMax, \NsFLawGapMax, \NsFCosGapMax{} and \NsFDbarGapMax, all \NsFNInsideSixteenInterval{} of the \NsFNQuantities{} values lie inside the sixteen-document intervals, and the half-widths halve (at most \NsFCosHwMax{} for the cosine against \NsFCosHwSixteenMax); the document-to-document spread of these window means within the corpus is small, and the sixteen-document intervals describe it.

\begin{table}[htbp]
\centering
\small
\caption{The controlled decomposition (window means) on sixteen WikiText-103 articles of 192 tokens beside the FineWeb-Edu values of Table~\ref{tab:switchci}.}
\label{tab:wiki}
\fittable{\input{supplement/table_wiki_r1}}
\end{table}

\paragraph{Damping.} The average damping $\bar D$ of Section~\ref{sec:setup} with one random direction per token, orthogonal to the MLP's input, $\|\epsilon u\| = 0.02\|x'\|$, the MLP evaluated in single precision (the forward pass of the 7B and 8B models and of Phi-3-mini runs in brain floating point, bfloat16, and each layer's normaliser and MLP are evaluated in single precision for the finite differences). $D_{\mathrm{prev}}$ uses the direction of $c_{l-1}$ projected orthogonal to $x'$ at the same tokens; the ratio of $D_{\mathrm{prev}}$ to $\bar D$ in Table~\ref{tab:family} is the mean over documents of the per-document ratio of window means, with its interval and Wilcoxon test over documents in the text.

\paragraph{Equation~\eqref{eq:switch} in the controlled frame.} The write $v$ is the controlled $c_b$ at each token. For each vector $y$ that is projected ($a_j$, $m_j$, and $m_j$ with $v$ removed), the mean of $y$ over all tokens of the sixteen documents, in the same condition, is subtracted before projecting on $\hat v$; because $\hat v$ is zero on the outlier dimensions, they need no separate zeroing. With these definitions the left side of Equation~\eqref{eq:switch} is the mean over tokens of the controlled projection $\|v\|^{-1}\langle \hat v, \tilde c_j\rangle$, where $\tilde c_j$ is the controlled $c_j$, and $A + C - D$, which is $A$ plus the MLP's projection at the actual input, equals it to within $\SwIdentityGap$ (bfloat16 rounding in the larger models). In the raw frame $v = c_b$ and nothing is subtracted. The split by neuron group (Table~\ref{tab:group}) fixes each neuron's group at the model's actual input, computes the four group outputs at the actual input and at the input with $v$ removed, and projects each the same way; the four parts therefore sum to $C$ and $D$ exactly.

\paragraph{Checks and conventions of Equation~\eqref{eq:switch}.} Two checks support reading $D$ as a property of the MLP rather than of how the write is removed: with $v$ scaled by $0.5$ or $1.5$ instead of removed, the MLP's projection deviates from the line through its values at $0$ and $1$ by a median of \SwLinErrMed\,$D$ (at most \SwLinErrMax\,$D$ in any model); and removing $v$ from the layer's input instead, so that attention is recomputed too, gives \SwLayerRatioMin{} to \SwLayerRatioMax{} times the same damping; and removing only the part of $v$ orthogonal to $x'$, which leaves the normaliser's scale unchanged, lowers $D$ to \PerpDRatioMin{} to \PerpDRatioMax{} of its value in the gated models and $C$ by the same amount, so that $A + C - D$ is unchanged to four decimals. The split between $A$ and $C$ depends on the counterfactual: $A$ keeps attention's own response to $v$, and with $v$ removed from the layer's input instead, GPT-2's attention term is $\SpAlayerOneGPTTwo$ against $\SpAOneGPTTwo$ (Table~\ref{tab:switchraw}). The decomposition locates GPT-2's exception in its attention and co-writing; it does not say why GPT-2 co-writes more. Only per-token signs of projection and cosine must agree; the window means agree in every model. In raw terms Phi-3-mini and ProSparse-7B sit on the other side of the boundary, because their attention writes along the shared component that the controls remove (Appendix~\ref{app:raw}).

\paragraph{Tests.} Documents are the units. Intervals are $t$ intervals over the sixteen document means; paired comparisons use the Wilcoxon signed-rank test over documents, whose smallest attainable two-sided $p$ with sixteen units is $3\times10^{-5}$, reached when all sixteen differences have the same sign. The per-token test computes, within each document, the Spearman correlation between the outcome and a predictor after regressing the ranks of both on the controls (pair indicators, $\log\|x'\|$, $\log$ position, $\log\|c_{j-1}\|$, $\log\|c_j\|$), and reports the mean over documents with a $t$ interval.

\begin{table}[htbp]
\centering
\small
\caption{Equation~\eqref{eq:switch} with raw vectors, adjacent layers, window means in units of $\|c_{j-1}\|$. $D$, layer: the damping with the write removed from the layer's input, so that attention is recomputed. Linearity: median over pairs of the largest deviation of the MLP's projection at $s = 0.5$ and $1.5$ from the line through $s = 0$ and $1$, in units of $D$.}
\label{tab:switchraw}
\fittable{\input{supplement/table_switch_raw_r1}}
\end{table}

\begin{figure}[htbp]
\centering
\input{figures/protocol}
\caption{The adjacent-layer cosine on the window, on our approximation of the protocol of \citet{patrawala2025tlcm} (\TpNArticles{} WikiText-103 test articles of 2{,}048 tokens) and on ours (sixteen FineWeb-Edu documents of 192 tokens): (a) raw, with the raw cosine averaged over the middle half of the layers instead of the window; (b) controlled. Models in the order of Figure~\ref{fig:law}.}
\label{fig:protocol}
\end{figure}

\begin{table}[htbp]
\centering
\small
\caption{The decomposition of Figure~\ref{fig:law}: $A$, $C$, $D$ and $A + C - D$ with 95\% intervals over the sixteen documents, and the controlled cosine (window means, per unit of the controlled $\|c_{j-1}\|$).}
\label{tab:switchci}
\fittable{\input{table_switch_r1}}
\end{table}

\section{Which tokens cancel}
\label{app:pertoken}

 The lobe term varies from token to token, with how many strengthening neurons sit in the negative region and how deep. If it carries the damping, tokens where it is larger should cancel more within a pair. For each document we compute the Spearman correlation, over tokens and window pairs, between the controlled projection of $c_j$ on $c_{j-1}$ and the lobe term of MLP $j$ at that token, partialling out the pair, $\log\|x'\|$ (the normaliser's factor in Equation~\eqref{eq:trace}), $\log$ position, and $\log\|c_{j-1}\|$ and $\log\|c_j\|$. In every gated model the partial correlation is negative, between $\PtLobePartialMin$ and $\PtLobePartialMax$ (negative in \PtLobeNegDocsMin{} of the sixteen documents of each model; 95\% intervals exclude zero in all ten; Figure~\ref{fig:pertoken}), and it stays between $\PtLobeTokMin$ and $\PtLobeTokMax$ when the token's class (punctuation, number), on which \citet{patrawala2025tlcm} find the cancellation firing most often, and its log frequency are partialled out as well (a token is punctuation if its decoded string, ignoring surrounding whitespace, is only non-word characters, and a number if it is a digit followed by digits, commas or periods; the frequency is the token's count, under the model's tokenizer, in the first two million characters of the first 400 held-out documents prepared as the common set). The weakening term's correlation changes sign between models ($\PtWeakPartialMin$ to $+\PtWeakPartialMax$), and the total damping of Equation~\eqref{eq:trace} predicts the projection at least as well as the lobe term in every model ($\PtTotalPartialMin$ to $\PtTotalPartialMax$). This is a correlation within trained models.

\section{Over training}
\label{app:training}

Figure~\ref{fig:switch} follows the terms of Equation~\eqref{eq:switch} for adjacent layers over the training of OLMo-1B, Pythia-160m, TinyLlama-1.1B and OLMoE-1B-7B at checkpoints released by their authors. In OLMo-1B (\OlN{} checkpoints from 2B to 3T tokens, doubling up to 513B and then every 250B to 500B; the released model measured in the tables is a later revision) the window cosine crosses zero between \OlPreCrossTok{}B and \OlCrossTok{}B tokens, and the sum $A + C - D$ between \OlLawPreCrossTok{}B and \OlLawCrossTok{}B, where the damping of the write ($\OlLawPreD$, then $\OlLawCrossD$) passes co-writing ($\OlLawPreC$, then $\OlLawCrossC$). At the cosine's crossing the lobe term is \OlCrossLobe{} and the median read--write cosine is at its minimum, $\OlRWMinVal$: the neurons are anti-aligned, and the average damping (\OlCrossDbar) comes from open weakening neurons (its weakening term is \OlCrossWeak), the route of the non-gated models. Over the following tokens the read--write cosine rises to $+\OlRWLast$, the lobe term to \OlLobeLast, and the open-side damping stays near \OlWeakLast; at 3T the closed side carries \RClosedSideShareOLMoOneB\% of the average damping and closed strengthening neurons \GrDClosedStrShareOLMoOneB\% of the write's. TinyLlama-1.1B takes the other route: at its earliest released checkpoint, 105B tokens, its neurons are already aligned (median read--write cosine $+\TlRWA$) and the lobe term is \TlLobeA{} against a weakening term of \TlWeakA, and at 2.5T, the last checkpoint before its cosine turns negative, the closed side already carries the damping (\TlLobeF{} against \TlWeakF); over the run its median read--write cosine rises from $+\TlRWFirst$ to $+\TlRWLast$ and the lobe term from \TlLobeFirst{} to \TlLobeLast. In both, co-writing stays nearly constant (between \PyCMin{} and \PyCMax{} in Pythia and between \TlCMin{} and \TlCMax{} in TinyLlama) and attention stays small (at most \PyAMax{} and \TlAMax), while the damping of the previous write grows from $\PyDwFirst$ to $\PyDwLast$ and from $\TlDwFirst$ to $\TlDwLast$. TinyLlama's trajectory carries a caveat: its damping rises by \TlDwStepFlip{} between the 2T and 2.5T checkpoints, against at most \TlDwStepMaxOther{} in any other interval, where its authors attribute a jump in benchmark scores to a data-loading issue (an issue with the beginning-of-sequence token described in its release notes); its window cosine turns negative between \TlFlipBefore{} and \TlFlipAfter{} billion tokens. We report the trajectory as released and draw the training-time conclusion, that the damping grows while co-writing stays flat, from OLMo-1B and Pythia-160m. OLMoE-1B-7B, the mixture of experts (\OeN{} checkpoints from 41B to 5.1T tokens; Figure~\ref{fig:switch}d), has the same shape and has not reached the crossing: co-writing stays between \OeCMin{} and \OeCMax{} and attention within \OeAAbsMax{} of zero, while the damping of the previous write grows from $\OeDwFirst$ to $\OeDwLast$, the lobe term from \OeLobeFirst{} to \OeLobeLast{} and the median read--write cosine from $\OeRWFirst$ to $+\OeRWLast$; the controlled cosine falls from $+\OeCosFirst$ to $+\OeCosLast$ and $A + C - D$ from $+\OeLawFirst$ to $+\OeLawLast$, positive at every checkpoint.

\begin{figure}[htbp]
\centering
\input{figures/switch}
\caption{The terms of Equation~\eqref{eq:switch} for adjacent layers over training (window, controlled): (a)~OLMo-1B, a SwiGLU model, over its released checkpoints from 2B to 3T tokens; (b)~Pythia-160m, which has no gate; (c)~TinyLlama-1.1B, a SwiGLU model; (d)~OLMoE-1B-7B, the mixture of experts, over its released checkpoints from 41B to 5.1T tokens. The dashed line is the controlled adjacent cosine. TinyLlama's released checkpoint at 503B tokens has a higher loss on our documents (\TlLossB{} nats) than those at 105B and 1T tokens (\TlLossA{} and \TlLossC), and its jump between 2T and 2.5T coincides with a documented data issue (text).}
\label{fig:switch}
\end{figure}

\section{Older writes}
\label{app:lags}

\citet{patrawala2025tlcm} find that layers further apart do not cancel on average. Equation~\eqref{eq:switch} applies to any lag. Figure~\ref{fig:lags} follows a write $v = c_b$ to layers $b+1$, $b+2$ and $b+3$, and splits co-writing into two parts: what the MLP writes along $v$ when the layers in between have also had their components along $v$ removed from its input (its \emph{own} co-writing), and the rest, its response to what the layers in between wrote along $v$. Older writes are still damped: from lag~1 to lag~2, in the ten gated models, the MLP's damping per unit of the write falls only to \LagDRatioMin{} to \LagDRatioMax{} of its lag-1 value, and its own co-writing changes by a factor of \LagCRatioMin{} to \LagCRatioMax. What changes is what the MLP finds: the layer in between has already removed part of the write (its projection on $v$ is between $\LagQMin$ and $\LagQMax$), so part of the damping now undoes that removal (the response term, $+\LagRintMin$ to $+\LagRintMax$); and attention, which writes against the previous write in \LagANegWord{} of the ten gated models ($A$ between $\LagANegMin$ and $\LagANegMax$), writes along the older one (its term rises by \LagDeltaAMin{} to \LagDeltaAMax). These changes and the weaker damping at lag~2 all move the balance towards reinforcement; no one of them accounts for most of the shift in every model. At lag~2 the controlled cosine lies between $\LagCosTwoMin$ and $+\LagCosTwoMax$ and is still negative in \LagNCosTwoNeg{} of the ten gated models; at lag~3 between $\LagCosThreeMin$ and $+\LagCosThreeMax$, negative in \LagNCosThreeNeg. The cancellation is concentrated in adjacent pairs, not confined to them, and the concentration needs no selectivity of the damping for the previous layer: later MLPs damp an older write nearly as much per unit, but by the time the write reaches them part of it is gone and attention writes more along it, while what they co-write along it has not shrunk.

\begin{figure}[htbp]
\centering
\input{figures/lags}
\caption{Equation~\eqref{eq:switch} at lags 1--3 for every model (window, controlled, document means), one line per model, coloured by family: (a) the controlled cosine, (b) the damping $D$, (c) attention $A$, (d) own co-writing, what the MLP of layer $b+k$ writes along $c_b$ with $c_b$ and the intermediate layers' components along it removed from its input.}
\label{fig:lags}
\end{figure}

\section{The full decomposition}
\label{app:decomp}

The last term of Equation~\eqref{eq:trace}, the part of the trace of $M$ along $x'$ itself, is
\begin{equation*}
\frac{\hat x^{\top} M \hat x}{d-1} = \frac{1}{(d-1)\|x'\|}\sum_k (\hat x\cdot d_k)(u_k\cdot z)\big[\phi'(t_k)\,t_k + \phi(t_k)\big].
\end{equation*}

\paragraph{A mixture of experts.} In OLMoE-1B-7B each layer's MLP is $m(x') = \sum_e p_e\, m_e(z)$ over the eight experts $e$ that a linear router selects for the token out of 64, where $p_e$ is the softmax probability of expert $e$ over all 64 (the selected probabilities are not renormalised) and $m_e$ is a SwiGLU MLP of width 1{,}024. The selection is piecewise constant in $x'$, so at almost every input $J = \sum_e p_e J_e + \sum_e m_e\,(\nabla p_e)^{\top}$: every neuron term of Equation~\eqref{eq:trace} is taken per expert and weighted by $p_e$, and the router adds one term, $-(d-1)^{-1}\sum_e \nabla p_e\cdot P m_e$, with $\nabla p_e = s\,P\big(w\odot p_e(q_e - \sum_f p_f q_f)\big)$ for router rows $q_e$ and the normaliser's scale $s$. The sum of all terms matches $-(d-1)^{-1}\tr(PJP)$ from the Jacobian by automatic differentiation to five decimals at the tokens we checked. The router term is negligible: \RRouterOLMoE{} in the window (\RRouterShareOLMoE\% of $\bar D$) and at most \RRouterAbsMaxOLMoE{} in any layer. OLMoE's residual stream is small in its early layers (mean squared coordinate $\overline{x'^2}$ near $10^{-4}$ against the normaliser's $\epsilon = 10^{-5}$), so for this model $s = (\overline{x'^2} + \epsilon)^{-1/2}$ replaces $\sqrt d/\|x'\|$ and the normaliser's projector is $I - \rho\,\hat x\hat x^{\top}$ with $\rho = \overline{x'^2}/(\overline{x'^2} + \epsilon)$; for the dense models the difference is below the precision reported. In the split of $C$ and $D$ by neuron group (Table~\ref{tab:group}) the routing is the model's own at each input, so the four groups sum to the MLP's actual output, and a neuron of an expert that acts at the token at the actual input keeps the gate state it has there; the local damping along fixed directions (Table~\ref{tab:selectivity}) holds the routing fixed. The read--write cosine and the strengthening share pool the neurons of all 64 experts, and the closed fraction is over the neurons of the experts that act at each token. (Gemma's normaliser multiplies by $1 + w$, which replaces $w$ throughout.)

\paragraph{gpt-oss-20b.} OpenAI's open-weight mixture \citep{openai2025gptoss} is measured as released: 24 layers of width 2{,}880, 32 experts of width 2{,}880 with four routed to each token and the router's softmax taken over the four, expert weights in 4-bit MXFP4 run with the released kernels, the post-trained model (no base release exists). Its expert differs from SwiGLU in three places, all kept: the gate pre-activation $t = g\cdot z + b_g$ is clamped at $7$ and $\phi(t) = t\,\sigma(1.702\,t)$; the linear input $u\cdot z + b_u$ is clamped to $[-7, 7]$ and offset by one, so a neuron writes $\phi(t)\,(u\cdot z + b_u + 1)\,d_k$; and each expert adds an output bias. The terms of Equation~\eqref{eq:trace} follow with $(u\cdot z + b_u + 1)$ in the gate-path term, zero slope where a clamp is active (fewer than $0.3\%$ of linear inputs and $0.003\%$ of gates in the window), and the top-four softmax in the router term; the expert weights are dequantised from the released blocks for the analytic terms, and the finite differences use the released forward pass. On the common documents (loss \GoLoss) the analytic sum is \GoDbarAnalytic{} against a finite-difference $\bar D$ of \GoDbar: the identity holds, and the average damping is smaller than in any of the sixteen models, all of it the lobe term (\GoLobe; without the closed side $\GoCf$), from a strengthening share of \GoStrSharePct\%, a closed fraction of \GoClosed{} and a median read--write cosine of $+\GoRW$, with a residual stream of norm \GoXnormMin{} to \GoXnormMax{} across the window. Adjacent layers write along each other at every one of its \GoNPairs{} pairs (controlled window cosine $+\GoCos$, raw $+\GoCosRaw$; $+\GoTpWikiRaw$ raw on the protocol of \citet{patrawala2025tlcm}): in the controlled decomposition at lag~1, $A = +\GoA$, $C = \GoC$ and $D = \GoD$ ($[\GoDLo, \GoDHi]$ over documents), so the MLP's response to removing the whole previous write is nil, although its local damping along the raw write is \GoDprev, \GoDprevRatio{} times a random direction's; here the finite response is not the local one. In raw terms the MLP writes \GoCRaw{} of the write's norm along it with the write absent and \GoDRaw{} less with it present. It is not counted among the sixteen models and has only the measurements stated here.

The columns of Table~\ref{tab:family} are: $\bar D$: the average damping of Section~\ref{sec:setup}. Lobe: the closed-strengthening term of Equation~\eqref{eq:trace} (non-gated models: neurons with negative slope and aligned output). No closed side: $D$ from Equation~\eqref{eq:trace} with every closed-gate term removed. Weakening: open (non-gated: positive-slope) neurons whose output opposes their input. The ratio of $D_{\mathrm{prev}}$ to $\bar D$: local damping along the previous layer's raw write (outlier dimensions and layer mean kept) relative to a random direction, same tokens (Appendix~\ref{sec:spectrum}); omitted (---) where $\bar D < 0.05$, as a ratio of two quantities near zero. Read--write: median unweighted cosine between a neuron's linear input weight and output weight (the strengthening class uses the gain-weighted inner product). Closed: fraction of gate activations that are closed (non-gated: negative pre-activations). ReGLU$^\ast$: ReLU gate with threshold $0.01$.

\begin{table}[htbp]
\centering
\small
\setlength{\tabcolsep}{3pt}
\caption{The values of Figure~\ref{fig:sources}a with the other per-model quantities (window means over sixteen FineWeb-Edu documents; 95\% half-widths at most \RFamHalfWidth). Columns as defined above.}
\label{tab:family}
\fittable{\input{table_family_r1}}
\end{table}

\begin{table}[htbp]
\centering
\small
\setlength{\tabcolsep}{3pt}
\caption{The values of Figure~\ref{fig:sources}b, and the same split of the co-writing $C$ (window, controlled; each neuron's group fixed by its state at the actual input). The parts sum to $D$ and $C$ of Table~\ref{tab:switchci}. Document means; 95\% half-widths at most \GrHalfWidth.}
\label{tab:group}
\fittable{\input{table_group_r1}}
\end{table}

Table~\ref{tab:decomp} gives all terms of Equation~\eqref{eq:trace} for the gated models, including the term along $x'$, and of its LayerNorm form for the non-gated ones. For a non-gated MLP $m(x') = \sum_k \phi(w_k\cdot z + b_k)\,d_k$ with LayerNorm $z = \gamma\odot\tilde x + \beta$, $\tilde x = \sigma^{-1}(x' - \mu)$, the Jacobian of $z$ with respect to $x'$ is $\mathrm{diag}(\gamma)\,\sigma^{-1}\big(I - \tfrac1d\mathbf{1}\mathbf{1}^\top - \tfrac1d\tilde x\tilde x^\top\big)$, so neuron $k$ contributes
\begin{equation*}
\tr J_k = \frac{\phi'(a_k)}{\sigma}\Big[\langle d_k, \gamma\odot w_k\rangle - \tfrac1d\big((w_k\cdot\gamma)(\mathbf 1\cdot d_k) + (w_k\cdot(\gamma\odot\tilde x))(\tilde x\cdot d_k)\big)\Big],
\end{equation*}
and $\bar D = -(d-1)^{-1}\sum_k \tr J_k$ exactly, since this Jacobian annihilates $x'$ (which is $\mu\mathbf 1 + \sigma\tilde x$) and so equals its own projection. An active neuron ($\phi'(a_k) > 0$) whose output weight opposes its input weight ($\langle d_k,\gamma\odot w_k\rangle < 0$) therefore adds damping. The ``lobe term'' of Table~\ref{tab:family} for these models sums the neurons with $\phi'(a_k) < 0$ and $\langle d_k,\gamma\odot w_k\rangle > 0$; the weakening term sums those with $\phi'(a_k) > 0$ and $\langle d_k,\gamma\odot w_k\rangle < 0$.

\begin{table}[htbp]
\centering
\small
\setlength{\tabcolsep}{3pt}
\caption{Window terms of the trace decomposition. Gated models (top): through the gate, $-\phi'(g\cdot z)(u\cdot z)\langle w\odot d, g\rangle$, and through the linear input, $-\phi(g\cdot z)\langle w\odot d, u\rangle$, split by gate state and by the sign of $\langle w\odot d, u\rangle$ (strengthening or weakening), and the term along $x'$ (radial). Non-gated models (bottom): split by the sign of the activation's slope and of $\langle d, \gamma\odot w\rangle$. Each row sums to $D$ up to finite-difference noise.}
\label{tab:decomp}
\fittable{\input{supplement/table_decomp_r1}}
\end{table}

\begin{table}[htbp]
\centering
\small
\setlength{\tabcolsep}{3pt}
\caption{The local damping along the previous write (a finite difference along $c_{j-1}$ projected orthogonal to $x'$) and along a random direction orthogonal to $x'$, on the same tokens, split by neuron group with each neuron's group fixed at the model's actual input (window, document means).}
\label{tab:grouplocal}
\fittable{\input{supplement/table_group_local_r1}}
\end{table}

\begin{figure}[htbp]
\centering
\input{figures/spectrum}
\caption{The spectrum of damping across directions: minus the eigenvalues of the symmetric part of the MLP's Jacobian on the subspace orthogonal to $x'$ (four layers of the window, one token per document). (a) The 5th to 95th percentile over directions, the mean (the average damping $\bar D$), and the mean damping of the lobe part of the Jacobian and of everything else. (b) The fraction of directions damped by more than $0.8$ (nearly removed), damped by less, and amplified (negative damping).}
\label{fig:spectrum}
\end{figure}

\section{The raw cosine and outlier dimensions}
\label{app:raw}
\citet{patrawala2025tlcm} report the Microsoft Phi models as not cancelling, naming Phi-3. In Phi-3-mini the gated mechanism is present (Table~\ref{tab:family}), and its window controlled cosine is $\RCosPhiThreeMini$ (standardised: $\RCosStdPhiThreeMini$), but its raw cosine is $+\RCosRawPhiThreeMini$. Removing each layer's mean contribution alone gives $\RCosMeanRemPhiThreeMini$; zeroing one, two, five and twenty outlier dimensions without that gives $+\RCosTopOnePhiThreeMini$, $+\RCosTopTwoPhiThreeMini$, $+\RCosTopFivePhiThreeMini$ and $\RCosTopTwentyPhiThreeMini$, so it is the per-layer mean, a component shared by all tokens, that makes Phi-3-mini's raw cosine positive. Gemma-2B's raw cosine ($\RCosRawGemmaTwoB$) is shrunk by a single dimension: zeroing it gives $\RCosTopOneGemmaTwoB$. In ProSparse-7B the raw cosine ($\RCosRawProSparse$) comes from a shared component of the opposite sign: removing each layer's mean gives $+\RCosMeanRemProSparse$, while zeroing twenty dimensions leaves $\RCosTopTwentyProSparse$ (Section~\ref{sec:interventions}). Across all models, the standardised cosine of \citet{timkey2021rogue} agrees in sign with the controlled one (Table~\ref{tab:cosctl}).

\section{The damping across directions}
\label{sec:spectrum}

$\bar D$ is the average damping over directions. To see how it varies from direction to direction, we computed the full Jacobian of the MLP at one token per document in four layers of the window of each gated model and took the eigenvalues of its symmetric part on the subspace orthogonal to $x'$; minus each eigenvalue is the damping along the corresponding direction (Figure~\ref{fig:spectrum}). The damping is not uniform across directions. Across the ten models its 5th percentile over directions is between $\SpecQFiveMin$ and $\SpecQFiveMax$ and its 95th between $+\SpecQNinetyFiveMin$ and $+\SpecQNinetyFiveMax$; \SpecContractMin{} to \SpecContractMax{} of directions are contracted and the rest expanded, so a feature along an expanded direction is amplified, not removed. \citet{patrawala2025tlcm} see the same breadth in the Jacobian of the whole layer: about three quarters of its eigenvalues are negative, and the fifty eigenvectors that carry most of the previous layer's write have eigenvalues clustered near $-1$ (fully corrected) and $+0.5$ (reinforced), a bimodal distribution they find inconsistent with uniform attenuation. The lobe part of the Jacobian carries the mean. On its own it contracts \SpecLobeContractMin{} to \SpecLobeContractMax{} of directions, with mean damping between \SpecLobeMeanMin{} and \SpecLobeMeanMax, while the rest of the Jacobian has mean damping within \SpecRestAbsMaxSwi{} of zero in the six SwiGLU models other than Phi-3 and only spreads it; in Gemma-2B and Phi-3-mini, where weakening neurons add damping, its mean is $+\SpecRestMeanGemmaTwoB$ and $+\SpecRestMeanPhiThreeMini$.

\paragraph{The previous write against a random direction.} Two versions of the previous write are compared with a random direction on the same tokens, all projected orthogonal to $x'$. Along the controlled write (Table~\ref{tab:grouplocal}), the ratio of the local damping along it to that along a random direction is \CtlPrevLlamaThreeOneB{} (\CtlPrevLoLlamaThreeOneB--\CtlPrevHiLlamaThreeOneB) in Llama-3.2-1B, \CtlPrevQwenHalfB{} (\CtlPrevLoQwenHalfB--\CtlPrevHiQwenHalfB) in Qwen2.5-0.5B, \CtlPrevTinyLlama{} (\CtlPrevLoTinyLlama--\CtlPrevHiTinyLlama) in TinyLlama-1.1B, \CtlPrevLlamaTwoSevenB{} (\CtlPrevLoLlamaTwoSevenB--\CtlPrevHiLlamaTwoSevenB) in Llama-2-7B, \CtlPrevMistralSevenB{} (\CtlPrevLoMistralSevenB--\CtlPrevHiMistralSevenB) in Mistral-7B, \CtlPrevLlamaThreeEightB{} (\CtlPrevLoLlamaThreeEightB--\CtlPrevHiLlamaThreeEightB) in Llama-3.1-8B-Instruct, \CtlPrevGemmaTwoB{} (\CtlPrevLoGemmaTwoB--\CtlPrevHiGemmaTwoB) in Gemma-2B, \CtlPrevPhiThreeMini{} (\CtlPrevLoPhiThreeMini--\CtlPrevHiPhiThreeMini) in Phi-3-mini \CtlPrevOLMoOneB{} (\CtlPrevLoOLMoOneB--\CtlPrevHiOLMoOneB) in OLMo-1B and \CtlPrevOLMoE{} (\CtlPrevLoOLMoE--\CtlPrevHiOLMoE) in OLMoE-1B-7B (95\% intervals over sixteen documents); it is above one in all sixteen documents of \CtlPrevNAbove{} models, in \CtlPrevNMistralSevenB{} of sixteen for Mistral-7B and in none for Gemma-2B. Along the raw write (Table~\ref{tab:family}), which keeps the five outlier dimensions and each layer's mean component that the controls remove, the ratio is above one in all sixteen documents of \PrevRatioNAbove{} models, by \PrevRatioAboveMin\% to \PrevRatioAboveMax\% (interval ends from \PrevRatioLoMin\% to \PrevRatioHiMax\%), in Pythia-160m by \PrevRatioExcessPythia\%, and in Gemma-2B the raw write is damped \PrevRatioGemmaLo\% to \PrevRatioGemmaHi\% less; the two ratios measure different directions, and the raw one is the larger. A Wilcoxon test on sixteen paired documents has $p = 3\times10^{-5}$ as its smallest value, which every all-sixteen case attains; a Bonferroni correction over ten models leaves those at $3\times10^{-4}$. A uniform random direction has expected squared cosine $d^{-1}$ with any neuron's read direction, so this comparison tests only whether the write is damped more than an arbitrary direction; the next paragraph tests it against directions of the write's own kind.

\paragraph{The previous write against other layer writes.} Table~\ref{tab:selectivity} gives the local damping (membership of each neuron fixed at the actual input, summed over the four groups) along six unit directions orthogonal to $x'$ at every token of the window pairs: the token's own controlled previous write; the same layer's controlled write at another token of the document, drawn uniformly over the other positions; the same layer's controlled write at the same position of the next document; the same layer's attention write $a_j = x'_j - x_j$ with the outlier dimensions zeroed and its token mean removed; the token's own write with its coordinates permuted (one permutation per layer, so the norm is kept and the direction is not); and a Gaussian direction. In every gated model the other token's write is damped more than the token's own previous write (ratio of the write's damping to the other token's between \SelRatioPrevOtherMin{} and \SelRatioPrevOtherMax; the write is damped more than the other token's write in at most \SelNPrevAboveOtherMax{} of sixteen documents in any model, and in none in \SelNModelsOtherZeroWord{} of the ten), and so is the other document's write (ratio \SelRatioPrevOdocMin{} to \SelRatioPrevOdocMax). The other token's write exceeds the random direction by \SelOtherRandMin\% to \SelOtherRandMax\% and the other document's by \SelOdocRandMin\% to \SelOdocRandMax\%: layer writes as a class overlap the neurons' read directions more than a uniform direction does, and the token's own previous write sits at the low end of that class. The shuffled write is damped as a random direction is (within \SelShufRandAbsMax\%), the write two layers back as the previous write (within \SelLagTwoAbsMax\%), and the attention write of the same layer between \SelRatioPrevAttnMin{} and \SelRatioPrevAttnMax{} times as much as the previous write. The same table gives the share of MLP $j$'s neurons whose gate state differs between $x'$ and $x' - v$, the neurons whose group the counterfactual of Equation~\eqref{eq:switch} moves: \SelFlipMin\% to \SelFlipMax\% of neurons, carrying \SelFlipWMin\% to \SelFlipWMax\% of the activation mass $\sum_k|\phi(t_k)(u_k\cdot z)|$ at $x'$.

\begin{table}[htbp]
\centering
\small
\caption{The previous write against in-distribution directions (window pairs, means over tokens and documents). Local damping $-u^{\top}Ju$ along six unit directions $u$ orthogonal to $x'$; write : other is the per-document ratio of the write's damping to the other token's (documents in which the write is damped more, of sixteen); write : random the same against the random direction. Gate flips: the percentage of neurons whose gate state the removal of the write changes (in parentheses, weighted by activation mass).}
\label{tab:selectivity}
\fittable{\input{supplement/table_selectivity_r1}}
\end{table}

\paragraph{The write's energy over the eigendirections of the MLP's Jacobian.} \citet{patrawala2025tlcm} infer selectivity from the eigenvalues of the whole layer's symmetrised Jacobian weighted by the previous write: the fifty eigenvectors that carry most of the write have eigenvalues clustered near $-1$ and $+0.5$, which they find inconsistent with uniform attenuation. We ask the same of the MLP's Jacobian. At the four window layers and one token per document used for Figure~\ref{fig:spectrum}, with $S$ the symmetric part of $PJP$ and $(\lambda_i, q_i)$ its eigenpairs on the subspace orthogonal to $x'$, a unit direction $c$ has energies $e_i = \langle q_i, c\rangle^2$ summing to one, and its damping $-c^{\top}Sc = \sum_i e_i(-\lambda_i)$ is the energy-weighted mean of the damping over eigendirections. Table~\ref{tab:eigen} compares the previous write, the same layer's write at another token, and a random direction. The participation ratio $1/\sum_i e_i^2$, which is $d - 1$ for energy spread evenly and about $(d-1)/3$ for a Gaussian direction, is \EigPrevPRFracMin{} to \EigPrevPRFracMax{} of $d - 1$ for the previous write against \EigOtherPRFracMin{} to \EigOtherPRFracMax{} for another token's write and \EigRandPRFracMin{} to \EigRandPRFracMax{} for a random direction; the write's fifty largest-energy eigendirections hold \EigTopFiftyPrevMin\% to \EigTopFiftyPrevMax\% of it, against \EigTopFiftyRandMin\% to \EigTopFiftyRandMax\% for a random direction (the ranges follow $d$, which runs from \EigDimMinusOneQwenHalfB{} to \EigDimMinusOneLlamaThreeEightB{} plus one). The energy on eigendirections damped by more than $0.8$ is \EigFullPrevMin\% to \EigFullPrevMax\% for the write against \EigFullRandMin\% to \EigFullRandMax\% for a random direction, and on expanded eigendirections \EigAmpPrevMin\% to \EigAmpPrevMax\% against \EigAmpRandMin\% to \EigAmpRandMax\%; in the two tails of the spectrum together (damping above $1.2$ or below $-0.5$) the write has \EigTailsPrevMin\% to \EigTailsPrevMax\% of its energy against \EigTailsRandMin\% to \EigTailsRandMax\%, and in the middle of the spectrum (damping between $0.1$ and $0.8$) \EigMidPrevMin\% to \EigMidPrevMax\% against \EigMidRandMin\% to \EigMidRandMax\%. The write is somewhat more concentrated than a random direction and places about twice the energy in both tails, the most contracted and the most expanded eigendirections alike; it does not concentrate on a few strongly contracting eigendirections. The lobe part of the Jacobian, $J_{\mathrm{lobe}}$ of Figure~\ref{fig:spectrum}, damps the write and a random direction within \EigLobeQGapPctMax\% of each other, and the rest of the Jacobian damps the write by at most \EigRestQAbsMax{} in magnitude in any model. The whole layer's Jacobian is treated next.

\paragraph{The whole layer's Jacobian.} \citet{patrawala2025tlcm} (their Section~5.2) linearise layer $i + 1$ at a token, $J = \partial c_{i+1}(t) / \partial x_{i+1}(t)$ with the prefix fixed, symmetrise it, project the previous layer's raw write $a = c_i(t)$ on the eigenvectors, and histogram the eigenvalues of the fifty eigendirections that carry most of $|{\bf Q}^{\top} a|^2$ over fifty token--layer pairs of Llama-3.1-8B-Instruct on one fixed text where the per-token cosine between $c_i$ and $c_{i+1}$ lies in $[-0.3, -0.1]$ (``TLCM'' pairs) and fifty where it lies within $0.015$ of zero; they find the first group bimodal, with modes near $-1$ and $+0.5$, and the second peaked at zero, and conclude that the correction is selective. We repeat this with their released code's text and pair selection (window pairs $4 \le i < 20$, tokens up to 249, Jacobians by reverse-mode differentiation of the layer in single precision), and add to each pair a random direction and the same layer's write at another token of the text, passed through the same Jacobian; a third group of pairs drawn from the window with no condition on the cosine; and the split of $J$ into the attention sublayer's Jacobian and the remainder (the MLP's path, including its response to attention's response). Table~\ref{tab:layerjac} gives, for each group and model, the share of the fifty eigenvalues below $-0.75$ and above $+0.25$ for the three directions and the first-order projection $c^{\top}\bar J c$ along each unit direction, which by their Appendix~B is the sign that decides correction. On their model and pairs the previous write's fifty eigenvalues are bimodal as they report, with \LjTlcmPrevTopBelowLlamaThreeEightB\% below $-0.75$ and \LjTlcmPrevTopAboveLlamaThreeEightB\% above $+0.25$ against \LjTlcmRandTopBelowLlamaThreeEightB\% and \LjTlcmRandTopAboveLlamaThreeEightB\% for a random direction; the same layer's write at another token has \LjTlcmOtherTopBelowLlamaThreeEightB\% and \LjTlcmOtherTopAboveLlamaThreeEightB\%. The first-order projection is $\LjTlcmPrevWmeanLlamaThreeEightB$ for the previous write, $\LjTlcmOtherWmeanLlamaThreeEightB$ for another token's write and $\LjTlcmRandWmeanLlamaThreeEightB$ for a random direction; across the \LjTlcmNModelsWord{} gated models run on their text the previous write's projection lies between $\LjTlcmPrevWmeanMin$ and $\LjTlcmPrevWmeanMax$, another token's between $\LjTlcmOtherWmeanMin$ and $\LjTlcmOtherWmeanMax$, and a random direction's between $\LjTlcmRandWmeanMin$ and $\LjTlcmRandWmeanMax$, the previous write contracted more than another token's write in \LjTlcmNPrevWmeanBelowOther{} of \LjTlcmNModels{} models and less in the rest, and more than a random direction in \LjTlcmNPrevWmeanBelowRand. Attention's part of the previous write's projection is at most \LjTlcmAttnQuadAbsMax{} in magnitude; the MLP's path carries $\LjTlcmMlpQuadMin$ to $\LjTlcmMlpQuadMax$. At the pairs selected for a cosine within $0.015$ of zero, the same statistics are smaller for all three directions alike (first-order projections $\LjNoTlcmPrevWmeanLlamaThreeEightB$, $\LjNoTlcmOtherWmeanLlamaThreeEightB$ and $\LjNoTlcmRandWmeanLlamaThreeEightB$ on their model): their two groups differ in how much the layer's Jacobian contracts at that token, not in whether it singles out the previous write, and since by their Appendix~B the per-token cosine is, to first order, the sign of $a^{\top}\bar J a$, selecting pairs by it selects Jacobians that contract $a$. The signature from which they conclude that the correction is selective, energy on eigenvalues near $-1$ alongside energy on positive eigenvalues, is the signature of a layer write of that layer at any token, and of a random direction in attenuated form; it places the previous write in the class of layer writes, and the MLP's part of it is the damping of Section~\ref{sec:mechanism}.

\begin{table}[htbp]
\centering
\small
\caption{The whole-layer Jacobian test of \citet{patrawala2025tlcm} with two comparison directions. Pairs: token--layer pairs on their text selected by the per-token cosine between $c_i$ and $c_{i+1}$ (their TLCM and non-TLCM groups) or drawn from the window without condition; $n$ pairs per group. For each pair, the fifty eigendirections of the symmetrised layer Jacobian carrying most of a direction's energy: the shares of their eigenvalues below $-0.75$ and above $+0.25$ (means over pairs), and the first-order projection $c^{\top}\bar J c$ (negative: contraction), for the previous write, the same layer's write at another token, and a random direction. ``Common documents'' uses the paper's sixteen documents in place of their text.}
\label{tab:layerjac}
\fittable{\input{supplement/table_layerjac_r1}}
\end{table}

\begin{table}[htbp]
\centering
\small
\caption{The write's energy over the eigendirections of the MLP's symmetrised Jacobian (four window layers, one token per document). Participation ratio of the energies as a fraction of $d - 1$ (a Gaussian direction gives about $1/3$), and the share of the energy on eigendirections damped by more than $0.8$ and on expanded ones, for the previous write, the same layer's write at another token, and a random direction.}
\label{tab:eigen}
\fittable{\input{supplement/table_eigen_r1}}
\end{table}

\section{Inference-time interventions and converted models}
\label{app:inference}

\paragraph{At inference, in the pretrained models.} Removing the closed side from every MLP of the ten gated models at once brings the window damping within \LoDampAbsMax{} of zero in all of them, but the models stop working as language models (loss \LoLossMin{} to \LoLossMax{} nats); a size-matched control that zeroes open entries at random also breaks them (loss \OcLossMin{} to \OcLossMax) yet leaves the damping at \OcDampRatioMin{} to \OcDampRatioMax{} of its value and the cosine negative in \OcNCosNeg{} of them (Figure~\ref{fig:lobeoff}). Among broken models it is the closed side, not the amount of activation removed, that carries the damping; broken models say nothing about cancellation in working ones. A gentler intervention, CATS-style thresholding \citep{lee2024cats} that zeroes the 50\% of gate activations smallest in magnitude (thresholds set on separate calibration text), keeps the models working; \CatsFiveZeroedClosedMin\% to \CatsFiveZeroedClosedMax\% of the entries it zeroes are closed, but only the smallest closed ones, \CatsFiveMassMin\% to \CatsFiveMassMax\% of the closed side's activation mass. The damping falls to \CatsFiveDampRatioMin{} to \CatsFiveDampRatioMax{} of its value and the cosine rises in all \CatsNModels{} models, at a loss cost of \CatsFiveDLossMin{} to \CatsFiveDLossMax{} nats. Damping and cancellation move together as the closed side is removed, and every step also degrades the model. Table~\ref{tab:catsctl} separates the two: in the same run, open entries ($t > 0$) are zeroed independently at random at rates $p$ from $0.5\%$ to $30\%$, the closed side left intact, and the rate whose loss matches the model's CATS loss is read off by interpolation ($p$ between \CcCtlPMin\% and \CcCtlPMax\%). At that loss the control leaves the damping at \CcCtlDRatioMin{} to \CcCtlDRatioMax{} of its value and moves the cosine by at most \CcCtlCosDeltaAbsMax{} (in \CcNCtlCosRises{} of \CcNModels{} models upward), while CATS lowers the damping to \CcCatsDRatioMin{} to \CcCatsDRatioMax{} and raises the cosine by \CcCatsCosDeltaMin{} to \CcCatsCosDeltaMax{}; in all \CcNModelsWord{} models the CATS cosine rise exceeds the control's. What moves the damping and the cosine is the removal of the closed side's small entries, not the loss cost: a graded association, not a clean removal.

\begin{table}[htbp]
\centering
\small
\caption{A loss-matched control for CATS (window means on the common documents). CATS at 50\% zeroes the gate activations smallest in magnitude, mostly closed ones; the control zeroes open entries at random at the rate $p$ whose loss matches the CATS loss (interpolated between the rates run). $\bar D$ ratio: window average damping relative to the unmodified model; $\Delta\cos$: change of the controlled window cosine.}
\label{tab:catsctl}
\fittable{\input{supplement/table_catsctl_r1}}
\end{table}

\paragraph{In converted 7B models.} Two public models start from Llama-2-7B's weights, replace the gate's SiLU by a ReLU, and retrain: ReluLLaMA-7B on about 5B tokens with language-modelling and distillation losses \citep{relullama}, ProSparse-7B (threshold $0.01$) on 34.6B tokens including instruction data, with a sparsity regulariser \citep{song2025prosparse}. Both remain working language models (losses \RLossTwoReluLlama{} and \RLossTwoProSparse{} against \RLossTwoLlamaTwoSevenB), and they keep \DampKeepReluLlama\% and \DampKeepProSparse\% of Llama-2-7B's window damping (Table~\ref{tab:family}): their strengthening neurons survive (median read--write cosines $+\RRWReluLlama$ and $+\RRWProSparse$ against $+\RRWLlamaTwoSevenB$), but a closed ReLU gate contributes exactly zero. Paired over the sixteen documents, the window damping changes by $\PairDampReluLlama$ $[\PairDampReluLlamaLo, \PairDampReluLlamaHi]$ and $\PairDampProSparse$ $[\PairDampProSparseLo, \PairDampProSparseHi]$ and the controlled cosine by $+\PairCosReluLlama$ and $+\PairCosProSparse$ (every document in the same direction). The confounds limit this: the conversions differ from their parent in distillation or in instruction data and a regulariser; the parent's own window cancellation is weak (controlled $\RCosLlamaTwoSevenB$); and ProSparse cancels in raw terms throughout ($\RCosRawProSparse$) through its attention, which writes against a component shared by all tokens while its MLP damps within \PsLateCtlDAbsMax{} of zero (Appendix~\ref{app:raw}), a route neither Llama-2-7B nor ReluLLaMA-7B shows and whose origin we leave open.

\begin{figure}[htbp]
\centering
\input{figures/lobeoff}
\caption{Every MLP changed at once, at inference: the closed side removed, $\phi(t)\to\phi(t)\,\mathbf 1[t>0]$, or open entries zeroed at random with probability equal to the closed side's share of each token's $\sum|\phi|$ (size-matched control). (a) Window damping, (b) window controlled cosine, (c) loss on the sixteen documents. Both changes stop the models working as language models.}
\label{fig:lobeoff}
\end{figure}

\begin{figure}[htbp]
\centering
\input{figures/pertoken}
\caption{Per-token test: partial Spearman correlation between the controlled projection of $c_j$ on $c_{j-1}$ and each term of MLP $j$ at the same token (controlling for pair, $\log\|x'\|$, $\log$ position, $\log\|c_{j-1}\|$ and $\log\|c_j\|$, and for the lobe term also for punctuation, number class and log unigram frequency), mean over sixteen documents with 95\% intervals. A negative value means tokens with more of the term cancel more.}
\label{fig:pertoken}
\end{figure}

\section{Training the 51M models}
\label{app:train}

A Llama-style decoder with width 512, 8 layers, 8 heads, context 1{,}024 and a gated MLP of width 1{,}408, with pre-RMSNorm, rotary positions with base $10^4$, no biases, and tied input and output embeddings: 51.5M parameters, of which 25.8M are the embedding. Initialisation $\mathcal N(0, 0.02^2)$, with the attention output and MLP down-projection scaled by $\tfrac14$, the inverse square root of twice the number of layers. AdamW with $\beta=(0.9, 0.95)$, weight decay $0.1$ on matrices only, gradient clipping at 1, peak learning rate $10^{-3}$ with 2\% linear warm-up and cosine decay to zero, batches of $128\times1{,}024$ tokens, 7{,}629 steps (1.0B tokens), mixed precision in bfloat16. Data: shards 000--001 of FineWeb-Edu sample-10BT tokenized with GPT-2's BPE, 1.49B training and 15.4M validation tokens. For a given seed all arms share initial weights and data order. The window for these eight-layer models is layers 2--5. At eight points during training we record validation loss and the window damping and controlled cosine on sixteen validation sequences of 1{,}024 tokens. The final models are measured afterwards on sixteen other validation sequences of 1{,}024 tokens, with validation loss on 160 sequences. Each run takes 57 minutes on one RTX~5090. In the SwiGLU run the strengthening neurons' alignment $\cos(w\odot d_k, u_k)$ rises from \GrFirstAlign{} at \GrFirstTok{}B tokens to \GrMidAlign{} at \GrMidTok{}B and \GrLastAlign{} at \GrLastTok{}B (seed~1, window layers), and the lobe term from \GrFirstLobe{} to \GrLastLobe.

\begin{figure}[htbp]
\centering
\input{figures/arms}
\caption{Training from scratch without the gate's negative region. Each row is a 51M-parameter SwiGLU model trained for 1B tokens that differs only in the closed side of the gate activation, $\phi(t)$ for $t \le 0$ (the open side is SiLU in every row): scaled by $\lambda$ in both the forward pass and the gradient, kept in the forward pass with no gradient through it, or removed from the forward pass with SwiGLU's gradient (straight-through). Two seeds per row. (a) The average damping $\bar D$ (gray) and its lobe term (blue). (b) The fraction of gates closed. (c) The median cosine between the neurons' input and output weights. (d) Validation loss. Without the closed side ($\lambda = 0$) the damping falls to \ArmLamZeroShare\% of SwiGLU's at a loss the seeds cannot distinguish; none of these models cancels. Values in Table~\ref{tab:arms}.}
\label{fig:arms}
\end{figure}

\begin{table}[htbp]
\centering
\small
\caption{The values of Figure~\ref{fig:arms} (seeds 0 and 1). Read--write: median unweighted cosine between a neuron's linear input and output weights. Str.\ share: fraction of the window's neurons with $\langle w\odot d_k, u_k\rangle > 0$. Closed: fraction of gate activations that are closed.}
\label{tab:arms}
\fittable{\input{table_arms_r1}}
\end{table}

\paragraph{The arms of Figure~\ref{fig:arms}.} The closed side of the gate activation, $\phi(t)$ for $t \le 0$, is scaled by $\lambda$ in both the forward pass and the gradient ($\lambda \in \{0, 0.5, 1, 2\}$); or kept in the forward pass with its output treated as a constant in the backward pass (no gradient through the closed side); or removed from the forward pass with the gradient of SwiGLU (a straight-through estimator, a training-time probe rather than a model one would deploy). The analysis was written down, with its date, before the first arm was trained: the elasticity per seed,
\begin{equation*}
e = \frac{\ln \bar D_{\lambda=2} - \ln \bar D_{\lambda=0.5}}{\ln 4},
\end{equation*}
with $e \le 0.3$ in both seeds to be read as the damping held near a level and $e \ge 0.8$ as the damping following the closed side's strength, anything between as partial compensation with no claim; and a two-by-two reading of the median read--write cosine over the cells (forward closed side present or absent) $\times$ (gradient through it present or absent), an effect counting only when it has the same sign in both seeds and is at least three times the largest within-cell difference between seeds. The elasticities were \ArmElasticityZero{} and \ArmElasticityOne{}; both the forward and the gradient effect on the alignment met the rule, with the pattern that the alignment is present only in the cell that has both.

\paragraph{A larger run.} To reach the regime in which Pythia-160m begins to cancel (between steps \PyFlipBefore{} and \PyFlipAfter, about 8B to 17B tokens at its batch size), we trained a SwiGLU model of Pythia-160m's shape, 12 layers of width 768 with 12 heads (125M parameters with GPT-2's vocabulary), for 12B tokens with batches of $512\times1{,}024$ tokens and peak learning rate $6\times10^{-4}$, otherwise as above. The batches are random windows of the same 1.49B-token training set, so this run saw its data about eight times over, which the 51M runs and Pythia did not. It does not cancel, though with eight passes over the data the run does not separate the number of unique tokens from repetition: at 12B tokens the window damping is \CzSwiD{} (lobe term \CzSwiLobe, median read--write cosine $+\CzSwiRw$) and the controlled cosine $+\CzSwiCos$, at a validation loss of \CzSwiVal; in the training-time probe the cosine fell from $+\CzSwiProbeCosMax$ to $+\CzSwiProbeCosLast$ over the run while the damping rose to \CzSwiProbeGainLast, which is the shape of Figure~\ref{fig:switch} before its crossing. Its $\lambda = 0$ twin, trained on the same batches from the same initial weights, does not cancel either (controlled cosine $+\CzLamZeroCos$): it has \CzLamZeroShareOfSwi\% of the SwiGLU arm's damping (\CzLamZeroD{} against \CzSwiD; lobe term zero, weakening term \CzLamZeroWeak), again without the strengthening alignment (share \CzLamZeroStr{} against \CzSwiStr, median read--write cosine $\CzLamZeroRw$ against $+\CzSwiRw$), at a validation loss of \CzLamZeroVal{} against \CzSwiVal{} (one seed; the difference, \CzLossDiff{} nats, is within the seed-to-seed spread of the 51M runs). The closed side's share of the damping is larger at this budget than at 51M (\ArmLamZeroShare\% there).

\section{Continued pretraining with the closed side annealed away}
\label{app:contpre}

\paragraph{Setup.} Qwen2.5-0.5B (three seeds) and Llama-3.2-1B (one seed) are trained for \CpSteps{} steps of \CpBatchTokensK{}k tokens (\CpTokensM M tokens): sequences of 1{,}024 tokens drawn at random from shard 002 of the FineWeb-Edu sample-10BT subset, tokenised with the model's tokenizer (the common documents come from shards 000--001), with AdamW ($\beta = (0.9, 0.95)$, weight decay $0.1$ on matrices only, gradient clipping at 1), peak learning rate $\CpLr$ after 100 warm-up steps with cosine decay to a tenth of the peak, bfloat16 autocast over single-precision weights, in two arms per seed from the same weights and the same sequence of batches (the seed sets the order of the batches). In the annealed arm every gated MLP's activation is $\phi(t)$ for $t > 0$ and $\lambda\,\phi(t)$ for $t \le 0$, in the forward pass and the gradient, with $\lambda = 1$ for the first \CpAnnealStart{} steps, falling linearly to $0$ at step \CpAnnealEnd{} (\CpAnnealEndM M tokens) and $0$ thereafter; the control arm keeps $\lambda = 1$. Validation loss is measured every 400 steps on 64 held-out sequences of 1{,}024 tokens from the same shard. The checkpoints at steps 0, 800, 1{,}600, 2{,}400, 3{,}200 and \CpSteps{} are measured as the released models are (the controlled decomposition at lag~1 of Table~\ref{tab:switchci}; the average damping and its split by neuron group of Tables~\ref{tab:family} and~\ref{tab:group}), each at its own $\lambda$. Table~\ref{tab:contpre} lists the values plotted in Figure~\ref{fig:contpre}, and Table~\ref{tab:contpre-llama} and Figure~\ref{fig:contpre-llama} those of Llama-3.2-1B. The two arms of one seed take about \CpQwenHalfBHours{} hours for Qwen2.5-0.5B and \CpLlamaThreeOneBHours{} for Llama-3.2-1B on one RTX~5090.

\begin{table}[htbp]
\centering
\small
\caption{Continued pretraining of Qwen2.5-0.5B, three seeds (window means on the common documents at each checkpoint). $\lambda$: the factor on the closed side in the annealed arm. Loss: on the common documents. $\bar D$: average damping; lobe: its lobe term; $D$: damping of the previous write (controlled); $\cos$: controlled cosine between adjacent contributions.}
\label{tab:contpre}
\fittable{\input{supplement/table_contpre_qwen05_r1}}
\end{table}

\begin{table}[htbp]
\centering
\small
\caption{Continued pretraining of Llama-3.2-1B, one seed; columns as in Table~\ref{tab:contpre}.}
\label{tab:contpre-llama}
\fittable{\input{supplement/table_contpre_llama1b_r1}}
\end{table}

\begin{figure}[htbp]
\centering
\input{figures/contpre_LlamaThreeOneB}
\caption{Continued pretraining of Llama-3.2-1B, one seed, with the closed side annealed away (solid; $\lambda$ in panel c, right axis) or left in place (dashed), as in Figure~\ref{fig:contpre}. Values in Table~\ref{tab:contpre-llama}.}
\label{fig:contpre-llama}
\end{figure}

\paragraph{What happened in Qwen2.5-0.5B.} The values are means over the three seeds, which agree to within \CpSeedRangeMax{} at the end of the run on every quantity below (Table~\ref{tab:contpre} lists each seed). In the control arm, \CpTokensM M tokens of continued training move the window cosine from $\CpStartCos$ to $\CpEndControlCos$ and $D$ from \CpStartD{} to \CpEndControlD; the loss on the common documents rises by \CpControlLossDrift{} and the validation loss by \CpControlValDrift{} as the model adapts to the shard. In the annealed arm $D$, $\bar D$ and the lobe term fall with $\lambda$: at $\lambda = \CpMidLam$ the cosine is $\CpMidCos$ against $\CpMidControlCos$ in the control, and at the first checkpoint after $\lambda$ reaches zero (step \CpZeroStep) it is $+\CpZeroCos$, where it stays ($+\CpEndAnnealCos$ at the end). With $\lambda = 0$ the lobe term is zero by construction; the weakening term of open neurons, the route of the non-gated models and of the straight-through arm of Table~\ref{tab:arms}, is \CpEndAnnealWeak{} against \CpEndControlWeak{} in the control, so the model has not rebuilt the damping through it within \CpZeroTokensM M tokens, and $\bar D$ is \CpEndAnnealDbar{} against \CpEndControlDbar. Co-writing falls from \CpStartC{} to \CpEndAnnealC{} (\CpCRatio{} of the control's \CpEndControlC) while attention's term does not move (\CpEndAnnealA{} against \CpEndControlA), so $A + C - D$ ends at $+\CpEndAnnealLaw$ against $\CpEndControlLaw$. The validation loss rises as $\lambda$ falls, to \CpLamZeroValLoss{} at step \CpAnnealEnd{} against \CpLamZeroControlValLoss{} in the control, and recovers to \CpEndAnnealValLoss{} against \CpEndControlValLoss{} by the end, still falling.

\paragraph{What happened in Llama-3.2-1B.} Llama-3.2-1B cancels more strongly than Qwen2.5-0.5B on the common documents (cosine $\CpLlamaThreeOneBStartCos$, $D = \CpLlamaThreeOneBStartD$), and the control keeps it ($\CpLlamaThreeOneBEndControlCos$ and $D = \CpLlamaThreeOneBEndControlD$ at the end). The control's loss on the common documents rises by \CpLlamaThreeOneBControlLossDrift{} (validation by \CpLlamaThreeOneBControlValDrift), more than Qwen2.5-0.5B's \CpQwenHalfBControlLossDrift: at this learning rate the model moves further from its release, and the 192-token common documents are affected more than the 1{,}024-token validation sequences. In the annealed arm the cosine is $\CpLlamaThreeOneBMidCos$ at $\lambda = \CpLlamaThreeOneBMidLam$ and $+\CpLlamaThreeOneBZeroCos$ at the first checkpoint with $\lambda = 0$, where it stays ($+\CpLlamaThreeOneBEndAnnealCos$ at the end). $D$ ends at $\CpLlamaThreeOneBEndAnnealD$ and $\bar D$ at $\CpLlamaThreeOneBEndAnnealDbar$, both slightly negative: with the closed side gone, the MLP writes a little more along a direction when it is present than when it is absent, and the weakening term of the open neurons is \CpLlamaThreeOneBEndAnnealWeak{} against \CpLlamaThreeOneBEndControlWeak{} in the control. Co-writing falls from \CpLlamaThreeOneBStartC{} to \CpLlamaThreeOneBEndAnnealC{} (\CpLlamaThreeOneBCRatio{} of the control's \CpLlamaThreeOneBEndControlC), and $A + C - D$ ends at $+\CpLlamaThreeOneBEndAnnealLaw$ against $\CpLlamaThreeOneBEndControlLaw$. The loss cost is larger than Qwen2.5-0.5B's: \CpLlamaThreeOneBEndLossGap{} nats on the common documents and \CpLlamaThreeOneBEndValGap{} in validation at the end, from \CpLlamaThreeOneBLamZeroValGap{} when $\lambda$ reached zero. Whether a longer run closes the loss gaps is untested.

%% file: supplement/table_cos_controls_r1.tex
\begin{tabular}{lrrrrrrrrr}
\toprule
 & \multicolumn{8}{c}{window} & all pairs \\
\cmidrule(lr){2-9}
model & raw & top-1 & top-2 & top-5 & top-20 & mean removed & controlled & standardised & controlled \\
\midrule
Llama-3.2-1B & $-0.181$ & $-0.183$ & $-0.205$ & $-0.210$ & $-0.220$ & $-0.219$ & $-0.222$ & $-0.216$ & $-0.117$ \\
Qwen2.5-0.5B & $-0.172$ & $-0.180$ & $-0.180$ & $-0.185$ & $-0.190$ & $-0.182$ & $-0.184$ & $-0.169$ & $-0.079$ \\
TinyLlama-1.1B & $-0.024$ & $-0.026$ & $-0.028$ & $-0.028$ & $-0.034$ & $-0.029$ & $-0.029$ & $-0.042$ & $+0.020$ \\
Llama-2-7B & $-0.031$ & $-0.036$ & $-0.037$ & $-0.037$ & $-0.041$ & $-0.039$ & $-0.041$ & $-0.046$ & $+0.025$ \\
Mistral-7B & $-0.026$ & $-0.029$ & $-0.034$ & $-0.035$ & $-0.040$ & $-0.033$ & $-0.035$ & $-0.047$ & $+0.014$ \\
Llama-3.1-8B-Instruct & $-0.118$ & $-0.125$ & $-0.128$ & $-0.128$ & $-0.130$ & $-0.127$ & $-0.127$ & $-0.125$ & $-0.043$ \\
Gemma-2B & $-0.015$ & $-0.121$ & $-0.115$ & $-0.117$ & $-0.119$ & $-0.170$ & $-0.145$ & $-0.156$ & $-0.032$ \\
Phi-3-mini & $+0.083$ & $+0.059$ & $+0.030$ & $+0.020$ & $-0.053$ & $-0.047$ & $-0.052$ & $-0.067$ & $+0.015$ \\
OLMo-1B & $-0.168$ & $-0.169$ & $-0.170$ & $-0.172$ & $-0.173$ & $-0.165$ & $-0.170$ & $-0.170$ & $-0.115$ \\
OLMoE-1B-7B & $+0.063$ & $+0.045$ & $+0.045$ & $+0.045$ & $+0.035$ & $+0.033$ & $+0.031$ & $+0.006$ & $+0.091$ \\
ReluLLaMA-7B & $+0.089$ & $+0.074$ & $+0.073$ & $+0.073$ & $+0.068$ & $+0.074$ & $+0.073$ & $+0.063$ & $+0.077$ \\
ProSparse-7B & $-0.218$ & $-0.232$ & $-0.235$ & $-0.244$ & $-0.255$ & $+0.058$ & $+0.055$ & $+0.046$ & $-0.008$ \\
Pythia-160m & $-0.105$ & $-0.125$ & $-0.136$ & $-0.136$ & $-0.141$ & $-0.148$ & $-0.155$ & $-0.154$ & $-0.159$ \\
GPT-2 & $+0.220$ & $+0.220$ & $+0.216$ & $+0.162$ & $+0.153$ & $+0.140$ & $+0.148$ & $+0.146$ & $+0.202$ \\
GPT-2 medium & $+0.188$ & $+0.173$ & $+0.162$ & $+0.124$ & $+0.079$ & $+0.094$ & $+0.091$ & $+0.076$ & $+0.154$ \\
OPT-1.3B & $+0.225$ & $+0.223$ & $+0.222$ & $+0.223$ & $+0.221$ & $+0.214$ & $+0.212$ & $+0.208$ & $+0.255$ \\
\bottomrule
\end{tabular}

%% file: supplement/table_wiki_r1.tex
\begin{tabular}{lrrrrrrrrr}
\toprule
 & \multicolumn{5}{c}{WikiText-103, 192 tokens} & \multicolumn{4}{c}{FineWeb-Edu, 192 tokens (Table~\ref{tab:switchci})} \\
\cmidrule(lr){2-6}\cmidrule(lr){7-10}
model & $A$ & $C$ & $D$ & $A + C - D$ & $\cos$ & $A$ & $C$ & $D$ & $A + C - D$ \\
\midrule
Llama-3.2-1B & $-0.025$ & $+0.276$ & $0.500$ & $-0.248$ & $-0.227$ & $-0.022$ & $+0.275$ & $0.497$ & $-0.244$ \\
Qwen2.5-0.5B & $-0.060$ & $+0.214$ & $0.349$ & $-0.195$ & $-0.190$ & $-0.060$ & $+0.208$ & $0.336$ & $-0.189$ \\
TinyLlama-1.1B & $+0.013$ & $+0.113$ & $0.154$ & $-0.027$ & $-0.038$ & $+0.022$ & $+0.114$ & $0.145$ & $-0.010$ \\
Llama-2-7B & $+0.002$ & $+0.099$ & $0.159$ & $-0.058$ & $-0.055$ & $+0.008$ & $+0.100$ & $0.151$ & $-0.043$ \\
Mistral-7B & $0.000$ & $+0.138$ & $0.190$ & $-0.053$ & $-0.050$ & $+0.006$ & $+0.135$ & $0.177$ & $-0.036$ \\
Llama-3.1-8B-Instruct & $-0.010$ & $+0.152$ & $0.280$ & $-0.138$ & $-0.130$ & $-0.007$ & $+0.148$ & $0.276$ & $-0.135$ \\
Gemma-2B & $-0.083$ & $+0.213$ & $0.279$ & $-0.150$ & $-0.148$ & $-0.087$ & $+0.215$ & $0.273$ & $-0.145$ \\
Phi-3-mini & $+0.002$ & $+0.092$ & $0.147$ & $-0.053$ & $-0.057$ & $+0.005$ & $+0.095$ & $0.148$ & $-0.048$ \\
OLMo-1B & $-0.043$ & $+0.184$ & $0.325$ & $-0.184$ & $-0.180$ & $-0.037$ & $+0.180$ & $0.318$ & $-0.175$ \\
OLMoE-1B-7B & $-0.004$ & $+0.172$ & $0.128$ & $+0.041$ & $+0.011$ & $+0.005$ & $+0.190$ & $0.118$ & $+0.076$ \\
\midrule
ReluLLaMA-7B & $+0.004$ & $+0.066$ & $0.000$ & $+0.070$ & $+0.056$ & $+0.011$ & $+0.071$ & $-0.006$ & $+0.089$ \\
ProSparse-7B & $-0.015$ & $+0.060$ & $-0.006$ & $+0.051$ & $+0.041$ & $-0.012$ & $+0.068$ & $-0.011$ & $+0.067$ \\
\midrule
Pythia-160m & $+0.028$ & $+0.160$ & $0.351$ & $-0.164$ & $-0.155$ & $+0.036$ & $+0.160$ & $0.361$ & $-0.165$ \\
GPT-2 & $+0.131$ & $+0.175$ & $0.140$ & $+0.166$ & $+0.139$ & $+0.140$ & $+0.179$ & $0.141$ & $+0.178$ \\
GPT-2 medium & $+0.084$ & $+0.132$ & $0.120$ & $+0.096$ & $+0.088$ & $+0.089$ & $+0.138$ & $0.127$ & $+0.100$ \\
OPT-1.3B & $+0.090$ & $+0.103$ & $-0.018$ & $+0.212$ & $+0.180$ & $+0.109$ & $+0.117$ & $-0.025$ & $+0.251$ \\
\bottomrule
\end{tabular}

%% file: supplement/table_switch_raw_r1.tex
\begin{tabular}{lrrrrrrrrr}
\toprule
 & \multicolumn{5}{c}{raw frame, $v$ removed from the MLP input} & \multicolumn{3}{c}{controlled, $v$ removed from the layer input} & \\
\cmidrule(lr){2-6}\cmidrule(lr){7-9}
model & $A$ & $C$ & $D$ & $A + C - D$ & $\cos$ raw & $A$ & $C$ & $D$ & linearity \\
\midrule
Llama-3.2-1B & $-0.028$ & $+0.282$ & $+0.452$ & $-0.199$ & $-0.181$ & $+0.002$ & $+0.228$ & $+0.450$ & $0.13$ \\
Qwen2.5-0.5B & $-0.065$ & $+0.212$ & $+0.324$ & $-0.176$ & $-0.172$ & $+0.009$ & $+0.183$ & $+0.311$ & $0.10$ \\
TinyLlama-1.1B & $+0.025$ & $+0.128$ & $+0.159$ & $-0.005$ & $-0.024$ & $+0.016$ & $+0.112$ & $+0.143$ & $0.08$ \\
Llama-2-7B & $+0.015$ & $+0.106$ & $+0.153$ & $-0.031$ & $-0.031$ & $+0.013$ & $+0.099$ & $+0.150$ & $0.13$ \\
Mistral-7B & $+0.003$ & $+0.149$ & $+0.178$ & $-0.026$ & $-0.026$ & $+0.017$ & $+0.132$ & $+0.174$ & $0.08$ \\
Llama-3.1-8B-Instruct & $-0.023$ & $+0.158$ & $+0.262$ & $-0.126$ & $-0.118$ & $+0.006$ & $+0.131$ & $+0.259$ & $0.14$ \\
Gemma-2B & $-0.004$ & $+0.236$ & $+0.256$ & $-0.024$ & $-0.015$ & $+0.010$ & $+0.195$ & $+0.254$ & $0.18$ \\
Phi-3-mini & $+0.109$ & $+0.132$ & $+0.130$ & $+0.111$ & $+0.083$ & $+0.009$ & $+0.097$ & $+0.150$ & $0.16$ \\
OLMo-1B & $-0.035$ & $+0.164$ & $+0.301$ & $-0.172$ & $-0.168$ & $+0.008$ & $+0.155$ & $+0.293$ & $0.13$ \\
OLMoE-1B-7B & $+0.017$ & $+0.243$ & $+0.141$ & $+0.119$ & $+0.063$ & $+0.038$ & $+0.180$ & $+0.108$ & $0.10$ \\
\midrule
ReluLLaMA-7B & $+0.086$ & $+0.029$ & $+0.007$ & $+0.108$ & $+0.089$ & $+0.019$ & $+0.084$ & $+0.007$ & $0.18$ \\
ProSparse-7B & $-0.252$ & $+0.016$ & $+0.023$ & $-0.260$ & $-0.218$ & $+0.009$ & $+0.075$ & $-0.004$ & $0.22$ \\
\midrule
Pythia-160m & $+0.059$ & $+0.209$ & $+0.381$ & $-0.114$ & $-0.105$ & $+0.053$ & $+0.160$ & $+0.361$ & $0.17$ \\
GPT-2 & $+0.182$ & $+0.216$ & $+0.135$ & $+0.263$ & $+0.220$ & $+0.131$ & $+0.185$ & $+0.148$ & $0.06$ \\
GPT-2 medium & $+0.153$ & $+0.171$ & $+0.116$ & $+0.208$ & $+0.188$ & $+0.094$ & $+0.136$ & $+0.125$ & $0.08$ \\
OPT-1.3B & $+0.117$ & $+0.134$ & $-0.007$ & $+0.259$ & $+0.225$ & $+0.098$ & $+0.121$ & $-0.022$ & --- \\
\bottomrule
\end{tabular}

%% file: figures/protocol.tex
\begin{tikzpicture}
\begin{groupplot}[group style={group size=2 by 1, horizontal sep=0.5cm, yticklabels at=edge left}, paper, zero x, scale only axis,
  width=\dimexpr0.5\linewidth-2.2cm\relax, height=5.2cm, xmin=-0.35, xmax=0.3, ymin=-0.6, ymax=\FigYMax, modelrows={figures/data/figproto.csv}{\FigYTicks}, enlarge y limits=false,
  extra y ticks={\FigSepCG, \FigSepNC}, extra y tick labels={}, extra y tick style={grid=major, grid style={draw=axisgray, line width=0.4pt}, major tick length=0pt}, xlabel={window mean of $\cos(c_{j-1}, c_j)$}]
\nextgroupplot[title={(a) raw}, legend to name=protoleg, legend columns=3, legend style={/tikz/every even column/.append style={column sep=0.5cm}}]
\addplot[dot=series1] table[x=wraw, y=y] {figures/data/figproto.csv}; \addlegendentry{WikiText-103, 2{,}048 tokens}
\addplot[diamond=series2] table[x=fraw, y=y] {figures/data/figproto.csv}; \addlegendentry{FineWeb-Edu, 192 tokens}
\addplot[tri=series3] table[x=wmh, y=y] {figures/data/figproto.csv}; \addlegendentry{WikiText-103, middle half of the layers instead of the window}
\nextgroupplot[title={(b) controlled}]
\addplot[dot=series1] table[x=wctl, y=y] {figures/data/figproto.csv}; \addplot[diamond=series2] table[x=fctl, y=y] {figures/data/figproto.csv};
\end{groupplot}
\end{tikzpicture}

\smallskip
\pgfplotslegendfromname{protoleg}

%% file: table_switch_r1.tex
\begin{tabular}{lrrrrr}
\toprule
model & $A$ & $C$ & $D$ & $A + C - D$ & $\cos$ ctrl. \\
\midrule
Llama-3.2-1B & $-0.022$ {\scriptsize$[-0.024, -0.021]$} & $+0.275$ {\scriptsize$[+0.272, +0.279]$} & $+0.497$ {\scriptsize$[+0.493, +0.501]$} & $-0.244$ {\scriptsize$[-0.247, -0.242]$} & $-0.222$ \\
Qwen2.5-0.5B & $-0.060$ {\scriptsize$[-0.061, -0.058]$} & $+0.208$ {\scriptsize$[+0.205, +0.210]$} & $+0.336$ {\scriptsize$[+0.333, +0.340]$} & $-0.189$ {\scriptsize$[-0.191, -0.186]$} & $-0.184$ \\
TinyLlama-1.1B & $+0.022$ {\scriptsize$[+0.020, +0.023]$} & $+0.114$ {\scriptsize$[+0.112, +0.116]$} & $+0.145$ {\scriptsize$[+0.142, +0.149]$} & $-0.010$ {\scriptsize$[-0.014, -0.006]$} & $-0.029$ \\
Llama-2-7B & $+0.008$ {\scriptsize$[+0.006, +0.010]$} & $+0.100$ {\scriptsize$[+0.098, +0.101]$} & $+0.151$ {\scriptsize$[+0.147, +0.154]$} & $-0.043$ {\scriptsize$[-0.047, -0.038]$} & $-0.041$ \\
Mistral-7B & $+0.006$ {\scriptsize$[+0.004, +0.007]$} & $+0.135$ {\scriptsize$[+0.133, +0.137]$} & $+0.177$ {\scriptsize$[+0.171, +0.183]$} & $-0.036$ {\scriptsize$[-0.042, -0.030]$} & $-0.035$ \\
Llama-3.1-8B-Instruct & $-0.007$ {\scriptsize$[-0.008, -0.006]$} & $+0.148$ {\scriptsize$[+0.146, +0.150]$} & $+0.276$ {\scriptsize$[+0.272, +0.279]$} & $-0.135$ {\scriptsize$[-0.138, -0.132]$} & $-0.127$ \\
Gemma-2B & $-0.087$ {\scriptsize$[-0.089, -0.085]$} & $+0.215$ {\scriptsize$[+0.211, +0.218]$} & $+0.273$ {\scriptsize$[+0.270, +0.277]$} & $-0.145$ {\scriptsize$[-0.148, -0.143]$} & $-0.145$ \\
Phi-3-mini & $+0.005$ {\scriptsize$[+0.003, +0.007]$} & $+0.095$ {\scriptsize$[+0.093, +0.097]$} & $+0.148$ {\scriptsize$[+0.145, +0.151]$} & $-0.048$ {\scriptsize$[-0.052, -0.044]$} & $-0.052$ \\
OLMo-1B & $-0.037$ {\scriptsize$[-0.039, -0.035]$} & $+0.180$ {\scriptsize$[+0.178, +0.183]$} & $+0.318$ {\scriptsize$[+0.313, +0.323]$} & $-0.175$ {\scriptsize$[-0.178, -0.171]$} & $-0.170$ \\
OLMoE-1B-7B & $+0.005$ {\scriptsize$[+0.002, +0.008]$} & $+0.190$ {\scriptsize$[+0.184, +0.195]$} & $+0.118$ {\scriptsize$[+0.113, +0.123]$} & $+0.076$ {\scriptsize$[+0.067, +0.086]$} & $+0.031$ \\
\midrule
ReluLLaMA-7B & $+0.011$ {\scriptsize$[+0.009, +0.013]$} & $+0.071$ {\scriptsize$[+0.069, +0.073]$} & $-0.006$ {\scriptsize$[-0.008, -0.004]$} & $+0.089$ {\scriptsize$[+0.084, +0.093]$} & $+0.073$ \\
ProSparse-7B & $-0.012$ {\scriptsize$[-0.014, -0.009]$} & $+0.068$ {\scriptsize$[+0.066, +0.069]$} & $-0.011$ {\scriptsize$[-0.012, -0.010]$} & $+0.067$ {\scriptsize$[+0.063, +0.071]$} & $+0.055$ \\
\midrule
Pythia-160m & $+0.036$ {\scriptsize$[+0.034, +0.038]$} & $+0.160$ {\scriptsize$[+0.158, +0.163]$} & $+0.361$ {\scriptsize$[+0.357, +0.366]$} & $-0.165$ {\scriptsize$[-0.171, -0.160]$} & $-0.155$ \\
GPT-2 & $+0.140$ {\scriptsize$[+0.137, +0.144]$} & $+0.179$ {\scriptsize$[+0.176, +0.181]$} & $+0.141$ {\scriptsize$[+0.139, +0.144]$} & $+0.178$ {\scriptsize$[+0.173, +0.182]$} & $+0.148$ \\
GPT-2 medium & $+0.089$ {\scriptsize$[+0.087, +0.091]$} & $+0.138$ {\scriptsize$[+0.136, +0.140]$} & $+0.127$ {\scriptsize$[+0.124, +0.130]$} & $+0.100$ {\scriptsize$[+0.096, +0.103]$} & $+0.091$ \\
OPT-1.3B & $+0.109$ {\scriptsize$[+0.106, +0.112]$} & $+0.117$ {\scriptsize$[+0.115, +0.119]$} & $-0.025$ {\scriptsize$[-0.026, -0.025]$} & $+0.251$ {\scriptsize$[+0.247, +0.256]$} & $+0.212$ \\
\bottomrule
\end{tabular}

%% file: figures/switch.tex
\begin{tikzpicture}
\begin{groupplot}[group style={group size=4 by 1, horizontal sep=0.95cm}, width=0.255\linewidth, height=4.4cm, paper, zero y, ymin=-0.25, ymax=0.45]
\nextgroupplot[xmode=log, xlabel={training tokens (billions)}, ylabel={window mean}, title={(a) OLMo-1B}, xmin=1.5, xmax=4000]
\addplot[series1, mark=*, mark size=1.3pt] table[x=x, y=Dw] {figures/data/tabolmotrain.csv};
\addplot[series2, mark=square*, mark size=1.3pt] table[x=x, y=C] {figures/data/tabolmotrain.csv};
\addplot[series3, mark=triangle*, mark size=1.5pt] table[x=x, y=A] {figures/data/tabolmotrain.csv};
\addplot[black!70, dashed, mark=o, mark size=1.3pt] table[x=x, y=cos] {figures/data/tabolmotrain.csv};
\nextgroupplot[xmode=log, xlabel={training step}, title={(b) Pythia-160m},
  legend to name=switchleg, legend columns=4, legend style={/tikz/every even column/.append style={column sep=0.3cm}}]
\addplot[series1, mark=*, mark size=1.3pt] table[x=x, y=Dw] {figures/data/tabpythiatrain.csv}; \addlegendentry{damping $D$}
\addplot[series2, mark=square*, mark size=1.3pt] table[x=x, y=C] {figures/data/tabpythiatrain.csv}; \addlegendentry{co-writing $C$}
\addplot[series3, mark=triangle*, mark size=1.5pt] table[x=x, y=A] {figures/data/tabpythiatrain.csv}; \addlegendentry{attention $A$}
\addplot[black!70, dashed, mark=o, mark size=1.3pt] table[x=x, y=cos] {figures/data/tabpythiatrain.csv}; \addlegendentry{$\cos(c_{j-1}, c_j)$}
\nextgroupplot[xlabel={training tokens (billions)}, title={(c) TinyLlama-1.1B}, xmin=0, xmax=3100]
\addplot[series1, mark=*, mark size=1.3pt] table[x=x, y=Dw] {figures/data/tabtinyllamatrain.csv};
\addplot[series2, mark=square*, mark size=1.3pt] table[x=x, y=C] {figures/data/tabtinyllamatrain.csv};
\addplot[series3, mark=triangle*, mark size=1.5pt] table[x=x, y=A] {figures/data/tabtinyllamatrain.csv};
\addplot[black!70, dashed, mark=o, mark size=1.3pt] table[x=x, y=cos] {figures/data/tabtinyllamatrain.csv};
\nextgroupplot[xmode=log, xlabel={training tokens (billions)}, title={(d) OLMoE-1B-7B}, xmin=30, xmax=7000]
\addplot[series1, mark=*, mark size=1.3pt] table[x=x, y=Dw] {figures/data/tabolmoetrain.csv};
\addplot[series2, mark=square*, mark size=1.3pt] table[x=x, y=C] {figures/data/tabolmoetrain.csv};
\addplot[series3, mark=triangle*, mark size=1.5pt] table[x=x, y=A] {figures/data/tabolmoetrain.csv};
\addplot[black!70, dashed, mark=o, mark size=1.3pt] table[x=x, y=cos] {figures/data/tabolmoetrain.csv};
\end{groupplot}
\end{tikzpicture}

\smallskip
\pgfplotslegendfromname{switchleg}

%% file: figures/lags.tex
\begin{tikzpicture}
\begin{groupplot}[group style={group size=4 by 1, horizontal sep=1.1cm}, paper, scale only axis, width=\dimexpr0.25\linewidth-1.15cm\relax, height=3.4cm,
  xmin=0.8, xmax=3.2, xtick={1,2,3}, xlabel={lag}, unbounded coords=jump, every axis plot/.append style={line width=0.8pt, mark size=1.4pt}]
\nextgroupplot[title={(a) controlled cosine}, zero y, legend to name=lagleg, legend columns=3, legend style={/tikz/every even column/.append style={column sep=0.5cm}}]
\addplot[series1, mark=*] table[x=lag, y=cos] {figures/data/figlagsg.csv}; \addlegendentry{gated}
\addplot[series3, mark=triangle*] table[x=lag, y=cos] {figures/data/figlagsc.csv}; \addlegendentry{ReLU conversions}
\addplot[series2, mark=diamond*] table[x=lag, y=cos] {figures/data/figlagsn.csv}; \addlegendentry{no gate}
\nextgroupplot[title={(b) damping $D$}, zero y]
\addplot[series1, mark=*] table[x=lag, y=D] {figures/data/figlagsg.csv}; \addplot[series3, mark=triangle*] table[x=lag, y=D] {figures/data/figlagsc.csv}; \addplot[series2, mark=diamond*] table[x=lag, y=D] {figures/data/figlagsn.csv};
\nextgroupplot[title={(c) attention $A$}, zero y]
\addplot[series1, mark=*] table[x=lag, y=A] {figures/data/figlagsg.csv}; \addplot[series3, mark=triangle*] table[x=lag, y=A] {figures/data/figlagsc.csv}; \addplot[series2, mark=diamond*] table[x=lag, y=A] {figures/data/figlagsn.csv};
\nextgroupplot[title={(d) own co-writing}, zero y]
\addplot[series1, mark=*] table[x=lag, y=Cown] {figures/data/figlagsg.csv}; \addplot[series3, mark=triangle*] table[x=lag, y=Cown] {figures/data/figlagsc.csv}; \addplot[series2, mark=diamond*] table[x=lag, y=Cown] {figures/data/figlagsn.csv};
\end{groupplot}
\end{tikzpicture}

\smallskip
\pgfplotslegendfromname{lagleg}

%% file: table_family_r1.tex
\begin{tabular}{llrrrrrrrrr}
\toprule
model & MLP & $\bar D$ & lobe & no closed side & weakening & $D_{\mathrm{prev}}:\bar D$ & read--write & closed & $\cos$ ctrl. & $\cos$ raw \\
\midrule
Llama-3.2-1B & SwiGLU & $0.455$ & $0.470$ & $-0.017$ & $0.011$ & $1.08$ & $+0.370$ & $0.82$ & $-0.222$ & $-0.181$ \\
Qwen2.5-0.5B & SwiGLU & $0.304$ & $0.307$ & $+0.001$ & $0.031$ & $1.10$ & $+0.309$ & $0.80$ & $-0.184$ & $-0.172$ \\
TinyLlama-1.1B & SwiGLU & $0.160$ & $0.171$ & $-0.023$ & $0.007$ & $1.20$ & $+0.362$ & $0.72$ & $-0.029$ & $-0.024$ \\
Llama-2-7B & SwiGLU & $0.146$ & $0.130$ & $+0.002$ & $0.011$ & $1.12$ & $+0.321$ & $0.73$ & $-0.041$ & $-0.031$ \\
Mistral-7B & SwiGLU & $0.184$ & $0.193$ & $-0.015$ & $0.009$ & $1.12$ & $+0.403$ & $0.78$ & $-0.035$ & $-0.026$ \\
Llama-3.1-8B-Instruct & SwiGLU & $0.255$ & $0.251$ & $-0.006$ & $0.010$ & $1.10$ & $+0.365$ & $0.79$ & $-0.127$ & $-0.118$ \\
Gemma-2B & GeGLU & $0.309$ & $0.274$ & $+0.032$ & $0.060$ & $0.92$ & $+0.314$ & $0.83$ & $-0.145$ & $-0.015$ \\
Phi-3-mini & SwiGLU & $0.148$ & $0.099$ & $+0.034$ & $0.032$ & $1.13$ & $+0.163$ & $0.62$ & $-0.052$ & $+0.083$ \\
OLMo-1B & SwiGLU & $0.294$ & $0.172$ & $+0.100$ & $0.076$ & $1.10$ & $+0.038$ & $0.62$ & $-0.170$ & $-0.168$ \\
OLMoE-1B-7B & SwiGLU, MoE & $0.152$ & $0.125$ & $+0.022$ & $0.033$ & $1.13$ & $+0.199$ & $0.68$ & $+0.031$ & $+0.063$ \\
\midrule
ReluLLaMA-7B & ReGLU & $0.017$ & $0.000$ & $+0.017$ & $0.018$ & --- & $+0.299$ & $0.68$ & $+0.073$ & $+0.089$ \\
ProSparse-7B & ReGLU$^\ast$ & $0.016$ & $0.000$ & $+0.015$ & $0.013$ & --- & $+0.269$ & $0.80$ & $+0.055$ & $-0.218$ \\
\midrule
Pythia-160m & GELU & $0.279$ & $0.011$ & --- & $0.308$ & $1.31$ & $-0.362$ & $0.79$ & $-0.155$ & $-0.105$ \\
GPT-2 & GELU & $0.227$ & $0.036$ & --- & $0.236$ & $0.93$ & $-0.096$ & $0.85$ & $+0.148$ & $+0.220$ \\
GPT-2 medium & GELU & $0.140$ & $0.026$ & --- & $0.150$ & $1.11$ & $-0.102$ & $0.88$ & $+0.091$ & $+0.188$ \\
OPT-1.3B & ReLU & $0.026$ & $0.000$ & --- & $0.030$ & --- & $+0.026$ & $0.97$ & $+0.212$ & $+0.225$ \\
\bottomrule
\end{tabular}

%% file: table_group_r1.tex
\begin{tabular}{lrrrrrrrrr}
\toprule
 & \multicolumn{5}{c}{damping $D$ of the previous write} & \multicolumn{4}{c}{co-writing $C$} \\
\cmidrule(lr){2-6}\cmidrule(lr){7-10}
model & closed str. & closed weak. & open str. & open weak. & total & closed str. & closed weak. & open str. & open weak. \\
\midrule
Llama-3.2-1B & $+0.531$ & $-0.004$ & $-0.045$ & $+0.015$ & $+0.497$ & $+0.242$ & $+0.006$ & $+0.019$ & $+0.008$ \\
Qwen2.5-0.5B & $+0.355$ & $-0.013$ & $-0.049$ & $+0.044$ & $+0.336$ & $+0.174$ & $+0.005$ & $+0.002$ & $+0.025$ \\
TinyLlama-1.1B & $+0.203$ & $-0.003$ & $-0.069$ & $+0.015$ & $+0.145$ & $+0.044$ & $+0.011$ & $+0.053$ & $+0.005$ \\
Llama-2-7B & $+0.161$ & $-0.002$ & $-0.034$ & $+0.027$ & $+0.151$ & $+0.037$ & $+0.012$ & $+0.038$ & $+0.013$ \\
Mistral-7B & $+0.217$ & $-0.005$ & $-0.057$ & $+0.022$ & $+0.177$ & $+0.061$ & $+0.016$ & $+0.045$ & $+0.013$ \\
Llama-3.1-8B-Instruct & $+0.290$ & $-0.003$ & $-0.033$ & $+0.021$ & $+0.276$ & $+0.108$ & $+0.009$ & $+0.021$ & $+0.010$ \\
Gemma-2B & $+0.253$ & $-0.011$ & $-0.053$ & $+0.085$ & $+0.273$ & $+0.156$ & $+0.007$ & $-0.007$ & $+0.058$ \\
Phi-3-mini & $+0.121$ & $-0.003$ & $-0.029$ & $+0.059$ & $+0.148$ & $+0.030$ & $+0.012$ & $+0.028$ & $+0.025$ \\
OLMo-1B & $+0.227$ & $-0.012$ & $-0.012$ & $+0.115$ & $+0.318$ & $+0.097$ & $+0.015$ & $+0.018$ & $+0.049$ \\
OLMoE-1B-7B & $+0.142$ & $-0.013$ & $-0.058$ & $+0.047$ & $+0.118$ & $+0.041$ & $+0.025$ & $+0.064$ & $+0.060$ \\
\bottomrule
\end{tabular}

%% file: supplement/table_decomp_r1.tex
\begin{tabular}{lrrrrrrrr}
\toprule
 & \multicolumn{2}{c}{through the gate} & \multicolumn{4}{c}{through the linear input} & & \\
\cmidrule(lr){2-3}\cmidrule(lr){4-7}
model & closed & open & closed, str. & closed, weak. & open, str. & open, weak. & radial & $D$ \\
\midrule
Llama-3.2-1B & $+0.007$ & $+0.006$ & $+0.470$ & $-0.003$ & $-0.034$ & $+0.011$ & $0.000$ & $0.455$ \\
Qwen2.5-0.5B & $+0.006$ & $+0.008$ & $+0.307$ & $-0.009$ & $-0.038$ & $+0.031$ & $0.000$ & $0.304$ \\
TinyLlama-1.1B & $+0.014$ & $+0.011$ & $+0.171$ & $-0.002$ & $-0.040$ & $+0.007$ & $0.000$ & $0.160$ \\
Llama-2-7B & $+0.017$ & $+0.018$ & $+0.130$ & $-0.003$ & $-0.026$ & $+0.011$ & $0.000$ & $0.146$ \\
Mistral-7B & $+0.010$ & $+0.012$ & $+0.193$ & $-0.003$ & $-0.036$ & $+0.009$ & $0.000$ & $0.184$ \\
Llama-3.1-8B-Instruct & $+0.012$ & $+0.013$ & $+0.251$ & $-0.003$ & $-0.029$ & $+0.010$ & $0.000$ & $0.255$ \\
Gemma-2B & $+0.015$ & $+0.008$ & $+0.274$ & $-0.011$ & $-0.037$ & $+0.060$ & $0.000$ & $0.309$ \\
Phi-3-mini & $+0.021$ & $+0.027$ & $+0.099$ & $-0.005$ & $-0.025$ & $+0.032$ & $0.000$ & $0.148$ \\
OLMo-1B & $+0.044$ & $+0.045$ & $+0.172$ & $-0.021$ & $-0.021$ & $+0.076$ & $0.000$ & $0.294$ \\
OLMoE-1B-7B & $+0.011$ & $+0.014$ & $+0.125$ & $-0.006$ & $-0.025$ & $+0.033$ & $0.000$ & $0.152$ \\
ReluLLaMA-7B & $0.000$ & $+0.035$ & $0.000$ & $0.000$ & $-0.036$ & $+0.018$ & $0.000$ & $0.017$ \\
ProSparse-7B & $0.000$ & $+0.022$ & $0.000$ & $0.000$ & $-0.020$ & $+0.013$ & $0.000$ & $0.016$ \\
\midrule
 & \multicolumn{2}{c}{slope $<0$} & \multicolumn{2}{c}{slope $>0$} & & & & \\
\cmidrule(lr){2-3}\cmidrule(lr){4-5}
model & aligned & anti-aligned & aligned & anti-aligned & & & & $D$ \\
\midrule
Pythia-160m & $+0.011$ & $-0.028$ & $-0.012$ & $+0.308$ & & & & $0.279$ \\
GPT-2 & $+0.036$ & $-0.022$ & $-0.023$ & $+0.236$ & & & & $0.227$ \\
GPT-2 medium & $+0.026$ & $-0.020$ & $-0.016$ & $+0.150$ & & & & $0.140$ \\
OPT-1.3B & $0.000$ & $0.000$ & $-0.004$ & $+0.030$ & & & & $0.026$ \\
\bottomrule
\end{tabular}

%% file: supplement/table_group_local_r1.tex
\begin{tabular}{lrrrrrrrrrr}
\toprule
 & \multicolumn{5}{c}{along the previous write} & \multicolumn{5}{c}{along a random direction} \\
\cmidrule(lr){2-6}\cmidrule(lr){7-11}
model & closed str. & closed weak. & open str. & open weak. & total & closed str. & closed weak. & open str. & open weak. & total \\
\midrule
Llama-3.2-1B & $+0.499$ & $-0.003$ & $-0.036$ & $+0.014$ & $+0.474$ & $+0.475$ & $-0.003$ & $-0.029$ & $+0.011$ & $+0.456$ \\
Qwen2.5-0.5B & $+0.347$ & $-0.012$ & $-0.054$ & $+0.048$ & $+0.329$ & $+0.311$ & $-0.008$ & $-0.033$ & $+0.034$ & $+0.304$ \\
TinyLlama-1.1B & $+0.201$ & $-0.001$ & $-0.049$ & $+0.018$ & $+0.168$ & $+0.182$ & $0.000$ & $-0.033$ & $+0.010$ & $+0.159$ \\
Llama-2-7B & $+0.155$ & $-0.001$ & $-0.028$ & $+0.030$ & $+0.156$ & $+0.143$ & $+0.001$ & $-0.016$ & $+0.018$ & $+0.146$ \\
Mistral-7B & $+0.209$ & $-0.003$ & $-0.048$ & $+0.023$ & $+0.181$ & $+0.199$ & $0.000$ & $-0.029$ & $+0.014$ & $+0.184$ \\
Llama-3.1-8B-Instruct & $+0.273$ & $-0.001$ & $-0.030$ & $+0.021$ & $+0.263$ & $+0.260$ & $+0.001$ & $-0.021$ & $+0.016$ & $+0.255$ \\
Gemma-2B & $+0.261$ & $-0.005$ & $-0.065$ & $+0.106$ & $+0.297$ & $+0.280$ & $-0.003$ & $-0.032$ & $+0.063$ & $+0.309$ \\
Phi-3-mini & $+0.122$ & $+0.001$ & $-0.019$ & $+0.074$ & $+0.178$ & $+0.114$ & $+0.001$ & $-0.011$ & $+0.044$ & $+0.148$ \\
OLMo-1B & $+0.218$ & $-0.011$ & $-0.012$ & $+0.116$ & $+0.311$ & $+0.199$ & $-0.005$ & $0.000$ & $+0.100$ & $+0.294$ \\
OLMoE-1B-7B & $+0.142$ & $-0.008$ & $-0.030$ & $+0.062$ & $+0.166$ & $+0.133$ & $-0.003$ & $-0.018$ & $+0.040$ & $+0.152$ \\
\bottomrule
\end{tabular}

%% file: figures/spectrum.tex
\begin{tikzpicture}
\begin{groupplot}[group style={group size=2 by 1, horizontal sep=0.5cm, yticklabels at=edge left}, paper, scale only axis,
  width=\dimexpr0.5\linewidth-2.2cm\relax, height=4.2cm, ymin=-0.6, ymax=8.6, modelrows={figures/data/figspec.csv}{\FigGYTicks}, enlarge y limits=false]
\nextgroupplot[title={(a) damping across directions}, zero x, xmin=-0.6, xmax=1.4, xtick={-0.5,0,0.5,1}, xlabel={damping of a direction},
  legend to name=specleg, legend columns=4, legend style={/tikz/every even column/.append style={column sep=0.4cm}}]
\addlegendimage{area legend, fill=black!15, draw=black!15}\addlegendentry{5th--95th percentile}
\addplot[black!15, line width=5pt, hbar=qlen, forget plot] table[x=q05, y=y] {figures/data/figspec.csv};
\addplot[tick mark=black!85] table[x=mean, y=y] {figures/data/figspec.csv}; \addlegendentry{mean $\bar D$}
\addplot[dot=series1] table[x=lobe, y=y] {figures/data/figspec.csv}; \addlegendentry{lobe part}
\addplot[diamond=inkmuted] table[x=rest, y=y] {figures/data/figspec.csv}; \addlegendentry{rest}
\nextgroupplot[title={(b) share of directions}, xmin=0, xmax=1, xtick={0,0.25,0.5,0.75,1}, xlabel={fraction of directions},
  legend to name=specleg2, legend columns=3, legend style={/tikz/every even column/.append style={column sep=0.4cm}}]
\addlegendimage{area legend, fill=series1dark, draw=series1dark}\addlegendentry{damped by more than $0.8$}
\addlegendimage{area legend, fill=series1light, draw=series1light}\addlegendentry{damped by less}
\addlegendimage{area legend, fill=series2, draw=series2}\addlegendentry{amplified}
\addplot[series1dark, line width=8pt, hbar=full, forget plot] table[x expr=0, y=y] {figures/data/figspec.csv};
\addplot[series1light, line width=8pt, hbar=part, forget plot] table[x=full, y=y] {figures/data/figspec.csv};
\addplot[series2, line width=8pt, hbar=expand, forget plot] table[x=contract, y=y] {figures/data/figspec.csv};
\end{groupplot}
\end{tikzpicture}

\smallskip
\pgfplotslegendfromname{specleg}\hspace{1cm}\pgfplotslegendfromname{specleg2}

%% file: supplement/table_selectivity_r1.tex
\begin{tabular}{lrrrrrrrrr}
\toprule
 & \multicolumn{6}{c}{local damping along a unit direction orthogonal to $x'$} & \multicolumn{2}{c}{ratio} & gate flips \\
\cmidrule(lr){2-7}\cmidrule(lr){8-9}
model & previous write & another token's & another document's & attention write & shuffled write & random & write : other & write : random & \% \\
\midrule
Llama-3.2-1B & $0.474$ & $0.506$ & $0.513$ & $0.447$ & $0.455$ & $0.456$ & $0.94$ {\scriptsize(0/16)} & $1.04$ & 10 (4.3) \\
Qwen2.5-0.5B & $0.329$ & $0.352$ & $0.357$ & $0.350$ & $0.304$ & $0.304$ & $0.94$ {\scriptsize(0/16)} & $1.08$ & 10 (4.8) \\
TinyLlama-1.1B & $0.168$ & $0.190$ & $0.197$ & $0.191$ & $0.160$ & $0.159$ & $0.88$ {\scriptsize(0/16)} & $1.06$ & 12 (4.8) \\
Llama-2-7B & $0.156$ & $0.176$ & $0.184$ & $0.159$ & $0.146$ & $0.146$ & $0.89$ {\scriptsize(0/16)} & $1.07$ & 10 (3.4) \\
Mistral-7B & $0.181$ & $0.205$ & $0.218$ & $0.187$ & $0.184$ & $0.184$ & $0.88$ {\scriptsize(0/16)} & $0.98$ & 10 (4.0) \\
Llama-3.1-8B-Instruct & $0.263$ & $0.289$ & $0.296$ & $0.267$ & $0.255$ & $0.255$ & $0.91$ {\scriptsize(0/16)} & $1.03$ & 9 (3.6) \\
Gemma-2B & $0.297$ & $0.319$ & $0.324$ & $0.332$ & $0.309$ & $0.308$ & $0.93$ {\scriptsize(0/16)} & $0.96$ & 9 (6.0) \\
Phi-3-mini & $0.178$ & $0.180$ & $0.185$ & $0.212$ & $0.148$ & $0.148$ & $0.99$ {\scriptsize(1/16)} & $1.20$ & 11 (3.9) \\
OLMo-1B & $0.311$ & $0.340$ & $0.346$ & $0.294$ & $0.294$ & $0.294$ & $0.92$ {\scriptsize(0/16)} & $1.06$ & 13 (5.3) \\
OLMoE-1B-7B & $0.165$ & $0.169$ & $0.173$ & $0.183$ & $0.151$ & $0.152$ & $0.98$ {\scriptsize(3/16)} & $1.09$ & 12 (4.8) \\
\bottomrule
\end{tabular}

%% file: supplement/table_layerjac_r1.tex
\begin{tabular}{llrrrrrrrrrr}
\toprule
 & & & \multicolumn{3}{c}{top-50 eigenvalues below $-0.75$ (\%)} & \multicolumn{3}{c}{top-50 eigenvalues above $+0.25$ (\%)} & \multicolumn{3}{c}{$c^{\top}\bar J c$} \\
\cmidrule(lr){4-6}\cmidrule(lr){7-9}\cmidrule(lr){10-12}
model, text & pairs & $n$ & write & other token & random & write & other token & random & write & other token & random \\
\midrule
Llama-3.2-1B & cos in $[-0.3, -0.1]$ & 50 & 42 & 49 & 31 & 24 & 13 & 8 & $-0.470$ & $-0.507$ & $-0.446$ \\
Llama-3.2-1B & $|$cos$| \le 0.015$ & 15 & 38 & 45 & 25 & 25 & 16 & 10 & $-0.392$ & $-0.420$ & $-0.396$ \\
Llama-3.2-1B & any & 50 & 39 & 45 & 30 & 23 & 13 & 9 & $-0.439$ & $-0.467$ & $-0.435$ \\
Llama-3.2-1B, common documents & cos in $[-0.3, -0.1]$ & 50 & 39 & 50 & 33 & 26 & 15 & 8 & $-0.466$ & $-0.500$ & $-0.467$ \\
Llama-3.2-1B, common documents & $|$cos$| \le 0.015$ & 50 & 34 & 41 & 29 & 30 & 19 & 9 & $-0.370$ & $-0.414$ & $-0.422$ \\
Llama-3.2-1B, common documents & any & 50 & 40 & 47 & 32 & 25 & 16 & 10 & $-0.469$ & $-0.497$ & $-0.461$ \\
Qwen2.5-0.5B & cos in $[-0.3, -0.1]$ & 50 & 37 & 39 & 24 & 15 & 13 & 8 & $-0.398$ & $-0.429$ & $-0.351$ \\
Qwen2.5-0.5B & $|$cos$| \le 0.015$ & 19 & 28 & 31 & 22 & 26 & 18 & 12 & $-0.253$ & $-0.325$ & $-0.312$ \\
Qwen2.5-0.5B & any & 50 & 36 & 36 & 22 & 15 & 10 & 8 & $-0.370$ & $-0.411$ & $-0.342$ \\
Qwen2.5-0.5B, common documents & cos in $[-0.3, -0.1]$ & 50 & 38 & 39 & 21 & 15 & 12 & 7 & $-0.387$ & $-0.412$ & $-0.328$ \\
Qwen2.5-0.5B, common documents & $|$cos$| \le 0.015$ & 50 & 25 & 24 & 21 & 24 & 19 & 7 & $-0.224$ & $-0.299$ & $-0.314$ \\
Qwen2.5-0.5B, common documents & any & 50 & 32 & 31 & 18 & 18 & 12 & 8 & $-0.342$ & $-0.363$ & $-0.316$ \\
TinyLlama-1.1B & cos in $[-0.3, -0.1]$ & 50 & 35 & 33 & 11 & 23 & 18 & 13 & $-0.273$ & $-0.271$ & $-0.190$ \\
TinyLlama-1.1B & $|$cos$| \le 0.015$ & 50 & 28 & 25 & 10 & 25 & 21 & 16 & $-0.205$ & $-0.203$ & $-0.150$ \\
TinyLlama-1.1B & any & 50 & 26 & 26 & 9 & 25 & 19 & 16 & $-0.208$ & $-0.226$ & $-0.150$ \\
TinyLlama-1.1B, common documents & cos in $[-0.3, -0.1]$ & 50 & 34 & 33 & 10 & 23 & 17 & 13 & $-0.261$ & $-0.264$ & $-0.184$ \\
TinyLlama-1.1B, common documents & $|$cos$| \le 0.015$ & 50 & 27 & 27 & 10 & 25 & 19 & 15 & $-0.206$ & $-0.219$ & $-0.158$ \\
TinyLlama-1.1B, common documents & any & 50 & 28 & 27 & 11 & 24 & 19 & 12 & $-0.215$ & $-0.220$ & $-0.168$ \\
Llama-2-7B & cos in $[-0.3, -0.1]$ & 30 & 33 & 31 & 8 & 30 & 15 & 9 & $-0.260$ & $-0.245$ & $-0.196$ \\
Llama-2-7B & $|$cos$| \le 0.015$ & 30 & 27 & 21 & 4 & 25 & 19 & 10 & $-0.160$ & $-0.157$ & $-0.129$ \\
Llama-2-7B & any & 30 & 29 & 27 & 7 & 25 & 18 & 10 & $-0.183$ & $-0.185$ & $-0.149$ \\
Mistral-7B & cos in $[-0.3, -0.1]$ & 30 & 43 & 37 & 11 & 30 & 21 & 8 & $-0.296$ & $-0.280$ & $-0.227$ \\
Mistral-7B & $|$cos$| \le 0.015$ & 30 & 31 & 28 & 6 & 30 & 21 & 9 & $-0.200$ & $-0.209$ & $-0.182$ \\
Mistral-7B & any & 30 & 33 & 30 & 9 & 33 & 23 & 9 & $-0.217$ & $-0.222$ & $-0.187$ \\
Llama-3.1-8B-Instruct & cos in $[-0.3, -0.1]$ & 50 & 37 & 41 & 18 & 29 & 16 & 8 & $-0.339$ & $-0.350$ & $-0.306$ \\
Llama-3.1-8B-Instruct & $|$cos$| \le 0.015$ & 50 & 18 & 12 & 3 & 28 & 22 & 11 & $-0.125$ & $-0.131$ & $-0.131$ \\
Llama-3.1-8B-Instruct & any & 50 & 32 & 31 & 13 & 30 & 18 & 9 & $-0.275$ & $-0.287$ & $-0.245$ \\
Llama-3.1-8B-Instruct, common documents & cos in $[-0.3, -0.1]$ & 30 & 39 & 41 & 15 & 29 & 15 & 9 & $-0.348$ & $-0.349$ & $-0.285$ \\
Llama-3.1-8B-Instruct, common documents & $|$cos$| \le 0.015$ & 30 & 18 & 16 & 4 & 31 & 22 & 9 & $-0.143$ & $-0.140$ & $-0.148$ \\
Llama-3.1-8B-Instruct, common documents & any & 30 & 31 & 29 & 13 & 27 & 14 & 7 & $-0.273$ & $-0.299$ & $-0.266$ \\
Gemma-2B & cos in $[-0.3, -0.1]$ & 30 & 39 & 38 & 22 & 22 & 18 & 8 & $-0.358$ & $-0.454$ & $-0.355$ \\
Gemma-2B & $|$cos$| \le 0.015$ & 30 & 34 & 32 & 21 & 24 & 21 & 8 & $-0.319$ & $-0.359$ & $-0.334$ \\
Gemma-2B & any & 30 & 34 & 33 & 17 & 23 & 21 & 8 & $-0.292$ & $-0.357$ & $-0.298$ \\
Phi-3-mini & cos in $[-0.3, -0.1]$ & 30 & 32 & 33 & 12 & 29 & 22 & 12 & $-0.229$ & $-0.274$ & $-0.217$ \\
Phi-3-mini & $|$cos$| \le 0.015$ & 30 & 27 & 30 & 10 & 32 & 25 & 13 & $-0.174$ & $-0.249$ & $-0.176$ \\
Phi-3-mini & any & 30 & 25 & 29 & 9 & 35 & 30 & 14 & $-0.126$ & $-0.207$ & $-0.156$ \\
OLMo-1B & cos in $[-0.3, -0.1]$ & 50 & 37 & 39 & 19 & 24 & 10 & 7 & $-0.374$ & $-0.403$ & $-0.328$ \\
OLMo-1B & $|$cos$| \le 0.015$ & 11 & 22 & 21 & 9 & 27 & 17 & 11 & $-0.196$ & $-0.212$ & $-0.186$ \\
OLMo-1B & any & 50 & 32 & 34 & 16 & 23 & 12 & 8 & $-0.319$ & $-0.351$ & $-0.287$ \\
OLMo-1B, common documents & cos in $[-0.3, -0.1]$ & 50 & 36 & 37 & 17 & 24 & 12 & 7 & $-0.372$ & $-0.398$ & $-0.323$ \\
OLMo-1B, common documents & $|$cos$| \le 0.015$ & 50 & 24 & 23 & 9 & 23 & 16 & 9 & $-0.214$ & $-0.236$ & $-0.200$ \\
OLMo-1B, common documents & any & 50 & 35 & 37 & 19 & 25 & 12 & 8 & $-0.364$ & $-0.397$ & $-0.327$ \\
OLMoE-1B-7B & cos in $[-0.3, -0.1]$ & 30 & 29 & 28 & 12 & 32 & 14 & 8 & $-0.246$ & $-0.274$ & $-0.229$ \\
OLMoE-1B-7B & $|$cos$| \le 0.015$ & 30 & 26 & 25 & 10 & 28 & 14 & 11 & $-0.201$ & $-0.233$ & $-0.193$ \\
OLMoE-1B-7B & any & 30 & 26 & 23 & 10 & 29 & 18 & 15 & $-0.190$ & $-0.198$ & $-0.155$ \\
Llama-3.1-8B base & cos in $[-0.3, -0.1]$ & 30 & 37 & 37 & 14 & 28 & 17 & 9 & $-0.315$ & $-0.340$ & $-0.287$ \\
Llama-3.1-8B base & $|$cos$| \le 0.015$ & 30 & 16 & 11 & 3 & 29 & 20 & 10 & $-0.108$ & $-0.113$ & $-0.115$ \\
Llama-3.1-8B base & any & 30 & 32 & 32 & 14 & 31 & 18 & 10 & $-0.270$ & $-0.289$ & $-0.248$ \\
\bottomrule
\end{tabular}

%% file: supplement/table_eigen_r1.tex
\begin{tabular}{lrrrrrrrrrr}
\toprule
 & & \multicolumn{3}{c}{participation ratio / $(d-1)$} & \multicolumn{3}{c}{energy with damping $> 0.8$ (\%)} & \multicolumn{3}{c}{energy with damping $< 0$ (\%)} \\
\cmidrule(lr){3-5}\cmidrule(lr){6-8}\cmidrule(lr){9-11}
model & $d-1$ & write & another token's & random & write & another token's & random & write & another token's & random \\
\midrule
Llama-3.2-1B & 2047 & $0.30$ & $0.32$ & $0.34$ & $32$ & $32$ & $27$ & $22$ & $20$ & $22$ \\
Qwen2.5-0.5B & 895 & $0.28$ & $0.32$ & $0.33$ & $22$ & $22$ & $18$ & $28$ & $25$ & $28$ \\
TinyLlama-1.1B & 2047 & $0.25$ & $0.31$ & $0.33$ & $13$ & $12$ & $8$ & $38$ & $36$ & $36$ \\
Llama-2-7B & 4095 & $0.26$ & $0.32$ & $0.33$ & $9$ & $8$ & $5$ & $38$ & $36$ & $37$ \\
Mistral-7B & 4095 & $0.26$ & $0.31$ & $0.33$ & $9$ & $9$ & $6$ & $36$ & $35$ & $34$ \\
Llama-3.1-8B-Instruct & 4095 & $0.27$ & $0.32$ & $0.33$ & $15$ & $14$ & $11$ & $31$ & $29$ & $30$ \\
Gemma-2B & 2047 & $0.24$ & $0.32$ & $0.33$ & $20$ & $19$ & $17$ & $28$ & $26$ & $25$ \\
Phi-3-mini & 3071 & $0.24$ & $0.32$ & $0.33$ & $11$ & $9$ & $6$ & $40$ & $38$ & $39$ \\
OLMo-1B & 2047 & $0.30$ & $0.33$ & $0.34$ & $20$ & $20$ & $16$ & $27$ & $25$ & $27$ \\
OLMoE-1B-7B & 2047 & $0.29$ & $0.32$ & $0.33$ & $11$ & $10$ & $8$ & $40$ & $38$ & $38$ \\
\bottomrule
\end{tabular}

%% file: supplement/table_catsctl_r1.tex
\begin{tabular}{lrrrrrrrr}
\toprule
 & & \multicolumn{3}{c}{CATS at 50\%} & \multicolumn{4}{c}{open entries zeroed at random, loss matched} \\
\cmidrule(lr){3-5}\cmidrule(lr){6-9}
model & loss & $\Delta$loss & $\bar D$ ratio & $\Delta\cos$ & $p$ (\%) & $\Delta$loss & $\bar D$ ratio & $\Delta\cos$ \\
\midrule
Llama-3.2-1B & $2.531$ & $+0.257$ & $0.89$ & $+0.017$ & $9.1$ & $+0.257$ & $1.02$ & $-0.006$ \\
Qwen2.5-0.5B & $2.685$ & $+0.288$ & $0.81$ & $+0.021$ & $9.4$ & $+0.288$ & $1.01$ & $-0.005$ \\
TinyLlama-1.1B & $2.348$ & $+0.077$ & $0.87$ & $+0.012$ & $3.7$ & $+0.077$ & $1.02$ & $-0.004$ \\
Llama-2-7B & $2.018$ & $+0.056$ & $0.86$ & $+0.010$ & $5.5$ & $+0.056$ & $1.01$ & $-0.003$ \\
Mistral-7B & $2.049$ & $+0.207$ & $0.82$ & $+0.014$ & $7.0$ & $+0.207$ & $1.02$ & $-0.008$ \\
Llama-3.1-8B-Instruct & $2.399$ & $+0.121$ & $0.86$ & $+0.017$ & $8.2$ & $+0.121$ & $1.02$ & $-0.004$ \\
Gemma-2B & $2.459$ & $+0.349$ & $0.74$ & $+0.030$ & $12.0$ & $+0.349$ & $0.99$ & $+0.004$ \\
Phi-3-mini & $2.119$ & $+0.126$ & $0.88$ & $+0.007$ & $5.7$ & $+0.126$ & $1.00$ & $-0.001$ \\
OLMo-1B & $2.547$ & $+0.231$ & $0.89$ & $+0.007$ & $10.2$ & $+0.231$ & $0.98$ & $0.000$ \\
OLMoE-1B-7B & $2.327$ & $+0.154$ & $0.88$ & $+0.008$ & $13.1$ & $+0.154$ & $1.00$ & $-0.007$ \\
\bottomrule
\end{tabular}

%% file: figures/lobeoff.tex
\begin{tikzpicture}
\begin{groupplot}[group style={group size=3 by 1, horizontal sep=0.45cm, yticklabels at=edge left}, paper, scale only axis,
  width=\dimexpr0.333\linewidth-1.55cm\relax, height=4.2cm, ymin=-0.6, ymax=8.6, modelrows={figures/data/figlo.csv}{\FigGYTicks}, enlarge y limits=false]
\nextgroupplot[title={(a) window damping $\bar D$}, zero x, xmin=-0.08, xmax=0.5, xtick={0,0.2,0.4},
  legend to name=loleg, legend columns=3, legend style={/tikz/every even column/.append style={column sep=0.5cm}}]
\addplot[dot=black!75] table[x=D0, y=y] {figures/data/figlo.csv}; \addlegendentry{as released}
\addplot[diamond=series2] table[x=Dl, y=y] {figures/data/figlo.csv}; \addlegendentry{closed side removed}
\addplot[tri=series3] table[x=Do, y=y] {figures/data/figlo.csv}; \addlegendentry{size-matched open entries removed}
\nextgroupplot[title={(b) controlled cosine}, zero x, xmin=-0.3, xmax=0.3, xtick={-0.2,0,0.2}]
\addplot[dot=black!75] table[x=c0, y=y] {figures/data/figlo.csv}; \addplot[diamond=series2] table[x=cl, y=y] {figures/data/figlo.csv}; \addplot[tri=series3] table[x=co, y=y] {figures/data/figlo.csv};
\nextgroupplot[title={(c) loss (nats)}, xmin=0, xmax=14, xtick={0,4,8,12}]
\addplot[dot=black!75] table[x=l0, y=y] {figures/data/figlo.csv}; \addplot[diamond=series2] table[x=ll, y=y] {figures/data/figlo.csv}; \addplot[tri=series3] table[x=lo, y=y] {figures/data/figlo.csv};
\end{groupplot}
\end{tikzpicture}

\smallskip
\pgfplotslegendfromname{loleg}

%% file: figures/pertoken.tex
\begin{tikzpicture}
\begin{groupplot}[group style={group size=2 by 1, horizontal sep=0.5cm, yticklabels at=edge left}, paper, zero x, scale only axis,
  width=\dimexpr0.5\linewidth-2.2cm\relax, height=4.6cm, ymin=-0.6, ymax=8.6, modelrows={figures/data/figpt.csv}{\FigGYTicks}, enlarge y limits=false,
  xmin=-0.75, xmax=0.2, xtick={-0.6,-0.4,-0.2,0,0.2}, xlabel={partial Spearman correlation with the projection},
  error bars/error bar style={line width=0.7pt}, error bars/error mark options={draw=none}]
\nextgroupplot[title={(a) lobe term}, legend to name=ptleg, legend columns=4, legend style={/tikz/every even column/.append style={column sep=0.4cm}}]
\addplot[dot=series1, error bars/.cd, x dir=both, x explicit] table[x=lp, y expr=\thisrow{y}+0.17, x error plus=lpp, x error minus=lpm] {figures/data/figpt.csv}; \addlegendentry{lobe term}
\addplot[diamond=series2, error bars/.cd, x dir=both, x explicit] table[x=lt, y expr=\thisrow{y}-0.17, x error plus=ltp, x error minus=ltm] {figures/data/figpt.csv}; \addlegendentry{lobe term, token class and frequency also controlled}
\addlegendimage{tri=series3}\addlegendentry{weakening term}
\addlegendimage{sq=black!80}\addlegendentry{total damping}
\nextgroupplot[title={(b) weakening term and total damping}]
\addplot[tri=series3, error bars/.cd, x dir=both, x explicit] table[x=wp, y expr=\thisrow{y}+0.17, x error plus=wpp, x error minus=wpm] {figures/data/figpt.csv};
\addplot[sq=black!80, error bars/.cd, x dir=both, x explicit] table[x=tp, y expr=\thisrow{y}-0.17, x error plus=tpp, x error minus=tpm] {figures/data/figpt.csv};
\end{groupplot}
\end{tikzpicture}

\smallskip
\pgfplotslegendfromname{ptleg}

%% file: figures/arms.tex
\begin{tikzpicture}
\begin{groupplot}[group style={group size=4 by 1, horizontal sep=0.45cm, yticklabels at=edge left}, paper, scale only axis,
  width=\dimexpr0.25\linewidth-1.05cm\relax, height=2.9cm, ymin=-0.55, ymax=5.55, ytick={5,4,3,2,1,0},
  yticklabels={{$\lambda = 0$}, {$\lambda = 0.5$}, {$\lambda = 1$ (SwiGLU)}, {$\lambda = 2$}, {no gradient}, {straight-through}},
  y tick label style={font=\scriptsize, text=black}, major y tick style={draw=none}, ymajorgrids=false, xmajorgrids=true]
\nextgroupplot[title={(a) window damping}, xmin=-0.005, xmax=0.08, xtick={0,0.02,0.04,0.06}, xticklabel style={/pgf/number format/precision=2},
  legend style={at={(0.98,0.98)}, anchor=north east}]
\addplot[dot=black!75] table[x=D, y=y] {figures/data/figarms.csv}; \addlegendentry{$\bar D$}
\addplot[diamond=series1] table[x=lobe, y=y] {figures/data/figarms.csv}; \addlegendentry{lobe term}
\nextgroupplot[title={(b) gates closed}, xmin=0.5, xmax=0.95, xtick={0.6,0.7,0.8,0.9}]
\addplot[dot=series1] table[x=closed, y=y] {figures/data/figarms.csv};
\nextgroupplot[title={(c) median read--write cosine}, zero x, xmin=-0.04, xmax=0.18, xtick={0,0.05,0.1,0.15}]
\addplot[dot=series1] table[x=rw, y=y] {figures/data/figarms.csv};
\nextgroupplot[title={(d) validation loss}, xmin=3.4, xmax=3.85, xtick={3.4,3.5,3.6,3.7,3.8}, xticklabel style={/pgf/number format/precision=1}]
\addplot[dot=series1] table[x=val, y=y] {figures/data/figarms.csv};
\end{groupplot}
\end{tikzpicture}

%% file: table_arms_r1.tex
\begin{tabular}{llrrrrrr}
\toprule
arm & closed side (forward, gradient) & $\bar D$ & lobe term & read--write & str.\ share & closed & loss \\
\midrule
$\lambda = 0$ & $0$, $0$ & $0.0148$, $0.0134$ & $0.0000$, $0.0000$ & $-0.018$, $-0.013$ & $0.45$, $0.46$ & $0.83$, $0.81$ & $3.440$, $3.438$ \\
$\lambda = 0.5$ & $0.5\,\phi$, $0.5\,\phi$ & $0.0266$, $0.0256$ & $0.0190$, $0.0175$ & $+0.108$, $+0.100$ & $0.76$, $0.74$ & $0.78$, $0.77$ & $3.441$, $3.440$ \\
$\lambda = 1$ & $\phi$, $\phi$ (SwiGLU) & $0.0441$, $0.0451$ & $0.0370$, $0.0380$ & $+0.146$, $+0.151$ & $0.79$, $0.79$ & $0.71$, $0.71$ & $3.444$, $3.437$ \\
$\lambda = 2$ & $2\phi$, $2\phi$ & $0.0628$, $0.0570$ & $0.0457$, $0.0429$ & $+0.114$, $+0.120$ & $0.70$, $0.70$ & $0.58$, $0.59$ & $3.464$, $3.455$ \\
no gradient & $\phi$, $0$ & $0.0232$, $0.0276$ & $0.0092$, $0.0134$ & $-0.005$, $+0.014$ & $0.49$, $0.55$ & $0.88$, $0.89$ & $3.484$, $3.493$ \\
straight-through & $0$, $\phi$ & $0.0747$, $0.0693$ & $0.0000$, $0.0000$ & $+0.008$, $+0.005$ & $0.51$, $0.52$ & $0.60$, $0.61$ & $3.764$, $3.809$ \\
\bottomrule
\end{tabular}

%% file: supplement/table_contpre_qwen05_r1.tex
\begin{tabular}{rrrrrrrrrrrrr}
\toprule
 & & & & \multicolumn{5}{c}{closed side annealed away} & \multicolumn{4}{c}{control (activation unchanged)} \\
\cmidrule(lr){5-9}\cmidrule(lr){10-13}
seed & step & tokens (M) & $\lambda$ & loss & $\bar D$ & lobe & $D$ & $\cos$ & loss & $\bar D$ & $D$ & $\cos$ \\
\midrule
0 & 0 & 0 & $1.00$ & $2.685$ & $0.304$ & $0.307$ & $0.336$ & $-0.184$ & $2.685$ & $0.304$ & $0.336$ & $-0.184$ \\
0 & 800 & 52 & $0.67$ & $2.779$ & $0.250$ & $0.245$ & $0.267$ & $-0.143$ & $2.772$ & $0.282$ & $0.307$ & $-0.162$ \\
0 & 1600 & 105 & $0.22$ & $2.871$ & $0.161$ & $0.144$ & $0.160$ & $-0.082$ & $2.755$ & $0.282$ & $0.308$ & $-0.161$ \\
0 & 2400 & 157 & $0.00$ & $2.967$ & $0.017$ & $0.000$ & $-0.005$ & $+0.023$ & $2.735$ & $0.285$ & $0.312$ & $-0.164$ \\
0 & 3200 & 210 & $0.00$ & $2.900$ & $0.019$ & $0.000$ & $-0.003$ & $+0.022$ & $2.728$ & $0.289$ & $0.318$ & $-0.168$ \\
0 & 3800 & 249 & $0.00$ & $2.892$ & $0.020$ & $0.000$ & $-0.002$ & $+0.021$ & $2.726$ & $0.291$ & $0.320$ & $-0.170$ \\
\midrule
1 & 0 & 0 & $1.00$ & $2.685$ & $0.304$ & $0.307$ & $0.336$ & $-0.184$ & $2.685$ & $0.304$ & $0.336$ & $-0.184$ \\
1 & 800 & 52 & $0.67$ & $2.786$ & $0.250$ & $0.245$ & $0.267$ & $-0.143$ & $2.770$ & $0.280$ & $0.306$ & $-0.162$ \\
1 & 1600 & 105 & $0.22$ & $2.866$ & $0.160$ & $0.144$ & $0.160$ & $-0.082$ & $2.771$ & $0.282$ & $0.307$ & $-0.162$ \\
1 & 2400 & 157 & $0.00$ & $2.964$ & $0.017$ & $0.000$ & $-0.005$ & $+0.022$ & $2.757$ & $0.285$ & $0.312$ & $-0.165$ \\
1 & 3200 & 210 & $0.00$ & $2.903$ & $0.019$ & $0.000$ & $-0.003$ & $+0.021$ & $2.730$ & $0.290$ & $0.318$ & $-0.169$ \\
1 & 3800 & 249 & $0.00$ & $2.888$ & $0.019$ & $0.000$ & $-0.003$ & $+0.021$ & $2.723$ & $0.291$ & $0.319$ & $-0.170$ \\
\midrule
2 & 0 & 0 & $1.00$ & $2.685$ & $0.304$ & $0.307$ & $0.336$ & $-0.184$ & $2.685$ & $0.304$ & $0.336$ & $-0.184$ \\
2 & 800 & 52 & $0.67$ & $2.788$ & $0.251$ & $0.246$ & $0.269$ & $-0.143$ & $2.778$ & $0.283$ & $0.309$ & $-0.164$ \\
2 & 1600 & 105 & $0.22$ & $2.866$ & $0.160$ & $0.144$ & $0.159$ & $-0.083$ & $2.763$ & $0.281$ & $0.307$ & $-0.162$ \\
2 & 2400 & 157 & $0.00$ & $2.960$ & $0.017$ & $0.000$ & $-0.004$ & $+0.022$ & $2.750$ & $0.285$ & $0.312$ & $-0.165$ \\
2 & 3200 & 210 & $0.00$ & $2.900$ & $0.019$ & $0.000$ & $-0.003$ & $+0.022$ & $2.726$ & $0.289$ & $0.317$ & $-0.168$ \\
2 & 3800 & 249 & $0.00$ & $2.883$ & $0.020$ & $0.000$ & $-0.002$ & $+0.021$ & $2.720$ & $0.291$ & $0.320$ & $-0.170$ \\
\bottomrule
\end{tabular}

%% file: supplement/table_contpre_llama1b_r1.tex
\begin{tabular}{rrrrrrrrrrrrr}
\toprule
 & & & & \multicolumn{5}{c}{closed side annealed away} & \multicolumn{4}{c}{control (activation unchanged)} \\
\cmidrule(lr){5-9}\cmidrule(lr){10-13}
seed & step & tokens (M) & $\lambda$ & loss & $\bar D$ & lobe & $D$ & $\cos$ & loss & $\bar D$ & $D$ & $\cos$ \\
\midrule
0 & 0 & 0 & $1.00$ & $2.531$ & $0.455$ & $0.470$ & $0.497$ & $-0.222$ & $2.531$ & $0.455$ & $0.497$ & $-0.222$ \\
0 & 800 & 52 & $0.67$ & $2.687$ & $0.356$ & $0.370$ & $0.375$ & $-0.154$ & $2.675$ & $0.394$ & $0.422$ & $-0.177$ \\
0 & 1600 & 105 & $0.22$ & $3.005$ & $0.209$ & $0.221$ & $0.194$ & $-0.043$ & $2.765$ & $0.392$ & $0.419$ & $-0.167$ \\
0 & 2400 & 157 & $0.00$ & $3.107$ & $-0.027$ & $0.000$ & $-0.078$ & $+0.151$ & $2.805$ & $0.393$ & $0.421$ & $-0.169$ \\
0 & 3200 & 210 & $0.00$ & $3.101$ & $-0.030$ & $0.000$ & $-0.082$ & $+0.151$ & $2.782$ & $0.402$ & $0.433$ & $-0.177$ \\
0 & 3800 & 249 & $0.00$ & $3.108$ & $-0.030$ & $0.000$ & $-0.079$ & $+0.148$ & $2.767$ & $0.406$ & $0.438$ & $-0.181$ \\
\bottomrule
\end{tabular}

%% file: figures/contpre_LlamaThreeOneB.tex
\begin{tikzpicture}
\begin{groupplot}[group style={group size=3 by 1, horizontal sep=1.2cm}, width=0.33\linewidth, height=3.1cm, paper, xlabel={training tokens (millions)}, xmin=0, xmax=\CpTokensM]
\nextgroupplot[zero y, title={(a) along the previous write}, ymin=-0.25, ymax=0.4,
  legend to name=cplegLlamaThreeOneB, legend columns=6, legend style={/tikz/every even column/.append style={column sep=0.3cm}}]
\addplot[name path=aDhi, draw=none, forget plot] table[x=tok, y=aDmax] {figures/data/figcpLlamaThreeOneB.csv};
\addplot[name path=aDlo, draw=none, forget plot] table[x=tok, y=aDmin] {figures/data/figcpLlamaThreeOneB.csv};
\addplot[series1, fill opacity=0.18, draw=none, forget plot] fill between[of=aDhi and aDlo];
\addplot[name path=cDhi, draw=none, forget plot] table[x=tok, y=cDmax] {figures/data/figcpLlamaThreeOneB.csv};
\addplot[name path=cDlo, draw=none, forget plot] table[x=tok, y=cDmin] {figures/data/figcpLlamaThreeOneB.csv};
\addplot[series1, fill opacity=0.18, draw=none, forget plot] fill between[of=cDhi and cDlo];
\addplot[name path=aAChi, draw=none, forget plot] table[x=tok, y=aACmax] {figures/data/figcpLlamaThreeOneB.csv};
\addplot[name path=aAClo, draw=none, forget plot] table[x=tok, y=aACmin] {figures/data/figcpLlamaThreeOneB.csv};
\addplot[series2, fill opacity=0.18, draw=none, forget plot] fill between[of=aAChi and aAClo];
\addplot[name path=cAChi, draw=none, forget plot] table[x=tok, y=cACmax] {figures/data/figcpLlamaThreeOneB.csv};
\addplot[name path=cAClo, draw=none, forget plot] table[x=tok, y=cACmin] {figures/data/figcpLlamaThreeOneB.csv};
\addplot[series2, fill opacity=0.18, draw=none, forget plot] fill between[of=cAChi and cAClo];
\addplot[name path=acoshi, draw=none, forget plot] table[x=tok, y=acosmax] {figures/data/figcpLlamaThreeOneB.csv};
\addplot[name path=acoslo, draw=none, forget plot] table[x=tok, y=acosmin] {figures/data/figcpLlamaThreeOneB.csv};
\addplot[black, fill opacity=0.18, draw=none, forget plot] fill between[of=acoshi and acoslo];
\addplot[name path=ccoshi, draw=none, forget plot] table[x=tok, y=ccosmax] {figures/data/figcpLlamaThreeOneB.csv};
\addplot[name path=ccoslo, draw=none, forget plot] table[x=tok, y=ccosmin] {figures/data/figcpLlamaThreeOneB.csv};
\addplot[black, fill opacity=0.18, draw=none, forget plot] fill between[of=ccoshi and ccoslo];
\addplot[series1, mark=*, mark size=1.3pt] table[x=tok, y=aD] {figures/data/figcpLlamaThreeOneB.csv}; \addlegendentry{$D$, annealed}
\addplot[series1, dashed, mark=o, mark size=1.3pt] table[x=tok, y=cD] {figures/data/figcpLlamaThreeOneB.csv}; \addlegendentry{$D$, control}
\addplot[series2, mark=square*, mark size=1.3pt] table[x=tok, y=aAC] {figures/data/figcpLlamaThreeOneB.csv}; \addlegendentry{$A + C$, annealed}
\addplot[series2, dashed, mark=square, mark size=1.3pt] table[x=tok, y=cAC] {figures/data/figcpLlamaThreeOneB.csv}; \addlegendentry{$A + C$, control}
\addplot[black!70, mark=triangle*, mark size=1.5pt] table[x=tok, y=acos] {figures/data/figcpLlamaThreeOneB.csv}; \addlegendentry{cosine, annealed}
\addplot[black!70, dashed, mark=triangle, mark size=1.5pt] table[x=tok, y=ccos] {figures/data/figcpLlamaThreeOneB.csv}; \addlegendentry{cosine, control}
\nextgroupplot[zero y, title={(b) average damping and lobe term}, ymin=-0.02, ymax=0.35]
\addplot[name path=aDbarhi, draw=none, forget plot] table[x=tok, y=aDbarmax] {figures/data/figcpLlamaThreeOneB.csv};
\addplot[name path=aDbarlo, draw=none, forget plot] table[x=tok, y=aDbarmin] {figures/data/figcpLlamaThreeOneB.csv};
\addplot[series1, fill opacity=0.18, draw=none, forget plot] fill between[of=aDbarhi and aDbarlo];
\addplot[name path=cDbarhi, draw=none, forget plot] table[x=tok, y=cDbarmax] {figures/data/figcpLlamaThreeOneB.csv};
\addplot[name path=cDbarlo, draw=none, forget plot] table[x=tok, y=cDbarmin] {figures/data/figcpLlamaThreeOneB.csv};
\addplot[series1, fill opacity=0.18, draw=none, forget plot] fill between[of=cDbarhi and cDbarlo];
\addplot[name path=alobehi, draw=none, forget plot] table[x=tok, y=alobemax] {figures/data/figcpLlamaThreeOneB.csv};
\addplot[name path=alobelo, draw=none, forget plot] table[x=tok, y=alobemin] {figures/data/figcpLlamaThreeOneB.csv};
\addplot[series3, fill opacity=0.18, draw=none, forget plot] fill between[of=alobehi and alobelo];
\addplot[name path=clobehi, draw=none, forget plot] table[x=tok, y=clobemax] {figures/data/figcpLlamaThreeOneB.csv};
\addplot[name path=clobelo, draw=none, forget plot] table[x=tok, y=clobemin] {figures/data/figcpLlamaThreeOneB.csv};
\addplot[series3, fill opacity=0.18, draw=none, forget plot] fill between[of=clobehi and clobelo];
\addplot[series1, mark=*, mark size=1.3pt] table[x=tok, y=aDbar] {figures/data/figcpLlamaThreeOneB.csv};
\addplot[series1, dashed, mark=o, mark size=1.3pt] table[x=tok, y=cDbar] {figures/data/figcpLlamaThreeOneB.csv};
\addplot[series3, mark=diamond*, mark size=1.5pt] table[x=tok, y=alobe] {figures/data/figcpLlamaThreeOneB.csv};
\addplot[series3, dashed, mark=diamond, mark size=1.5pt] table[x=tok, y=clobe] {figures/data/figcpLlamaThreeOneB.csv};
\nextgroupplot[title={(c) loss on the common documents, $\lambda$}, ymin=2.6, ymax=3.05, axis y line*=left]
\addplot[name path=alosshi, draw=none, forget plot] table[x=tok, y=alossmax] {figures/data/figcpLlamaThreeOneB.csv};
\addplot[name path=alosslo, draw=none, forget plot] table[x=tok, y=alossmin] {figures/data/figcpLlamaThreeOneB.csv};
\addplot[black, fill opacity=0.18, draw=none, forget plot] fill between[of=alosshi and alosslo];
\addplot[name path=closshi, draw=none, forget plot] table[x=tok, y=clossmax] {figures/data/figcpLlamaThreeOneB.csv};
\addplot[name path=closslo, draw=none, forget plot] table[x=tok, y=clossmin] {figures/data/figcpLlamaThreeOneB.csv};
\addplot[black, fill opacity=0.18, draw=none, forget plot] fill between[of=closshi and closslo];
\addplot[black!85, mark=*, mark size=1.3pt] table[x=tok, y=aloss] {figures/data/figcpLlamaThreeOneB.csv};
\addplot[black!85, dashed, mark=o, mark size=1.3pt] table[x=tok, y=closs] {figures/data/figcpLlamaThreeOneB.csv};
\end{groupplot}
\begin{axis}[at={(group c3r1.south west)}, anchor=south west, width=0.33\linewidth, height=3.1cm, xmin=0, xmax=\CpTokensM, ymin=0, ymax=1.05, axis y line*=right, axis x line=none, ylabel={$\lambda$}, ylabel near ticks, ytick={0,0.5,1}, tick label style={font=\scriptsize, text=inksecond}, label style={font=\footnotesize}, axis line style={draw=axisgray, line width=0.4pt}, tick style={draw=axisgray, line width=0.4pt}]
\addplot[inkmuted, thick, no marks] table[x=tok, y=lam] {figures/data/figcpLlamaThreeOneB.csv};
\end{axis}
\end{tikzpicture}

\smallskip
\pgfplotslegendfromname{cplegLlamaThreeOneB}